\documentclass[11pt]{article}

\usepackage[final]{acl}
\usepackage{times}
\usepackage{latexsym}
\usepackage[T1]{fontenc}
\usepackage[utf8]{inputenc}
\usepackage{microtype}
\usepackage{inconsolata}
\usepackage{graphicx}
\usepackage{multirow}
\usepackage{booktabs}
\usepackage{xcolor}

\usepackage{tabularx}
\usepackage{array}
\usepackage{colortbl}
\usepackage{makecell}

\definecolor{headergray}{HTML}{E8E8E8}
\definecolor{rowalt}{HTML}{F5F5F5}
\definecolor{metarow}{HTML}{F0F0F0}

\renewcommand{\arraystretch}{1.35}
\newcommand{\user}[1]{\textit{#1}}

\usepackage{subcaption} 

\usepackage{soul} 
\usepackage{hyperref}
\usepackage{xurl}

\usepackage[most]{tcolorbox}
\usepackage{listings}
\usepackage{enumitem}
\usepackage{xcolor}

\usepackage{booktabs}
\usepackage{multirow}
\usepackage{adjustbox}   
\usepackage{longtable}
\usepackage{pdflscape}  

\newcommand{\data}{Wiki-Talkie}
\newcommand{\base}{\texttt{base}}
\newcommand{\profile}{\texttt{profile}}
\newcommand{\behavior}{\texttt{behavior}}
\newcommand{\behaviorreasoning}{\texttt{behavior+r}}
\newcommand{\history}{\texttt{history}}

\newcommand{\modelA}{Model~A}   
\newcommand{\modelB}{Model~B}   
\newcommand{\modelC}{Model~C}   
\newcommand{\modelD}{Model~D}   
\newcommand{\extractor}{the extractor model}

\title{Wiki-Talkie: Multilingual Benchmarking of \\ Persona-Based Agents on Real-World Discussions}

\author{
  Dennis Fucci\thanks{Work done during an internship at Amazon.} \\
  \texttt{dennisfucci@gmail.com} \\[0.6em]
  \textbf{Andrea Bacciu} \quad \textbf{Dong Liu} \quad \textbf{Weronika Łajewska} \quad \textbf{Saab Mansour} \\
  Amazon \\
  \texttt{\{andbac, liuadong, lajewska, saabm\}@amazon.com} \\
}

\begin{document}
\maketitle
\begin{abstract}

LLMs are increasingly deployed as autonomous agents in social environments, making it critical to study their ability to faithfully simulate human interactions. Central to this is grounding agents in realistic user personas, yet existing datasets rely on fictional personas and are limited to a handful of languages, lacking the empirical grounding necessary to evaluate behavioral fidelity across diverse populations. We introduce \data, a multilingual dataset of real-world conversations from Wikipedia Talk pages across five languages spanning two language families: Germanic (German, English) and Romance (Spanish, French, Italian), paired with personas derived from real user communities and encompassing sociodemographic attributes, self-descriptions, and behaviorally grounded interaction traits. Using \data, we evaluate agent interactional behavior on a next-turn generation task across various persona conditioning strategies. Our evaluation assesses whether agents collectively reproduce the distributional behavioral patterns observed in human discussions. Results show that user's comment history exemplifying interaction behavior consistently outperforms explicit persona information. In addition, models systematically underproduce negative or extreme sentiments, while over producing references and suggestions, revealing biases toward agreeableness and positivity. Crucially, these patterns hold robustly across languages, with small cross-lingual differences.\footnote{Code and data are available at \url{https://github.com/amazon-science/wikitalkie}.}
\end{abstract}

\section{Introduction}

Large Language Models (LLMs) are increasingly embedded in real-world social systems, from community moderation on Reddit\footnote{\url{https://redditinc.com/news/oterlu-team-joins-reddit-to-accelerate-safety-efforts}.} to experimental platforms like MoltBook\footnote{\url{https://www.moltbook.com}.} where LLMs autonomously interact at scale. These developments position LLMs not merely as tools, but as \textit{agents}: autonomous actors participating in governance, culture, and social interaction.
Central to the design of such agents is the \textit{role} they are asked to play \citep{shanahan-etal-2023-role, tseng-etal-2024-two-tales}, e.g. the \textit{persona}, that is, a composite of individual traits such as background facts, language behavior, and interaction style \citep{liu-etal-2025-persona}.  This approach has been shown to improve alignment with real human populations in applications such as personalized dialogues \citep{liu-etal-2025-persona}, linguistic annotation \citep{orlikowski-etal-2025-beyond}, and social simulations \citep{argyle-etal-2023-one, park-etal-2024-empowering}.

However, existing datasets for modeling persona-based conversational agents rely on fictional personas \citep{smith-etal-2020-put, jang-etal-2022-focus, xu-etal-2022-beyond, xu-etal-2022-long, yamashita-etal-2023-realpersonachat, zhao-etal-2025-personalens, lotfi-etal-2023-personalitychat, chen-etal-2022-cped, occhipinti-etal-2024-prodigy}, making it impossible to verify whether agent behavior faithfully reflects the opinions, linguistic patterns, and interaction dynamics of the actual human population being simulated.
Datasets extracting personas from large-scale media \citep{zheng-etal-2020-personaldialog, wu-etal-2021-personalized} offer greater realism but are limited to Chinese or English. To our best knowledge, there is no resource providing real-world, multilingual dialogues grounded in real personas within a cooperative setting where pragmatic phenomena such as conflict, misalignment, and evolving viewpoints can emerge---precisely the dynamics that characterize high-stakes scenarios.

To fill this gap, we introduce \textbf{\data}, a multilingual dataset of real-world discussions extracted from Wikipedia Talk pages across five languages, where registered users collaboratively deliberate on article edits. \data{} enables multifaceted evaluation of agents in a consensus-oriented discussion scenario, with optional conditioning on personas derived from Wikipedia user pages, and behavioral traits from comment histories.
To safeguard user privacy, all identities are anonymized and persona profiles are regenerated in a neutral fashion rather than verbatim copies, ensuring privacy protection while maintaining research utility.

Our evaluation targets what we term \textit{population fidelity}: whether agents collectively reproduce the distributional behavioral patterns observed in human discussions, as opposed to \textit{individual-level persona fidelity}, i.e., whether a single generated response faithfully reflects the provided persona of a specific user. This choice is principled: individual-level fidelity risks enabling faithful simulation of specific individuals, raising ethical concerns around impersonation and lack of consent \citep{desai-etal-2025-personas-evolved}, while population fidelity is more tractable and more relevant for applications such as community simulation and policy analysis. This aligns with an emerging consensus that evaluation of persona-based agents should operate at the population level \citep{mansour-etal-2025-paars, rupprecht-etal-2025-german-personas, fawaz-etal-2026-narriva}. To assess population fidelity, we frame
the task as \textit{next-turn generation}: given a thread of previous comments, an agent must generate the next comment simulating a specific Wikipedia editor. We conduct evaluations with multiple metrics: instance-level text similarity (ROUGE-L and BERTScore), which provides individual-level signals averaged across the entire dataset, and population-level interactional behavior, comparing sentiment and dialog act distributions \citep{searle-1969-speechacts} between human comments and those generated by agents powered by multiple state-of-the-art LLMs, across five languages.
We study how agents behave when conditioned on different types of persona information, comparing behavioral traits extracted from comment histories against explicit persona descriptions derived from user profiles. Results show that cross-lingual differences are small, consistent with prior work on multilingual LLM evaluation showing strong performance across high-resource languages \cite{xuan-etal-2025-mmlu, li-etal-2025-languageranker}. We also find that comment history consistently proves more effective than explicit persona information as a conditioning strategy, suggesting that \textit{showing} how a persona behaves outperforms \textit{telling} what a persona is.

Overall, our contributions are: \textit{1)} the construction of \data{}, a dataset of Wikipedia discussions supporting research on agentic interaction simulation, whose key advantages are: \textit{a)} multilingual coverage; \textit{b)} real-world discussions reflecting established communication norms; \textit{c)} ground-truth persona information linked to discussions; and \textit{d)} rich contextual information, with discussions paired with articles; \textit{2)} the evaluation of agents based on four LLMs under various conditioning strategies across five languages, providing insights into the limitations of current agentic simulation of human cooperative interactions.

\section{Related Work}

\paragraph{Datasets for Persona-based Evaluation}
Conversational datasets with persona-related attributes vary in source, language, and use case. Crowdsourced datasets dominate, exemplified by PersonaChat \cite{zhang-etal-2018-personachat}, where crowdworkers converse under fictional personas, and its many extensions: BlendedSkillTalk \citep{smith-etal-2020-put} adds empathy and knowledge, FoCus \citep{jang-etal-2022-focus} adds Wikipedia grounding, Multi-Session Chat \citep{xu-etal-2022-beyond} models longer-term interaction, DuLeMon \citep{xu-etal-2022-long} uses Chinese mutual personas, and RealPersonaChat \citep{yamashita-etal-2023-realpersonachat} collects real persona descriptions in Japanese. Other datasets rely on synthetic generation \citep{zhao-etal-2025-personalens, lotfi-etal-2023-personalitychat} or scripted content \citep{chen-etal-2022-cped, occhipinti-etal-2024-prodigy}, while social media datasets such as PersonalDialog \citep{zheng-etal-2020-personaldialog} and PER-CHAT \citep{wu-etal-2021-personalized} extract persona signals from large-scale platforms. Most rely on simulated role-play or synthetic generation and cover few languages: although XPersona \citep{lin-etal-2021-xpersona} extends PersonaChat multilingually via automatic translation, naturally occurring interaction datasets remain largely restricted to English and Chinese. Critically, no existing resource simultaneously provides: \textit{(i)} naturally occurring dialogues, \textit{(ii)} multilingual coverage, \textit{(iii)} real user-derived personas linked to the conversations, and \textit{(iv)} a cooperative yet contentious setting where disagreement, negotiation, and consensus-building emerge. We address this gap with \data{}.

\paragraph{LLM Social Behavior}
A growing body of work documents systematic biases in LLM-generated social behavior. LLMs exhibit sycophantic tendencies, preferring agreement over truthful disagreement \citep{malmqvist_etal_2025_sycophancy}, and display a positivity bias that suppresses negative or confrontational outputs \citep{santurkar-etal-2023-opinions}. Studies using LLMs to simulate human populations have shown that, while persona conditioning can improve alignment with survey response distributions \citep{argyle-etal-2023-one}, models often fail to reproduce the full diversity of human opinions and behaviors \citep{orlikowski-etal-2025-beyond}. However, these findings are predominantly based on English-only settings, synthetic survey tasks, or opinion elicitation, leaving open the question of whether such biases persist in naturalistic, multilingual, cooperative discussions. \data{} addresses this gap by providing a testbed where LLM social behavior can be evaluated against real human interactions across five languages in a collaborative setting with genuine disagreement and pragmatic complexity.

\paragraph{Datasets Derived from Wikipedia Talks} 
Wikipedia Talk pages remain underexplored despite providing large-scale, naturally occurring collaborative dialogue \cite{johnson-etal-2024-wikimedia}. Prior work has used them to study conversational behavior and harmful interactions \citep{wulczyn-etal-2017-personal-attacks, zhang-etal-2018-conversations}, dialogue act classification \cite{ferschke-etal-2012-behind}, dispute detection \cite{wang-cardie-2014-piece}, sociolinguistics \cite{prabhakaran-rambow-2016-corpus}, abuse \citep{cecillon-etal-2020-wac}, and links between discussions and edit histories \citep{jaidka-etal-2021-wikitalkedit}. WikiConv \citep{hua-etal-2018-wikiconv} reconstructs full conversational histories including edits and deletions, supporting tasks such as preemptive toxicity detection \citep{karan-snajder-2019-preemptive}. However, these datasets focus on discourse structure, conflict, or toxicity, remain mostly English-centric, and lack persistent persona representations for multilingual evaluation. To our knowledge, Talk pages have not been used to analyze LLM behavior in socially grounded, collaborative settings.

\section{Wiki-Talkie}

\begin{figure}[t!]
    \centering
    \includegraphics[width=0.9\textwidth, angle=90]{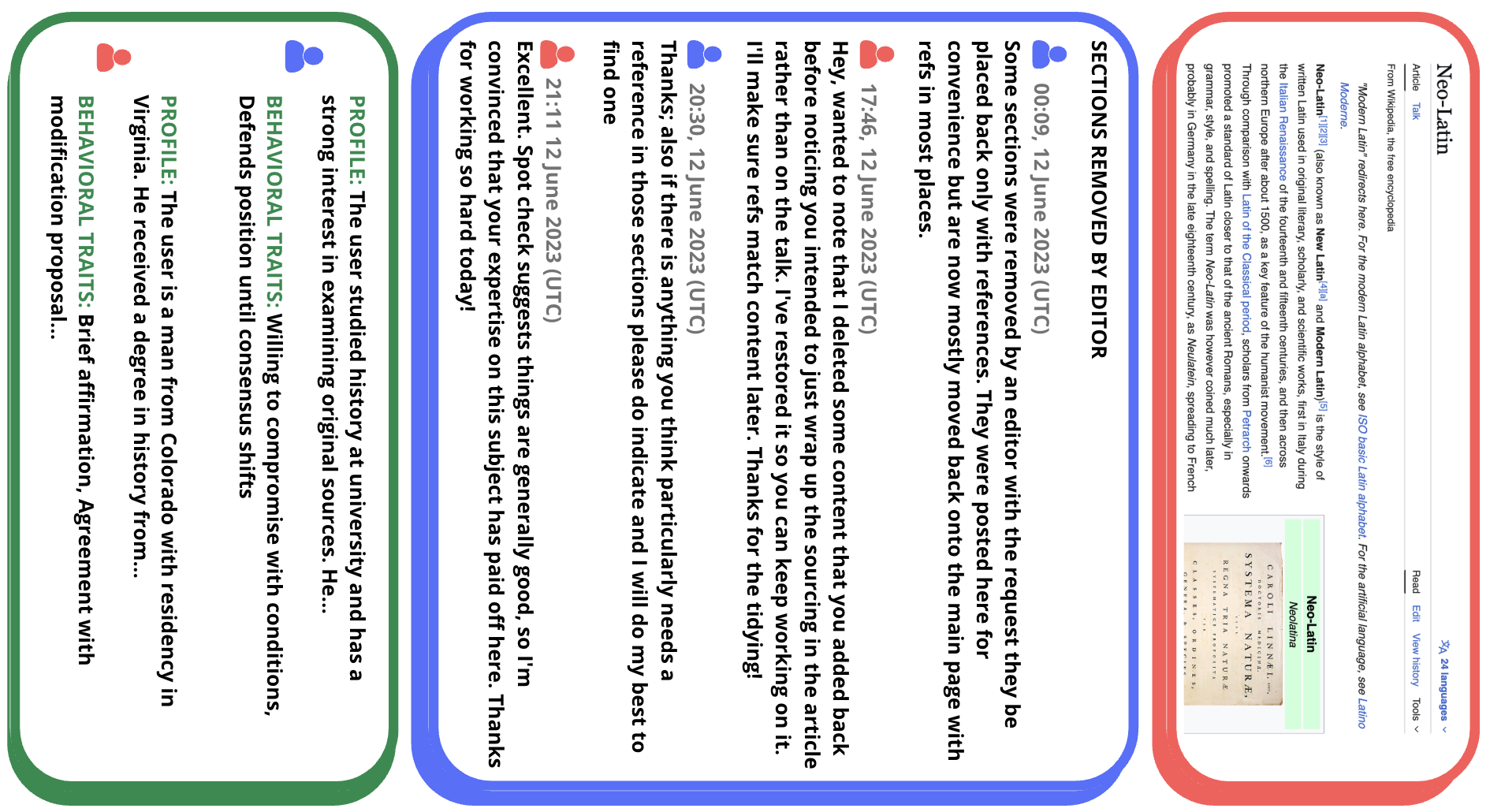}
    \caption{Example from \data{} with its three core components: \textcolor{red}{article}, \textcolor{blue}{thread}, and \textcolor{green!60!black}{persona}.}
    \label{fig:wikitalkie}
\end{figure}

We introduce \textbf{\data{}}, a multilingual dataset of real-world discussions extracted from Wikipedia Talk pages across five languages spanning two language families: Germanic (German, English) and Romance (Spanish, French, Italian). \data{} is built around three core components (see Figure~\ref{fig:wikitalkie}): \textit{1)} multilingual \textbf{threads} in which users deliberate on article content; \textit{2)} user \textbf{personas} from two complementary sources, a \textit{self-reported profile} derived from Wikipedia user pages (with attributes such as gender, profession, interests, and contribution topics) and \textit{behavioral traits} inferred from each user's comment history; and \textit{3)} the \textbf{article} context, paired with each thread as it appeared immediately before the discussion began. Together, these components make \data{} suited for research on persona-conditioned and document-grounded interaction modeling across languages.\footnote{Examples are in Appendix~\ref{app:examples}, Tables~\ref{tab:de}--\ref{tab:it}.}

Table~\ref{tab:threads_summary} presents summary statistics of \data{} across languages (full distributions in Appendix~\ref{app:statistics_distributions}). German contributes the largest portion (>42K threads, >134K turns), followed by French and English, while Spanish and Italian form smaller subsets. Despite differences in scale, structural characteristics are remarkably consistent across languages: threads contain on average 3.2--3.6 turns and involve 2.1--2.3 unique users.\footnote{These statistics are not necessarily representative of Wikipedia Talk pages at large, but reflect our construction pipeline (Section~\ref{sec:dataset_construction}).} Turn length varies more, ranging from 47.74 words (German) to 79.73 (Spanish), potentially reflecting differences in editorial culture or linguistic verbosity. High standard deviations across all metrics indicate substantial variability, consistent with discussions ranging from brief clarifications to extended debates. The number of unique users mirrors the thread distribution; further details on users and personas are in Appendix~\ref{app:statistics_users}.

\begin{table*}[ht]
    \centering
    \small
    \begin{tabular}{c|rrccccc}
    \toprule
    & \textbf{\# Threads} & \textbf{\# Turns} & \textbf{\# Users} & \textbf{Turns/Thread} & \textbf{Turns/User} & \textbf{Turn Length} & \textbf{Users/Thread} \\
    \midrule
    \texttt{de} & 38,711 & 123,520 & 6,282 & 3.19 $\pm$ 1.92 & 19.93 $\pm$ 53.93 & 48.09 $\pm$ 56.80 & 2.11 $\pm$ 0.35 \\
    \texttt{en} & 17,645 & 59,537 & 4,580 & 3.37 $\pm$ 2.33 & 13.19 $\pm$ 40.01 & 67.30 $\pm$ 81.68 & 2.06 $\pm$ 0.24 \\
    \texttt{es} & 5,229 & 17,019 & 2,307 & 3.25 $\pm$ 2.13 & 7.55 $\pm$ 19.41 & 79.80 $\pm$ 96.36 & 2.14 $\pm$ 0.42 \\
    \texttt{fr} & 23,983 & 86,041 & 4,952 & 3.59 $\pm$ 2.37 & 17.59 $\pm$ 58.68 & 66.32 $\pm$ 79.56 & 2.26 $\pm$ 0.61 \\
    \texttt{it} & 5,453 & 18,956 & 1,850 & 3.48 $\pm$ 2.35 & 10.39 $\pm$ 27.04 & 66.00 $\pm$ 84.43 & 2.18 $\pm$ 0.46 \\
    \bottomrule
    \end{tabular}
    \caption{Summary statistics of \data{} ground-truth threads, turns, users, and average metrics (mean $\pm$ std).}
    \label{tab:threads_summary}
\end{table*}

\subsection{Dataset Construction}
\label{sec:dataset_construction}

\paragraph{Threads}
The threads contained in Wiki-Talkie are extracted from raw Wikipedia Talk page dumps.\footnote{Dumps from October 2025: \url{https://dumps.wikimedia.org/<lang_code>wiki/20251001/}.}
Wikipedia Talk pages are largely unstructured, making clean conversation extraction a challenging preprocessing task \cite{ferschke-etal-2012-behind, hua-etal-2018-wikiconv}. 
We exploit standard conventions: MediaWiki heading syntax for topics, username-timestamp signatures, and colon-based reply indentation, and handcrafted regular expressions to parse threads into flat sequential turns ordered by timestamps~\footnote{See \url{https://en.wikipedia.org/wiki/Wikipedia:Talk_page_guidelines}.}. 
To address extraction errors, we prompted \extractor{} to revise threads by correcting turn boundaries and comment attributions, flagging irrecoverable cases for exclusion (prompt reported in Appendix \ref{app:prompts}, Figure \ref{fig:prompt-profiling}).

\paragraph{Persona Profiles}
For each unique user in the extracted threads, we fetched the corresponding Wikipedia user page via the MediaWiki API.\footnote{\url{https://www.mediawiki.org/wiki/API}, accessed through \texttt{mwclient}
(\url{https://github.com/mwclient/mwclient}).}
User pages are optional personal spaces where editors describe themselves, their interests, expertise, and contributions. We prompted \extractor{} to determine whether a page contained relevant biographical or self-descriptive content, excluding pages consisting solely of administrative templates or userboxes (prompt in Appendix~\ref{app:prompts}, Figure~\ref{fig:prompt-profile}). For pages with substantive content, the model also generated a natural-language profile summarizing all available details, following recent work on LLM-based persona extraction \citep{zhang-2024-guided, wu-etal-2025-personas, mansour-etal-2025-paars}. Because these are self-declared attributes from real users, the resulting personas avoid the stereotypical or synthetic quality of artificially constructed profiles and naturally incorporate sociodemographic aspects, making them valuable for evaluating how such factors influence agent behavior and biases. Alongside the profile, the model extracted predefined attributes in JSON format (e.g., primary editing topics, gender, profession) to support statistical analysis of user subgroups. Users with successfully generated profiles were subsampled to at most 10,000 per language, prioritizing presence of contribution topics, then gender (favoring women and non-binary users to counteract imbalance), then profession.
Only threads where all participants belong to the selected subset were retained.

\paragraph{Persona Behavioral Traits}
From each user's comments (across up to 50 randomly sampled threads), we extracted behavioral traits using \extractor{}, characterizing each user along eight dimensions including (dis)agreement strategy and editing behavior. Each dimension is filled with closed response options only when sufficient evidence is available, accompanied by the model's reasoning (prompts in Appendix~\ref{app:prompts}, Figures~\ref{fig:prompt-editorial-1}--\ref{fig:prompt-editorial-2}). Unlike self-declared profiles, these traits are inferred from concrete interactions, making them especially useful for evaluating how behavioral conditioning shapes agent output. A manual quality audit of profiles and traits (faithfulness, coherence, privacy preservation) is reported in Appendix~\ref{app:qualitative_audit}.

\paragraph{Articles}
For each thread, the discussed Wikipedia article is retrieved via the MediaWiki API to provide conversational context, using the revision immediately preceding the first comment's timestamp. This temporal alignment ensures the retrieved content reflects the article state when the discussion began.

Finally, all usernames are anonymized via deterministic hashing to prevent re-identification, and thread and article formatting is cleaned and normalized for consistency. Full details are in Appendix~\ref{app:anonymization}.

\section{Experimental Setting}
\label{sec:experimental}

\paragraph{Task}
Wiki-Talkie supports tasks including comment generation, summarization, and quality assessment. Here, we frame the task as \textit{next-turn generation}: given the preceding comments in a thread, an agent is asked to generate the next comment in the role of the real user who authored it. We always condition on at least one preceding comment (i.e., the first turn is never generated 
from scratch), and prior context always consists of ground-truth comments.
This task formulation was chosen deliberately to enable controlled, comparable evaluation across languages and conditioning strategies, resulting in over 100 experimental conditions (5 languages $\times$ 4 models $\times$ 5 conditioning methods).
Although Wiki-Talkie supports evaluation along several dimensions---including coherence with context, factual grounding in the associated article, and fidelity to persona conditioning---here we focus on interaction behavior of agents.

\paragraph{Data}
To keep generation costs manageable, we experimented with a subsample of Wiki-Talkie. We selected threads involving users with at least six comments in the corpus, ensuring a minimum five-comment history prior to the target comment. From these, we sampled up to 5,000 threads per language, maximizing the proportion of comments authored by females up to 50\%. Summary statistics are presented in Appendix \ref{app:data_sample}.

\begin{figure*}[t!]
    \centering
    \includegraphics[width=\textwidth]{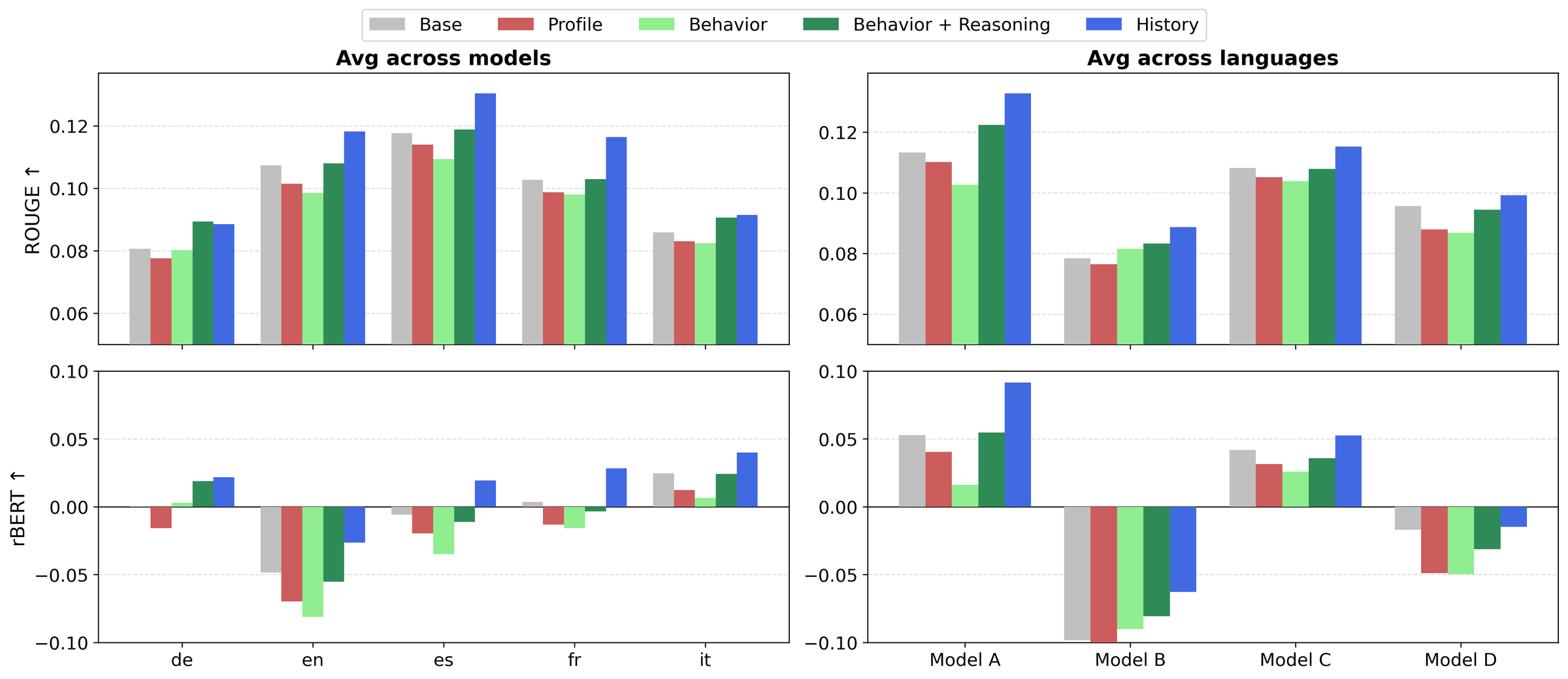}
    \caption{ROUGE-L and rescaled BERTScore between humans and agents across the five 
    conditioning methods.
    \textbf{Left:} averages across models for each language. \textbf{Right:} averages across languages for each model. 
    }
    \label{fig:text_similarity}
\end{figure*}

\paragraph{Conditioning Methods}
We defined five conditioning methods
to investigate how different types of persona information shape agent behavior in interactions:
\textit{i)} \texttt{\textbf{base}} role-play, where the agent is instructed that it is a Wikipedia editor with no additional information; 
\textit{ii)} \texttt{\textbf{profile}}, where the agent is provided with the self-reported profile (gender, age, profession, etc.) of the user who produced the target comment; 
\textit{iii)} \texttt{\textbf{behavior}}, where the agent is provided with the user's behavioral traits; 
\textit{iv)} \texttt{\textbf{behavior+r}}, same as \texttt{\textbf{behavior}} but including the extractor model's reasoning; 
\textit{v)} \texttt{\textbf{history}}, where the agent is provided with 5-20 comments by the same user (randomly sampled if more are available), enabling indirect conditioning through stylistic and behavioral patterns in prior contributions. The prompts for each method are provided in Appendix \ref{app:prompts}, Figures \ref{fig:prompt-base}, \ref{fig:prompt-profile}, \ref{fig:prompt-behavior}, \ref{fig:prompt-history}.

\paragraph{Models}
We run experiments with four state-of-the-art models spanning both open-source and proprietary systems: \modelA{} (proprietary), \modelB{}, \modelC{}, and \modelD{} (open-weights).\footnote{Model identities are withheld; we report results under anonymized labels (\modelA{}--\modelD{}).} All four are widely-used frontier LLMs at the top of their respective open-weights and proprietary classes at the time of writing, chosen to ensure that our findings reflect the behavior of strong, competitive systems rather than artifacts of weak models. Implementation and generation details are provided in Appendix~\ref{app:generation_parameters}.

\paragraph{Evaluation Metrics}
We report \textit{text similarity} via ROUGE-L~\citep{lin-2004-rouge} for surface lexical overlap and BERTScore~\citep{bertscore} for semantic similarity, both raw and rescaled against a baseline (rBERT) as recommended.\footnote{\url{https://github.com/Tiiiger/bert_score}.} Since generation is open-ended, these are coarse reference points rather than primary criteria.

Our primary focus is \textit{interactional behavior}. We extract dialogue acts using the taxonomy of \citet{ferschke-etal-2012-behind}, developed for Wikipedia Talk pages, considering 10 acts: \textsc{critique}, \textsc{request}, \textsc{reference}, \textsc{commit}, \textsc{report}, \textsc{inform}, \textsc{ask}, \textsc{positive}, \textsc{negative}, and \textsc{partial}.\footnote{Mapping to \citet{ferschke-etal-2012-behind} is in Appendix~\ref{app:act_validation}.} Both human and agent comments are annotated with \extractor{} as LLM annotator (prompt in Appendix~\ref{app:prompts}, Figure~\ref{fig:prompt-acts}), which achieves 0.76 F1 against human judgment (Appendix~\ref{app:act_validation}). Each comment can carry one or multiple acts. We additionally assess overall response tone via sentiment prediction on both human and agent comments.\footnote{Using the multilingual model \url{https://huggingface.co/tabularisai/multilingual-sentiment-analysis}, predicting on a five-point scale from \textit{very negative} to \textit{very positive}.}

All metrics are computed at the dataset level. BERTScore and ROUGE-L are averaged over all output-reference pairs. Sentiment and dialogue act distributions are compared via Wasserstein Distance (WD; \citealp{panaretos-2019-wd}) and Jensen-Shannon Divergence (JSD; \citealp{js-divergence}), reflecting the ordinal nature of sentiment and the nominal nature of acts. This aggregate perspective targets \textit{population fidelity}---whether agents collectively reproduce distributional behavioral tendencies---rather than individual-level persona fidelity. 
ROUGE-L and BERTScore are computed at the instance level against each target user's ground-truth response and then averaged across the dataset.

\section{Results}
\label{sec:results}

\subsection{Text Similarity}
\label{sec:results_text_similarity}

Figure~\ref{fig:text_similarity} shows ROUGE-L and rBERT scores averaged across models and languages; full results, including average comment length and raw BERTScore (which remains stable across setups, ${\sim}0.84$--$0.85$, and is thus less informative), are in Appendix~\ref{app:full_results}, Tables~\ref{tab:full_results_overall_1}--\ref{tab:full_results_overall_2}. While our primary target is population fidelity (Section~\ref{sec:experimental}), these instance-level metrics provide a complementary individual-level signal.

ROUGE-L scores never exceed $0.154$. While low in absolute terms, this is expected given the open-ended nature of the task, where the space of plausible responses is virtually unbounded; scores are thus best read as relative measures. rBERT hovers around zero, indicating that agent-human similarity is broadly comparable to random sentence-human pairs, further confirming \textbf{the difficulty of the task}.

\begin{figure*}[t!]
    \centering
    \includegraphics[width=\textwidth]{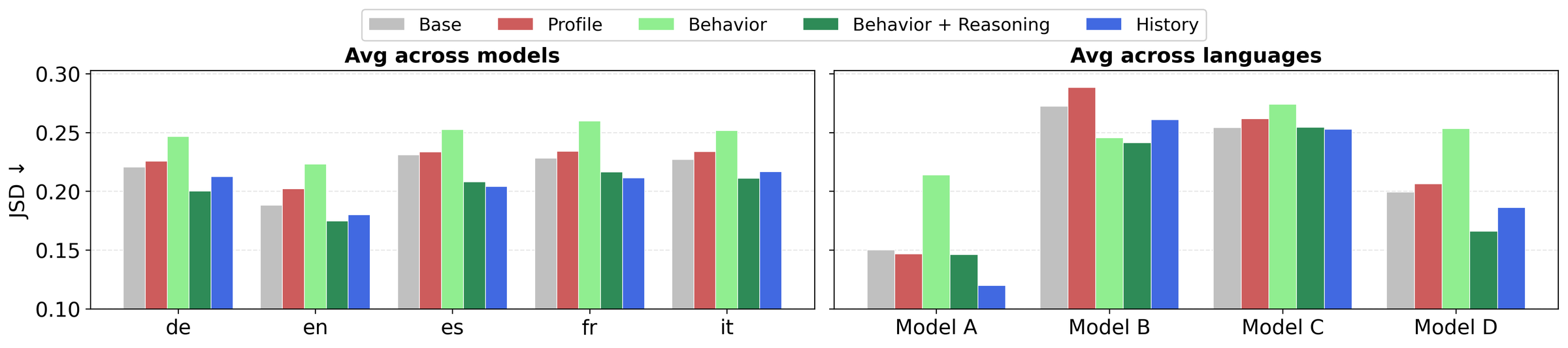}
    \caption{JSD between human and agent dialog act distributions across conditioning methods. \textbf{Left:} averages across models for each language. \textbf{Right:} averages across languages for each model.}
    \label{fig:acts}
\end{figure*}

Across both languages and models, \history{} consistently yields the highest similarity, with ROUGE-L up to $0.154$ (\modelA{} in \texttt{es}) and rBERT up to $0.105$ (\modelA{} in \texttt{de}). \behaviorreasoning{} is the second most effective, with consistent gains over \base{} (e.g., ROUGE-L up to $0.138$ for \modelA{} in \texttt{es}; rBERT up to $0.085$ for \modelA{} in \texttt{de}). By contrast, \profile{} generally degrades similarity relative to \base{} (e.g., rBERT drops on average from $0.053$ to $0.040$ for \modelA{}, and from $-0.017$ to $-0.049$ for \modelD{}), likely because prompting agents to reflect persona traits leads them to incorporate non-relevant details, producing longer and more elaborated responses without improving content alignment. Indeed, agent comments are consistently longer than human comments across all conditions (Appendix~\ref{app:full_results}). \behavior{} similarly fails to improve similarity for most models and languages, with only modest gains in some cases (e.g., \modelB{}'s ROUGE-L from $0.078$ to $0.082$).

Cross-lingually, ROUGE-L scores are consistently higher for \texttt{es} and \texttt{en} than for \texttt{de} and \texttt{it}, partially mirroring human comment length: longer reference texts provide richer surface for lexical overlap. rBERT trends are less pronounced; \texttt{en} falls slightly below the others, largely driven by \modelB{}'s particularly low English scores.

Overall, \textbf{concrete demonstrations}---through comment histories (\history{}) or behavioral traits paired with reasoning (\behaviorreasoning{})---\textbf{are more effective than abstract descriptions} (\behavior{}) \textbf{and sociodemographic information} (\profile{}) at steering agent output toward human-like text. This is the dominant pattern across models and languages, while \textbf{cross-lingual differences remain small}, though we note that all five languages are high-resource and European. The consistency is nonetheless non-trivial given that they span two language families and differ in Wikipedia community size and editorial norms.

\subsection{Interaction Behavior}
\label{sec:results_interaction_behavior}

\begin{figure*}[t!]
    \centering
    \includegraphics[width=\textwidth]{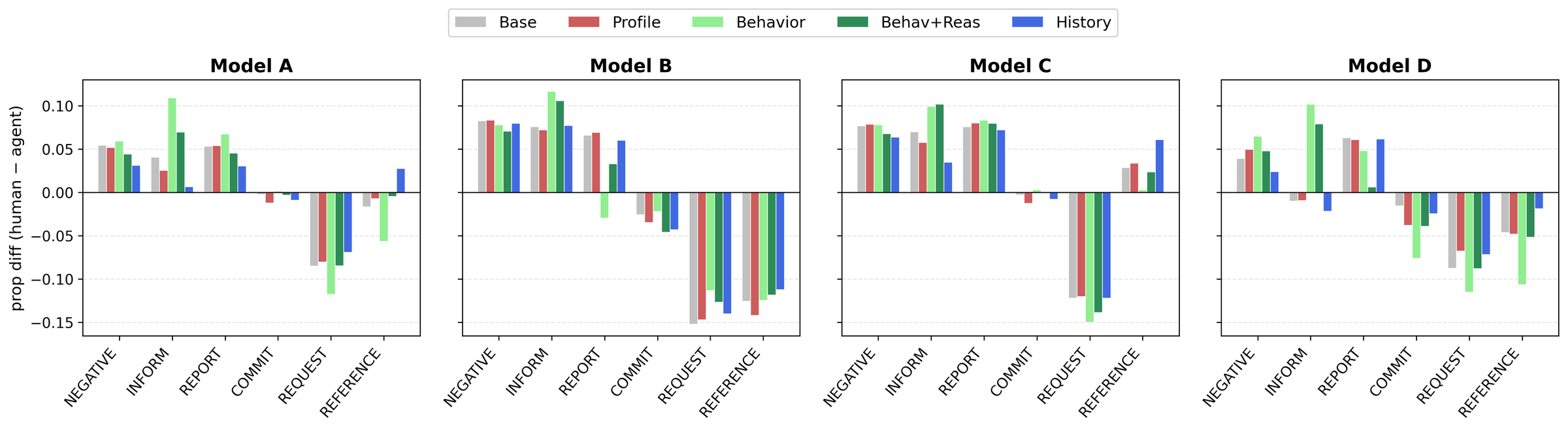}
    \caption{Proportion difference between human and agent dialog acts. Positive values indicate human over-production; negative values indicate agent over-production. Differences are averaged across languages.}
    \label{fig:gender_gap_per_label}
\end{figure*}

\begin{figure}[t!]
    \centering
    \includegraphics[width=0.45\textwidth]{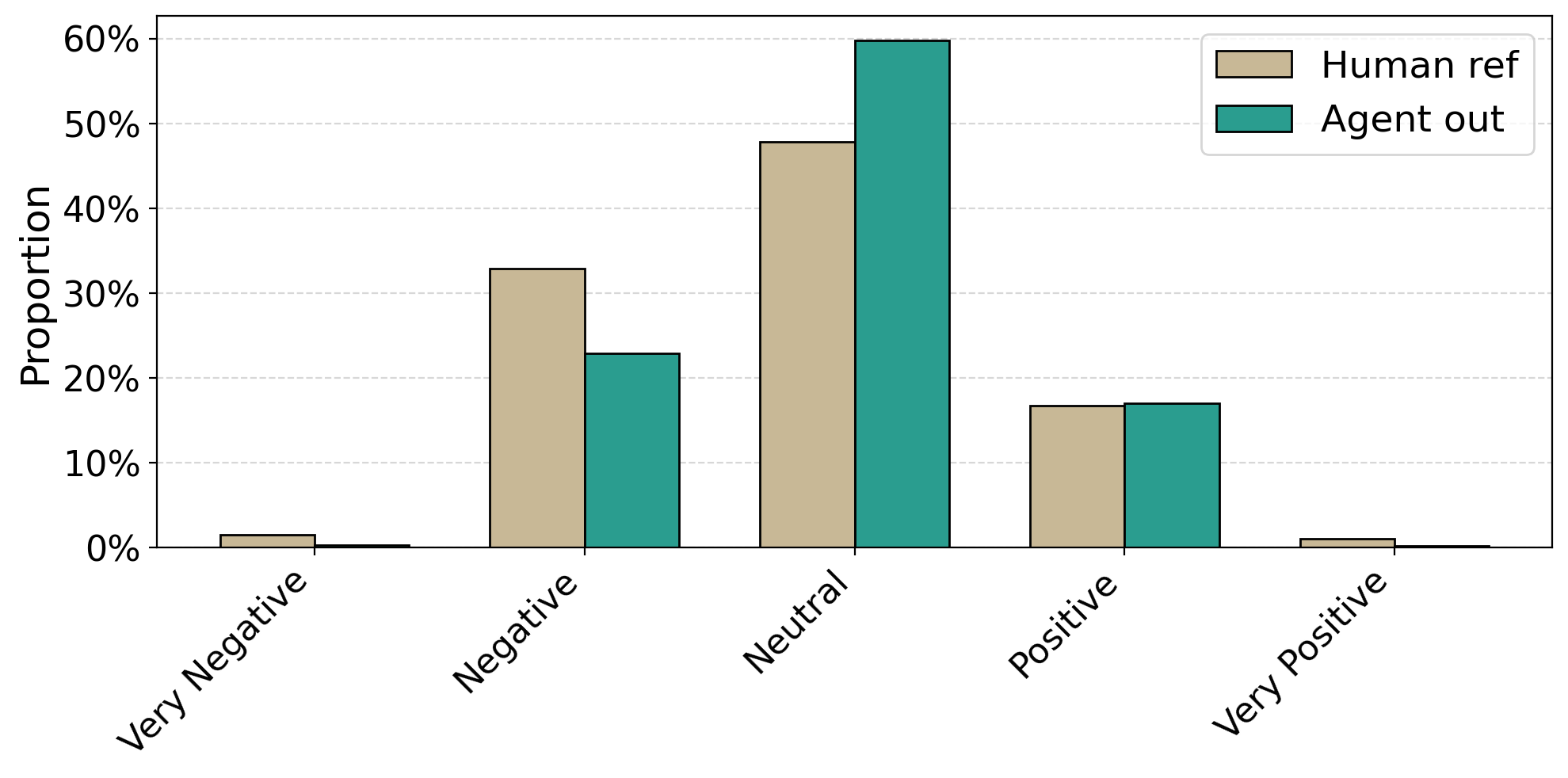}
    \caption{Distribution of sentiment scores for human and agent (\modelA{} under \history{}) comments for \texttt{en}.}
    \label{fig:sentiment}
\end{figure}

In the following we discuss evaluation of behavioral patterns through dialog act JSD and sentiment WD; detailed scores are reported in Appendix~\ref{app:full_results}, Tables~\ref{tab:acts_overall_1} and~\ref{tab:acts_overall_2}.

\paragraph{Acts JSD.}
Figure~\ref{fig:acts} presents average JSD scores aggregated across models (left) and languages (right). Scores range between $0.1$ and $0.3$, indicating only \textbf{moderate divergence between human and agent dialogue act distributions}. \history{} is the strongest method, achieving the lowest JSD overall ($0.085$ for \modelA{} in \texttt{en}). \behaviorreasoning{} is second, with consistent gains over \base{} (e.g., \modelB{}: $0.272 \to 0.241$; \modelD{}: $0.199 \to 0.166$). By contrast, \behavior{} is the weakest, increasing JSD across most models and languages (except \modelB{}, where it improves over \base{}, $0.272 \to 0.245$). \profile{} yields only marginal effects, with gains limited to \modelA{} ($0.150 \to 0.147$). These trends are broadly consistent across languages, though \texttt{en} achieves the lowest average JSD throughout (e.g., $0.188$ under \base{}, $0.175$ under \behaviorreasoning{}), likely reflecting English-centric pretraining. \textbf{Cross-lingual differences, however, remain marginal}.

Overall, these results reinforce the pattern observed for text similarity: \textbf{concrete behavioral demonstrations} (\history{}, \behaviorreasoning{}) \textbf{are consistently more effective than abstract descriptions or sociodemographic factors} (\behavior{}, \profile{}) at aligning agent behavior with human distributions, echoing the well-established advantage of few-shot over zero-shot conditioning~\cite{brown_etal_2020_fewshot}.

One might wonder whether this gap reflects a difference in information \emph{type} or simply in information \emph{quantity}, since \history{} provides 5--20 raw comments while \profile{} and \behavior{} supply compressed summaries. We address this concern in Appendix~\ref{app:profile_length}, where we show that performance under \profile{} does not improve consistently with profile length: ROUGE-L and rBERT are flat, and where acts and sentiment JSD change, they mostly worsen with longer profiles. Combined with the observation that \profile{} \emph{degrades} performance relative to \base{} despite adding tokens, this argues against a pure context-budget explanation and supports our interpretation that the operative factor is information type, not quantity.

\paragraph{Per-act differences}
To assess whether divergences concentrate on specific labels, we analyze per-label proportion differences ($\delta$) between human and agent distributions, where positive values indicate human over-production and negative values indicate agent over-production. Figure~\ref{fig:gender_gap_per_label} shows differences averaged across languages, omitting acts with negligible variation (\textsc{Ask}, \textsc{Critique}, \textsc{Positive}, \textsc{Partial}). Differences are computed between proportions rather than raw counts to account for lower act density in agent comments (Appendix~\ref{app:full_results}). Full breakdowns are in Tables~\ref{tab:acts_overall_1}--\ref{tab:acts_overall_2}.

Agents systematically over-produce \textsc{Request} (up to $\delta = -0.177$ for \modelC{} in \texttt{de} on \behavior{}) and \textsc{Commit} (up to $\delta = -0.108$ for \modelD{} in \texttt{fr} on \behavior{}), while humans produce more \textsc{Report} ($\delta = 0.088$ for \modelC{} in \texttt{fr} on \behavior{}). Together, this suggests that agents tend to \emph{operate} by requesting and committing to changes, yet report fewer actions already undertaken. \textsc{Inform} is under-produced by agents across most models, possibly reflecting insufficient factual grounding (\modelA{} and \modelD{} diverge in some methods). \textsc{Reference} shows a model-dependent pattern. \textsc{Negative} is consistently under-produced ($\delta$ from $0.020$ to $0.104$), consistent with sycophantic tendencies of LLMs~\cite{malmqvist_etal_2025_sycophancy}. Overall, the most systematic biases concern \emph{operativity} (over-production of \textsc{Request}, \textsc{Commit}) and \emph{negativity} (under-production of \textsc{Negative}, \textsc{Report}), \textbf{pointing to a consistent tendency for agents to be more action-oriented and less confrontational than humans.}

\paragraph{Per-act differences by gender}
To examine whether agents reproduce dialog act trends \emph{equitably} across genders, we analyze per-label proportion differences split by gender for two models with contrasting alignment levels (\modelA{}, higher; \modelB{}, lower) and two languages with similar contrast (English, higher; French, lower), under \history{} (best overall), \profile{} (the only method with explicit gender), and \base{} as reference. Differences are shown in Appendix~\ref{app:full_results}, Figure~\ref{fig:gender_acts}. Trends are consistent across genders for all labels and methods. Crucially, \profile{} does not behave systematically differently from \base{} or \history{}, indicating that \textbf{explicit gender conditioning neither amplifies nor reduces gender disparities in dialog act profiles}.

\paragraph{Sentiment}
Sentiment alignment broadly mirrors the dialog act trends: \history{} yields the best alignment (minimum WD $0.097$ for \modelC{} in \texttt{en}), followed by \behaviorreasoning{} ($0.168$ in \texttt{en}). Across languages, \texttt{en} shows the lowest WD ($0.315$ on average under \base{}) and \texttt{es} the highest ($0.452$), consistently across methods. As shown in Figure~\ref{fig:sentiment} for \modelA{} under \history{}, agent responses rarely exhibit sentiment extremes (\textit{very positive} or \textit{very negative}) and contain fewer \textit{negative} comments than humans, consistent with their lower rate of \textsc{Negative} acts. These patterns hold across models, conditioning methods, and languages (Appendix~\ref{app:full_results}, Figure~\ref{fig:sentiment_distributions}). Overall, sentiment converges with the broader finding that \textbf{agents tend to avoid negative or confrontational content across all conditioning methods}.

\section{Conclusions}

We introduced \data, a multilingual dataset derived from Wikipedia Talk pages to evaluate agent interactions across five languages, using four state-of-the-art LLMs under different persona conditioning strategies.
Our evaluation targets population fidelity---whether agents collectively reproduce the distributional behavioral patterns of human communities---complemented by instance-level text similarity metrics.
Our experiments show that grounding agents with concrete comment examples consistently outperforms abstract persona descriptions across all languages. Notably, sociodemographic information can degrade agent performance, an encouraging finding for privacy-preserving applications. In addition, agents are consistently more action-oriented and less confrontational than humans, revealing a bias toward agreeableness and avoidance of extremes. Cross-lingual differences remain small, with English showing only slightly greater alignment.
Overall, these results underscore both the promise and limitations of persona-conditioned generation, motivating further research into more effective simulation strategies.
While we focused on next-turn generation to enable controlled comparison across a large experimental space, Wiki-Talkie is designed to support richer evaluation settings, including multi-turn generation, summarization, factual grounding assessment, and quality evaluation, all of which we encourage the community to explore.

\section*{Limitations}
\label{sec:limitations}

\paragraph{Dataset Construction}
Despite our multi-stage extraction and revision pipeline, some residual noise remains in Wiki-Talkie, including occasional misattributed comments, imperfect comment boundaries, and formatting artifacts not fully resolved during preprocessing 
Such noise is inevitable at this scale and is unlikely to substantially affect evaluation or analysis.

\paragraph{Language Coverage}
\data{} currently covers five European languages (English, German, Spanish, French, and Italian), limiting its cultural and linguistic diversity. While these languages span two distinct families (Romance and Germanic) and vary in Wikipedia community size and editorial norms, they are all high-resource; findings may not generalize to language communities with different writing systems, discourse conventions, or editorial practices. Future work should extend \data{} to additional languages, particularly those underrepresented in NLP research. The dataset construction pipeline and evaluation framework are fully language-agnostic and straightforwardly extensible to lower-resource and non-European languages.

\paragraph{Dataset Richness}
The richness and consistency of user personas vary considerably within and across languages. Some profiles contain detailed self-reported attributes, while others are sparse. English is generally better represented in both quantity and quality, meaning persona conditioning experiments may be more reliable for English than for other languages.
Moreover, despite subsampling to increase the representation of women and non-binary users, the dataset remains gender-skewed, which may affect the robustness of gender-based analyses.

\paragraph{Persona Conditioning}
Our experiments focus exclusively on prompting-based persona conditioning, leaving alternative approaches to future work, such as fine-tuning (e.g., \citealp{thakur2024personas, takayama-etal-2025-persona}), retrieval-augmented conditioning (e.g., \citealp{huang-etal-2023-retrieval, salemi-etal-2024-lamp}), or embedding-based persona representations (e.g, \citealp{liu-etal-2025-llms}).

\paragraph{Information Density across Conditioning Methods}
Our conditioning methods are not strictly matched on information density: \history{} provides 5--20 raw comments, while \profile{} and \behavior{} supply compressed summaries. We probe this confound in Appendix~\ref{app:profile_length} by stratifying \profile{} performance by profile length and find no consistent improvement with profile length across four metrics, supporting an information-type rather than information-quantity interpretation. We did not perform the analogous analysis for \history{} length because the conditioning pool is concentrated in the 16--20 comment range, leaving lower-comment buckets too sparse to support a meaningful comparison. A fully matched ablation---e.g., comparing \history{} against equally long synthetic summaries distilled from the same comments---remains a direction for future work.

\paragraph{Scope of Persona Representations}
Our profiles and behavioral traits are produced by a single anonymized extractor model and a specific prompting design. A manual audit (Appendix~\ref{app:qualitative_audit}) indicates that these representations are largely faithful and well-grounded, with imprecisions limited to isolated factual details. Nonetheless, our findings about \profile{} and \behavior{} should be read as conclusions about sociodemographic and behavioral descriptors \emph{as we generated them}, rather than as universal claims about all possible persona representations. Alternative summarization strategies, richer attribute schemas, or different extractor models could in principle yield different results.

\paragraph{Evaluation Scope}
Our primary evaluation target is \textit{population fidelity}---whether agents collectively reproduce the distributional behavioral patterns observed in human discussions---rather than \textit{individual-level persona fidelity}, i.e., whether each generated response faithfully reflects the specific persona provided. This choice is principled: individual-level fidelity risks enabling faithful simulation of specific individuals, raising ethical concerns around impersonation and lack of consent \citep{desai-etal-2025-personas-evolved}, while population fidelity is more tractable and more relevant for applications such as community simulation and policy analysis, in line with recent work advocating population-level evaluation of persona-based agents \citep{mansour-etal-2025-paars, fawaz-etal-2026-narriva, rupprecht-etal-2025-german-personas}. 
Assessing fine-grained individual-level persona consistency remains a direction for future work.

\section*{Ethic Statement}

\subsection*{Privacy, Data Protection, and Contextual Integrity}

Wiki-Talkie is derived from Wikipedia dumps, released under a Creative Commons Attribution-ShareAlike 4.0 license. While publicly accessible, we recognize that contributors may reasonably expect their comments to be archived for research purposes but not necessarily used for persona-conditioned simulation. Accordingly, we treat this dataset as human-derived data requiring safeguards beyond copyright compliance, guided by the principles of data minimization, proportionality, contextual integrity, and harm reduction.

\paragraph{Dataset Construction and Anonymization}
The dataset does not contain original user pages verbatim. Instead, we generated neutral, natural-language re-elaborations summarizing publicly stated interests and interactional patterns, deliberately omitting usernames, real names, contact information, and external links. This process inherently introduces a layer of abstraction: the resulting profiles are approximate characterizations that may diverge from the original content through omission, rephrasing, or occasional inaccuracy. While this means profiles should not be treated as ground-truth descriptions of actual persons, we acknowledge that the same process may occasionally introduce erroneous attributes, including sensitive ones. We sought to mitigate this risk through manual inspection of a random sample of outputs (see Appendix \ref{app:qualitative_audit}).

Usernames are replaced using salted cryptographic hashing, where the salt is stored separately from the dataset and is accessible only to the principal investigator, preventing straightforward reversal or precomputation attacks (see Appendix \ref{app:anonymization}). Within comment text, we also anonymized mentions of third-party usernames to the extent possible. However, a residual risk remains where usernames appear in non-standard forms or where real names are mentioned in conversation.

\paragraph{Re-Identification Risk and Access Control}
Complete anonymity cannot be guaranteed when working with public textual data. While salted hashing prevents trivial username reconstruction, a content-based re-identification risk exists: because Wikipedia discussions are publicly archived and indexed, motivated individuals could potentially identify original authors via distinctive phrases or search queries. Our dataset does not introduce new exposure beyond what is already publicly accessible, but we recognize that aggregation and structuring may lower the cost of such searches.

To mitigate these risks, the dataset is released under the CC BY-SA 4.0 license, inherited from the underlying Wikipedia-derived content. Rather than hosting an open public mirror, we distribute it through a lightweight access request in which prospective users identify themselves and their intended use. This provides an audit trail and a point of contact while raising the effort required for indiscriminate bulk access. As a matter of research norms rather than a binding condition, we ask that recipients not attempt to de-anonymize contributors, not use the data to impersonate or manipulate editorial processes, and work from the most recent release.

We provide a dedicated contact address for individuals who wish to request removal of their data. Upon verification, corresponding entries are added to a blocklist and removed in subsequent releases. Our goal is risk mitigation rather than elimination, and we consider the residual risk proportionate to the scientific value of studying collaborative online deliberation at scale.

\paragraph{Informed Consent}
Individual informed consent was not sought, as contributions to Wikipedia are made under the CC BY-SA license and the Wikimedia Foundation's Terms of Use, which explicitly permit reuse and derivative works. This is consistent with established practice in NLP research on Wikipedia-derived data \citep{ferschke-etal-2012-behind, hua-etal-2018-wikiconv}, and the scale of contributors makes individual consent infeasible.
Nevertheless, we recognize that licensing permissibility does not fully resolve the ethical tension between contributors' original communicative intent and downstream research applications; we address this gap through the layered safeguards described above.

\paragraph{Harmful or Sensitive Content}
Wikipedia Talk pages may occasionally contain harassment, conflictual exchanges, or incidental personal disclosures. We do not seek to amplify or foreground such content; the dataset is designed to analyze aggregate interaction patterns in collaborative environments, not for reputational analysis of specific contributors. We did not apply active content filtering in order to preserve the ecological validity of the discussions. Access restrictions are intended to further prevent misuse or amplification of harmful material.

\subsection*{Experimental Ethics and Persona Conditioning}
The objective of this research is not to recreate, approximate, or impersonate specific individuals. Persona-conditioned agents are stochastic constructs derived from coarse behavioral summaries, used solely to examine population-level dynamics in simulated group discussions. Although individual-level conditioning is methodologically necessary to capture heterogeneity in interaction styles, all evaluation and reporting are conducted at the aggregate level. We do not evaluate model fidelity to specific users, publish individual outputs, or claim to model the beliefs, intentions, or identities of real contributors.

We recognize that repurposing public discourse for AI research raises novel ethical questions concerning digital representation and contextual integrity. Our ethical justification rests on proportionality: the scientific value of understanding large-scale collaborative behavior, combined with layered safeguards (anonymization, profile re-elaboration, controlled access, opt-out mechanisms), outweighs the residual and mitigated risks.

\clearpage

\bibliography{custom}

\appendix

\section{Further \data{} Details}
\label{app:dataset_details}

\subsection{Statistics}
\label{app:statistics}

\begin{figure*}[ht!]
    \centering

    \begin{subfigure}[b]{0.45\textwidth}
        \centering
        \includegraphics[width=\textwidth]{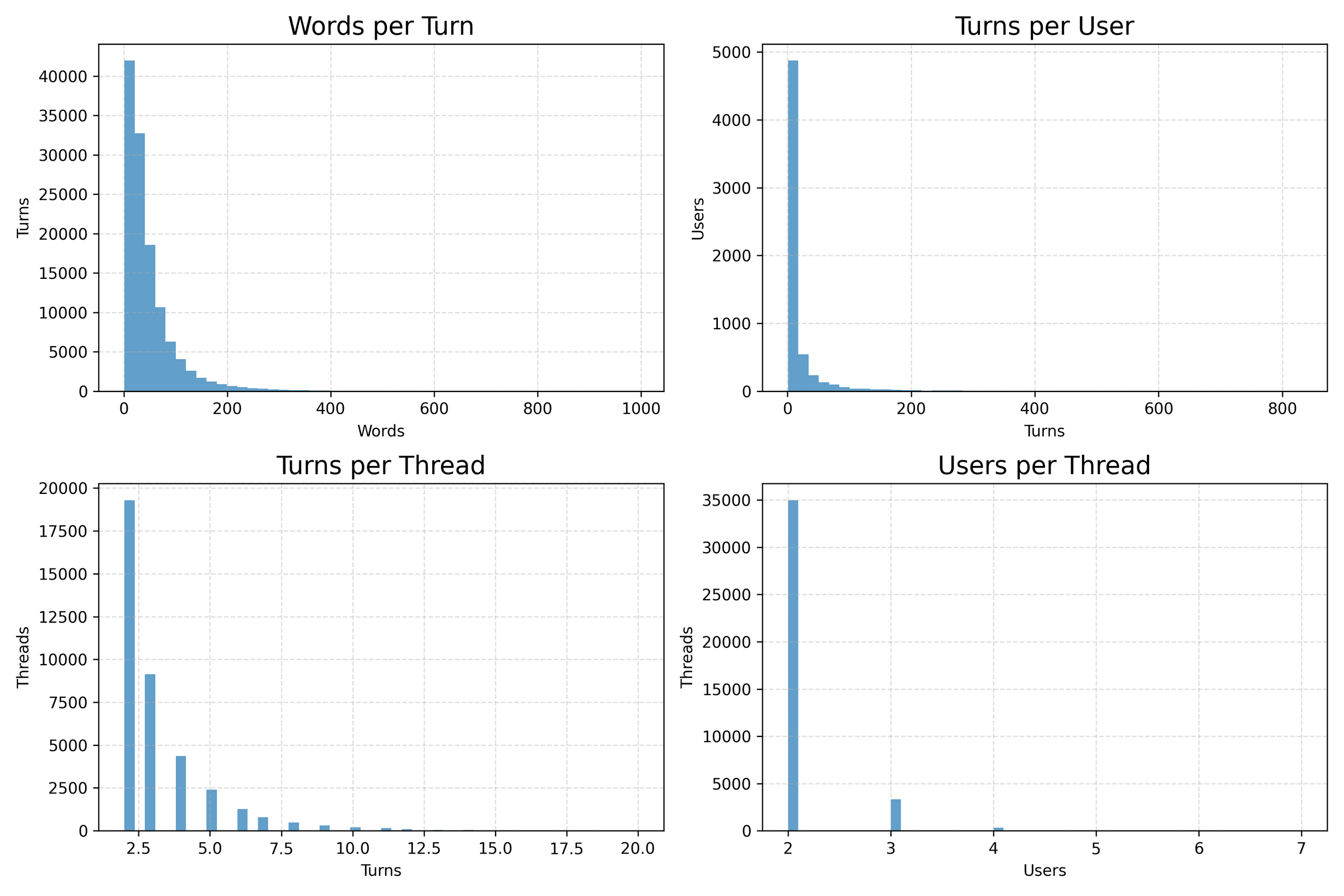}
        \subcaption{\texttt{de}}
    \end{subfigure}
    \hfill
    \begin{subfigure}[b]{0.45\textwidth}
        \centering
        \includegraphics[width=\textwidth]{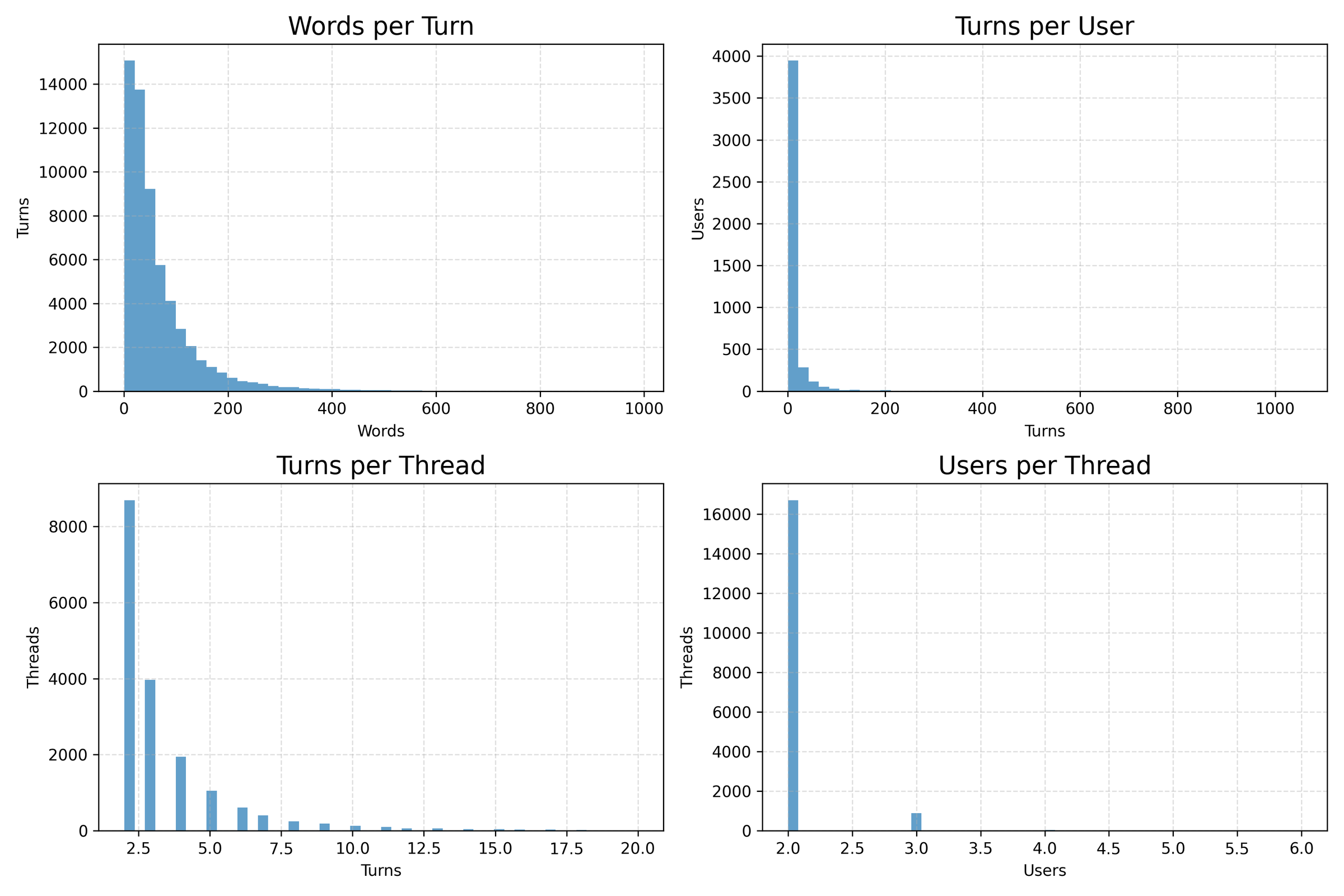}
        \subcaption{\texttt{en}}
    \end{subfigure}

    \par\vspace{1em} 

    \begin{subfigure}[b]{0.45\textwidth}
        \centering
        \includegraphics[width=\textwidth]{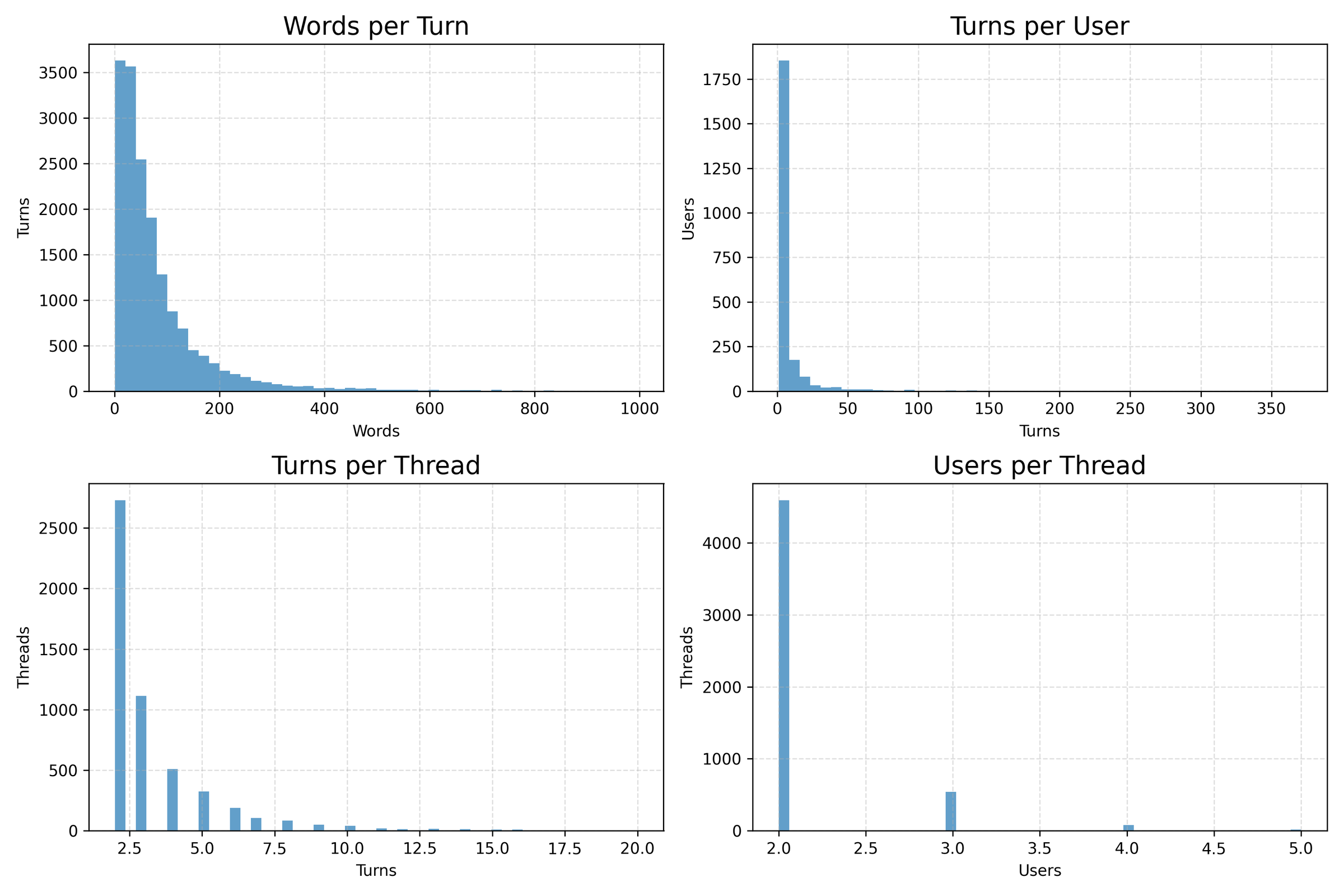}
        \subcaption{\texttt{es}}
    \end{subfigure}
    \hfill
    \begin{subfigure}[b]{0.45\textwidth}
        \centering
        \includegraphics[width=\textwidth]{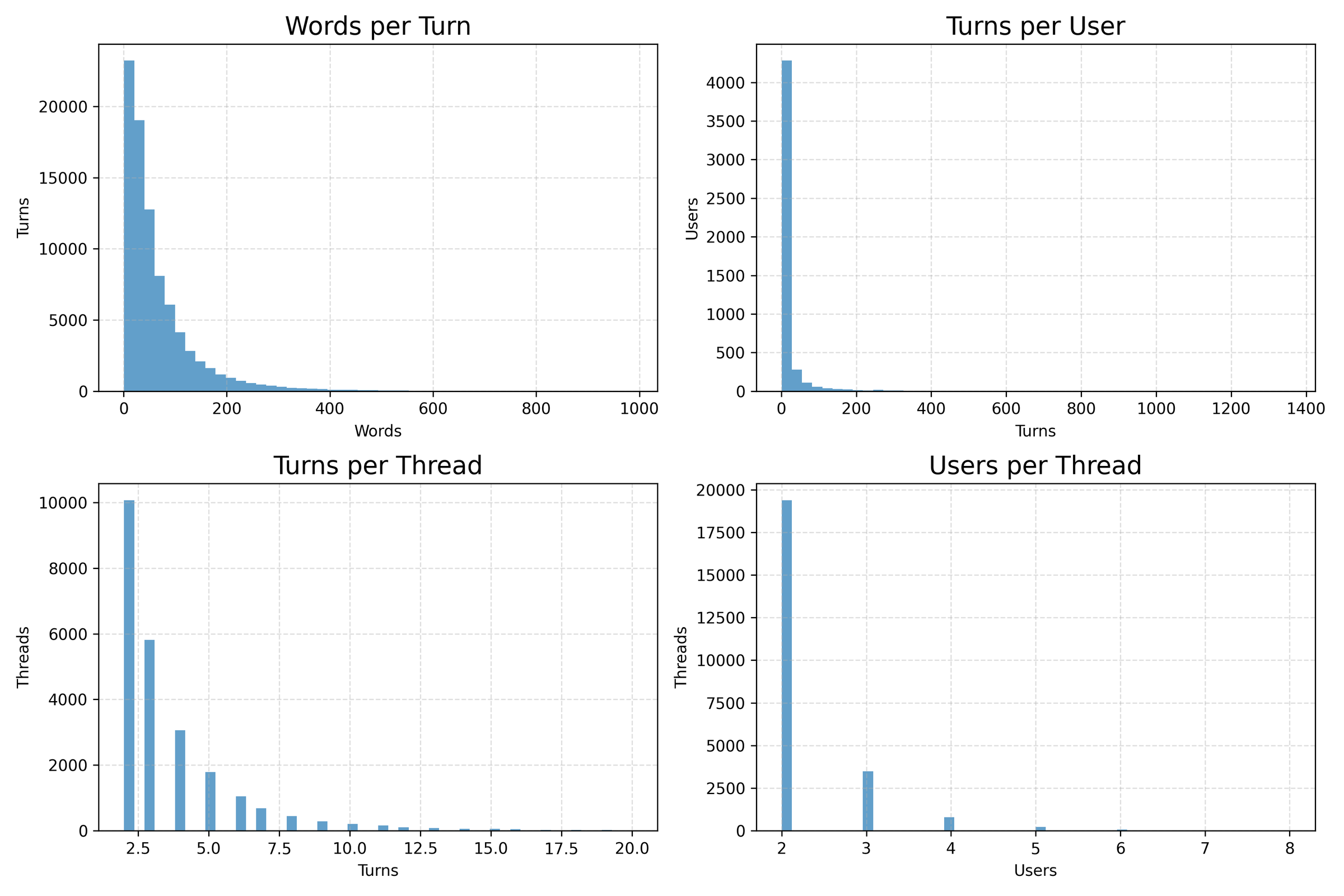}
        \subcaption{\texttt{fr}}
    \end{subfigure}

    \par\vspace{1em} 

    \begin{subfigure}[b]{0.45\textwidth}
        \centering
        \includegraphics[width=\textwidth]{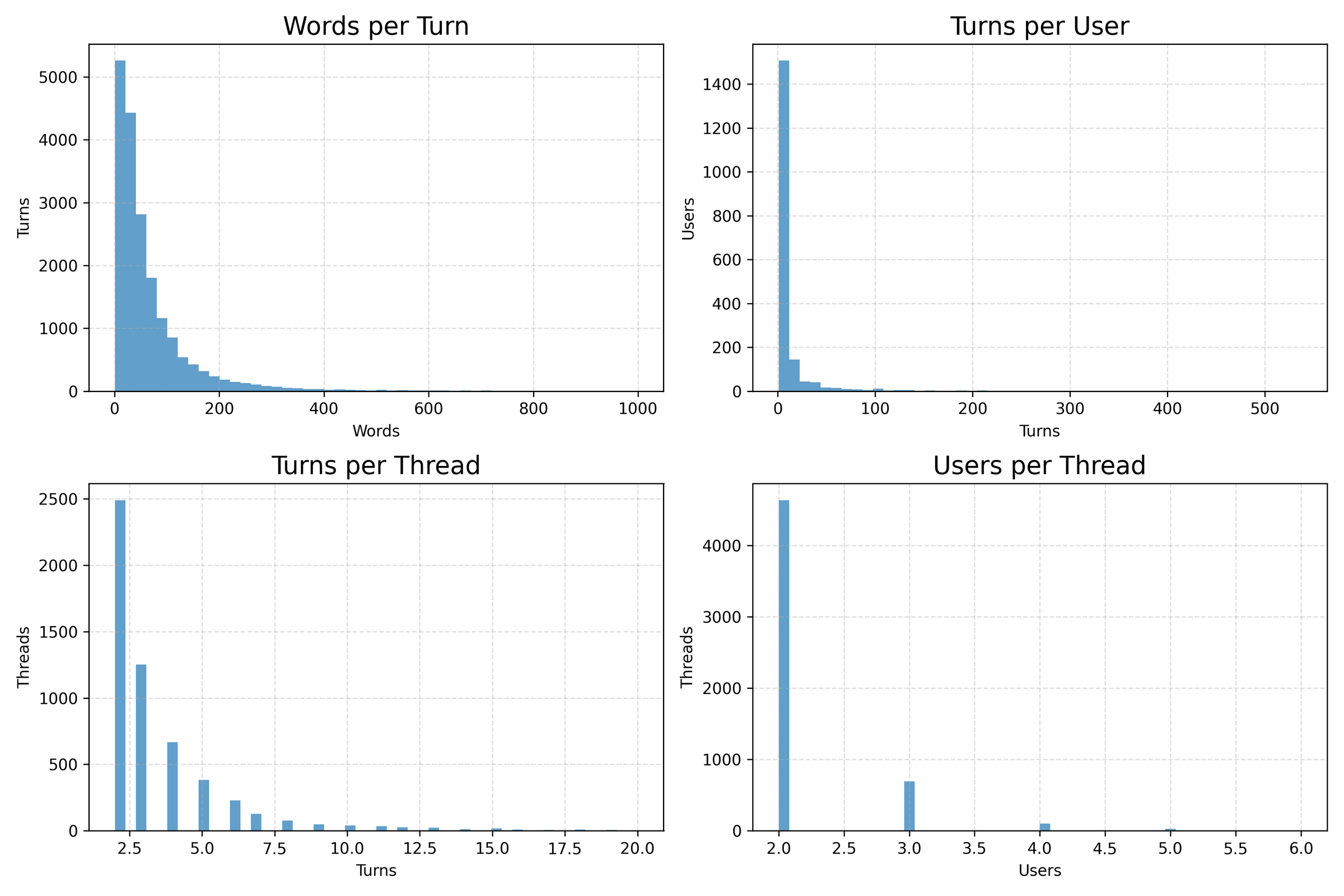}
        \subcaption{\texttt{it}}
    \end{subfigure}

    \caption{Distributions of words per turn, turns per user, turns per thread, users per thread across the five languages.}
    \label{fig:stats_plot}
\end{figure*}

\subsubsection{Distributions}
\label{app:statistics_distributions}

Figure \ref{fig:stats_plot} shows the distributions of the number of words per turn, turns per user, turns per thread, and users per thread across the five languages. These statistical properties exhibit remarkably consistent patterns across languages, suggesting that the conversational dynamics of the dataset are largely language-agnostic.

\paragraph{Words per Turn}
The distribution is heavily right-skewed in every language, with the vast majority of turns containing fewer than 200 words, indicating that most conversational exchanges are brief while a small number of turns involve substantially longer responses. The 1,000-word cap applied uniformly across all languages, as threads containing longer turns were excluded due to their higher likelihood of noise.

\paragraph{Turns per User}
This distribution is similarly right-skewed, with most users contributing only a handful of turns. The bulk of users fall below 50 turns across all languages, with the long tail suggesting a small cohort of highly active participants. The Italian and Spanish subsets exhibit a narrower spread, consistent with the smaller number of threads collected for these languages.

\paragraph{Turns per Thread}
Threads are predominantly short, with the modal thread containing between 2 and 5 turns across all languages. This suggests that most conversations are transactional or goal-oriented rather than extended dialogues, with very few threads exceeding 10 turns.

\paragraph{Users per Thread}
The vast majority of threads involve exactly 2 users, reflecting a typical dyadic conversation structure. Threads with 3 or more participants are rare across all languages, with the distribution dropping off sharply beyond 2 users.

\subsubsection{Users and Personas}
\label{app:statistics_users}

\begin{figure*}[ht!]
    \centering

    \begin{subfigure}[b]{0.45\textwidth}
        \centering
        \includegraphics[width=\textwidth]{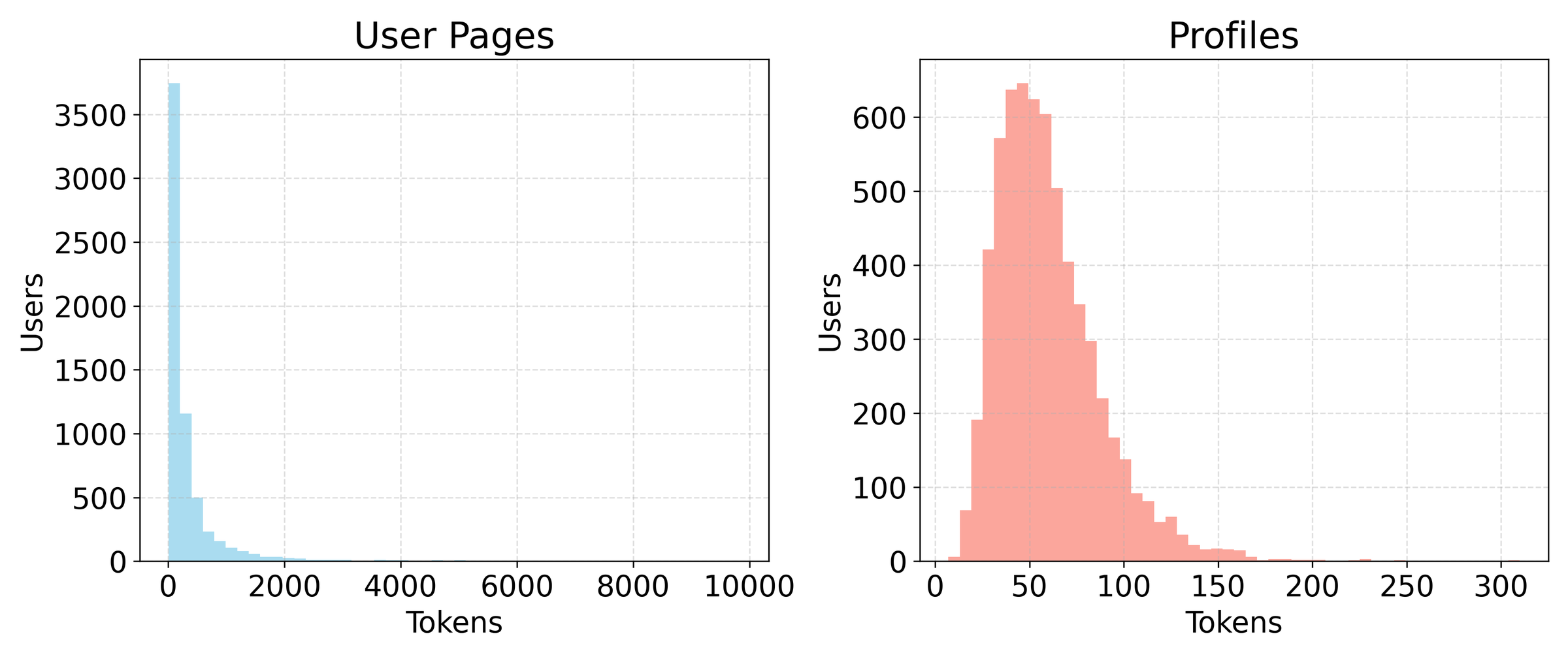}
        \subcaption{\texttt{de}}
    \end{subfigure}
    \hfill
    \begin{subfigure}[b]{0.45\textwidth}
        \centering
        \includegraphics[width=\textwidth]{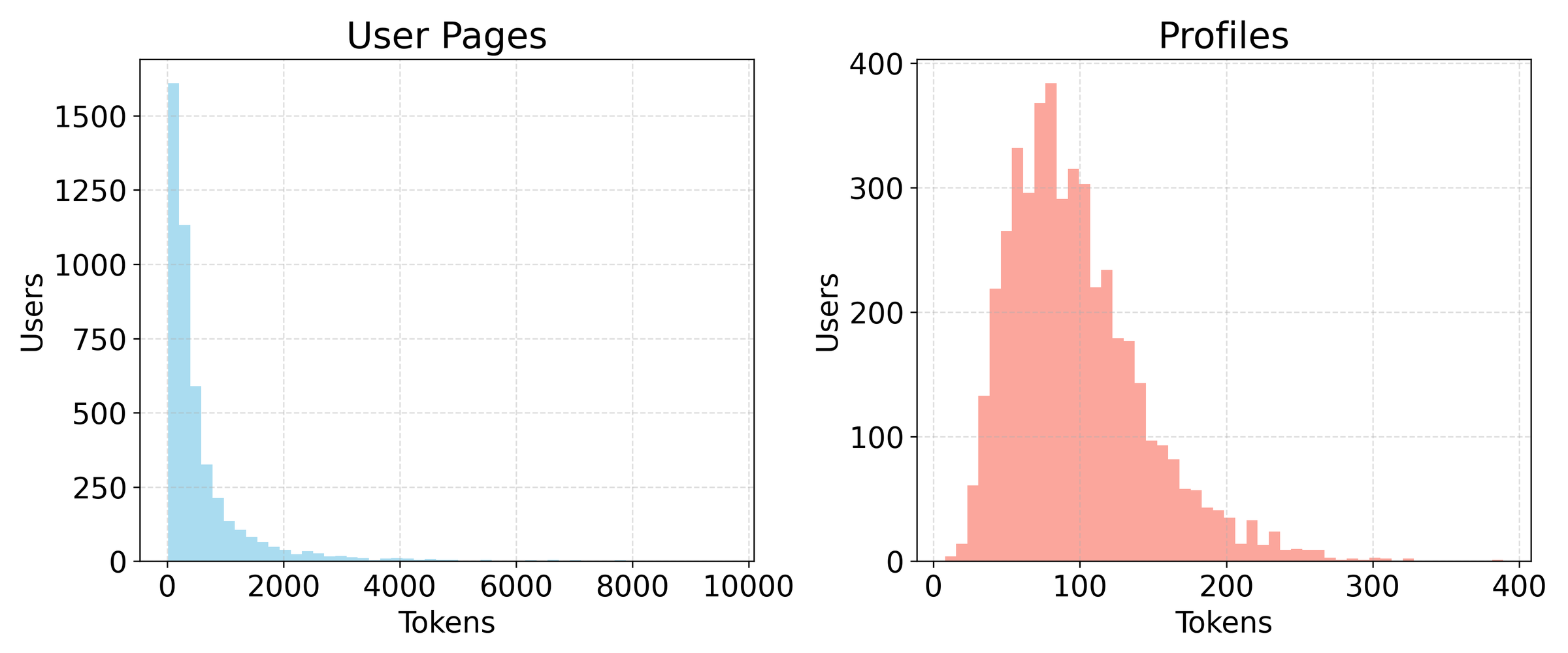}
        \subcaption{\texttt{en}}
    \end{subfigure}

    \par\vspace{1em} 

    \begin{subfigure}[b]{0.45\textwidth}
        \centering
        \includegraphics[width=\textwidth]{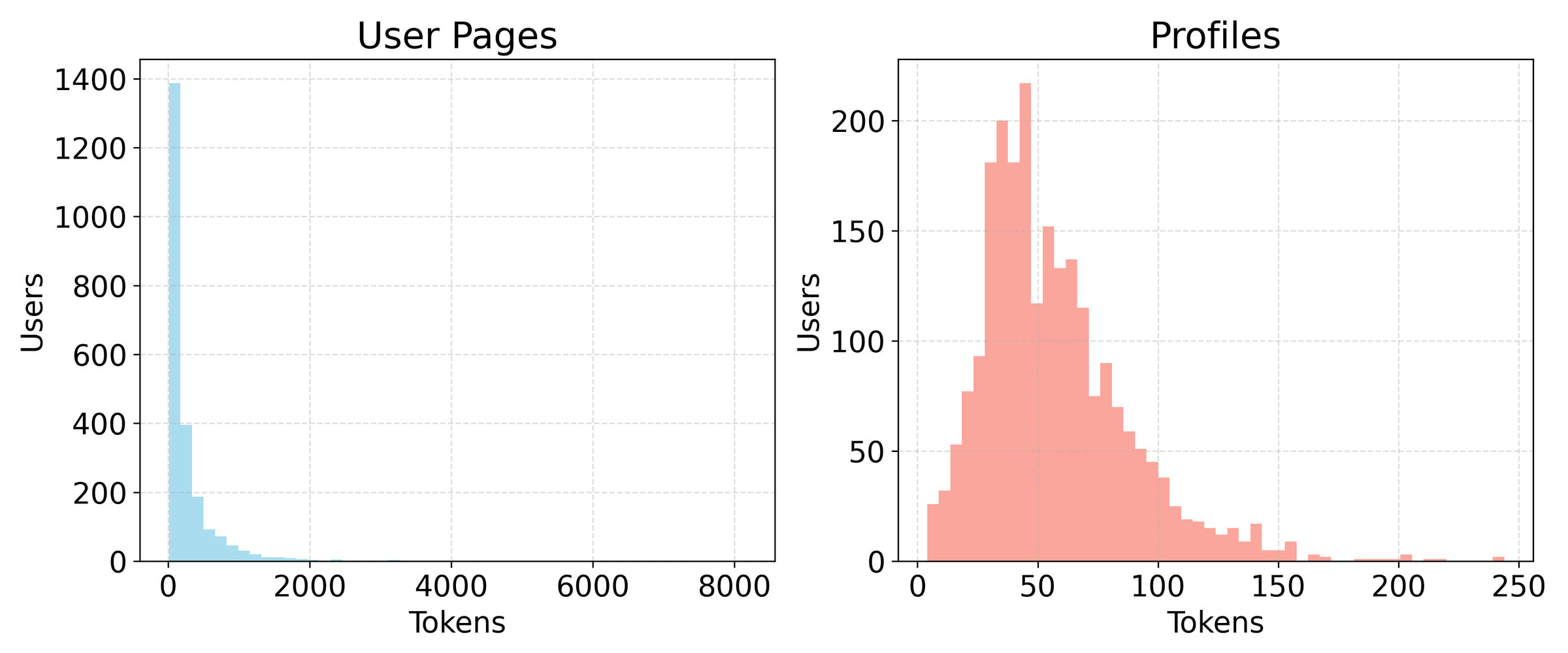}
        \subcaption{\texttt{es}}
    \end{subfigure}
    \hfill
    \begin{subfigure}[b]{0.45\textwidth}
        \centering
        \includegraphics[width=\textwidth]{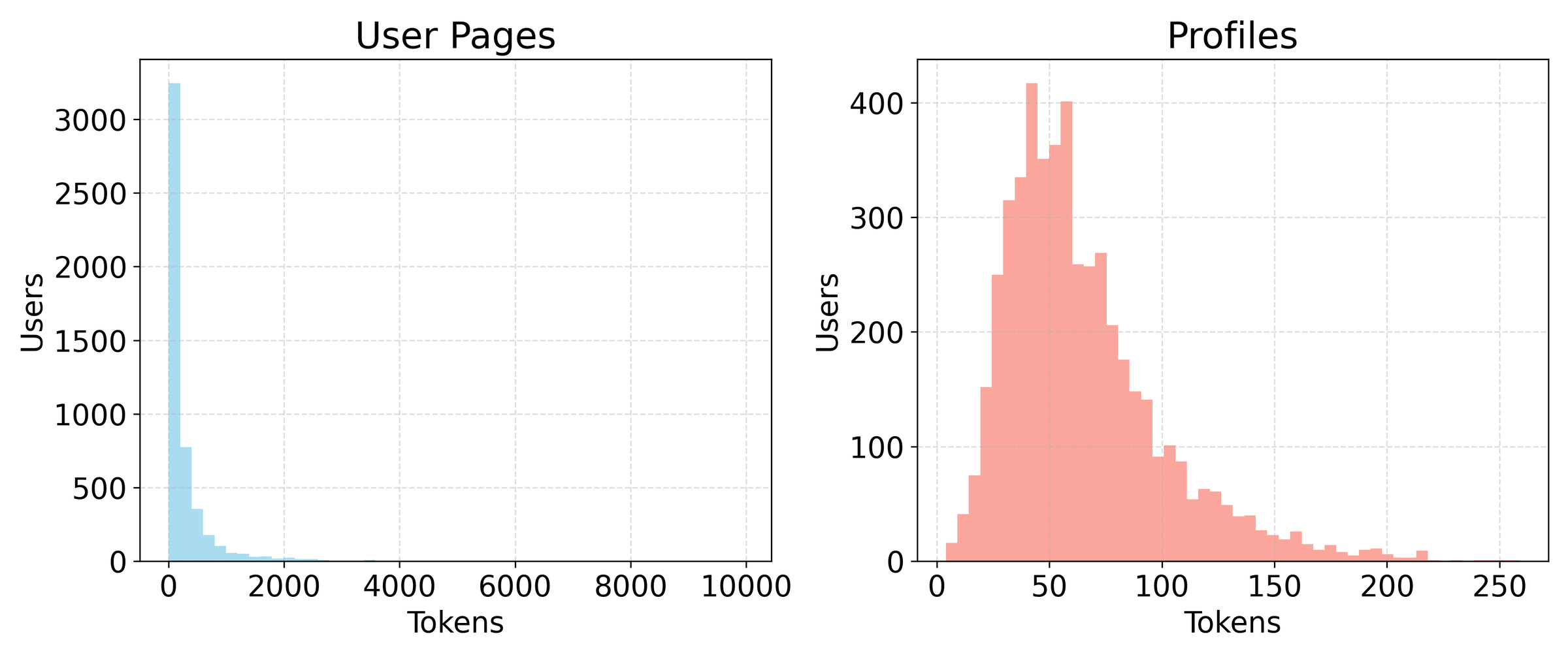}
        \subcaption{\texttt{fr}}
    \end{subfigure}

    \par\vspace{1em} 

    \begin{subfigure}[b]{0.45\textwidth}
        \centering
        \includegraphics[width=\textwidth]{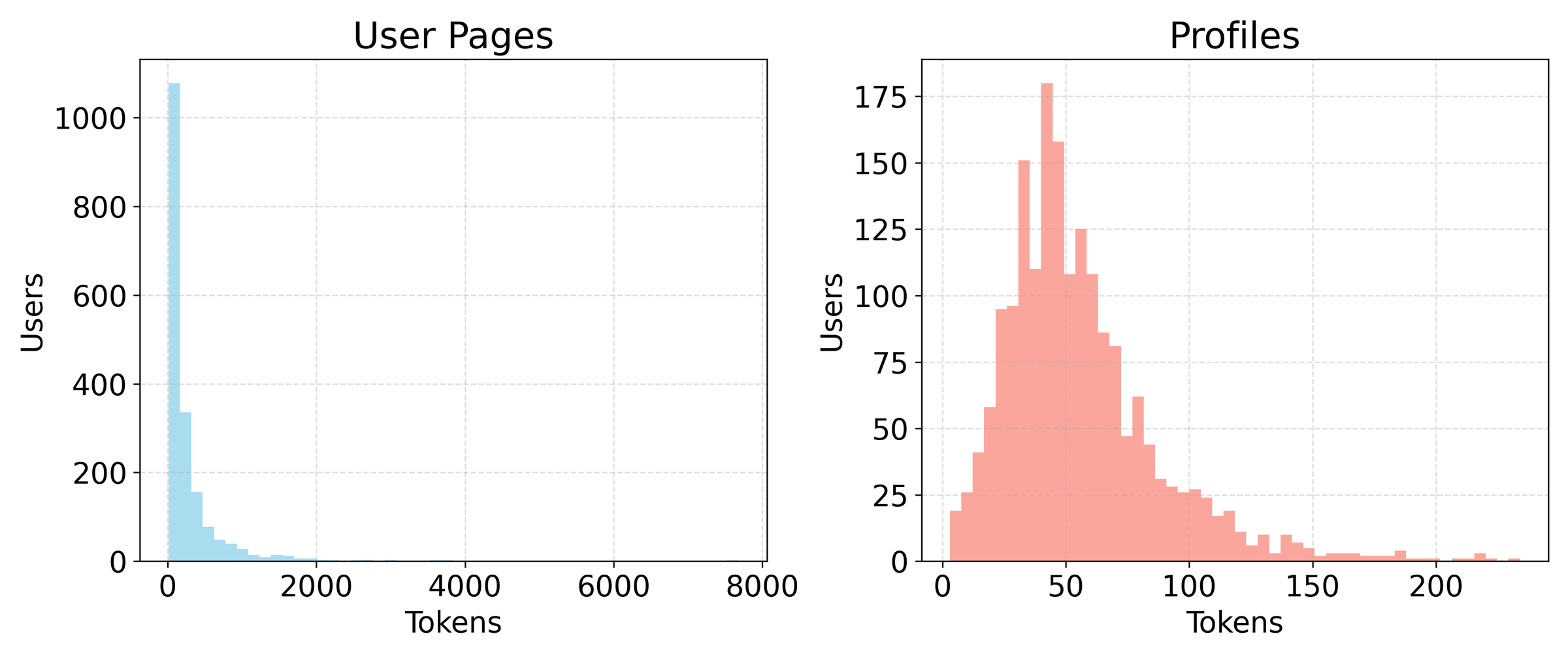}
        \subcaption{\texttt{it}}
    \end{subfigure}

    \caption{Word distributions for user pages and profiles.}
    \label{fig:user_distributions}
\end{figure*}

In Table~\ref{tab:user_stats}, we show statistics on users and personas, with Figure~\ref{fig:user_distributions} showing the corresponding word distributions for user pages and profiles. User pages are right-skewed across all languages, with most users having short pages and a minority producing substantially longer ones. Profiles follow a more bell-shaped distribution centered around 50--100 words, reflecting the relatively standardized length of the persona descriptions we construct from user metadata, and confirming that our persona extraction distills user information into concise, consistent summaries regardless of the underlying page length. 
English users exhibit the richest descriptions, with both the longest average user pages and profile lengths (98.07 words), and near-complete coverage of contribution topics (100\%), gender (67\%), and profession (66\%), reflecting the larger and more active English Wikipedia community. German users, while numerous, show lower rates of gender (36\%) and profession (26\%) disclosure despite high topic coverage (98.9\%). Spanish, French, and Italian users display more modest attribute coverage, particularly for profession, reflecting sparser self-descriptions in these communities. The column reporting users with at least five turns indicates a core of highly active contributors in each language, ranging from approximately 30\% (Spanish, Italian) to nearly 50\% (German).

\begin{figure*}[t!]
    \centering
    \includegraphics[width=0.8\textwidth]{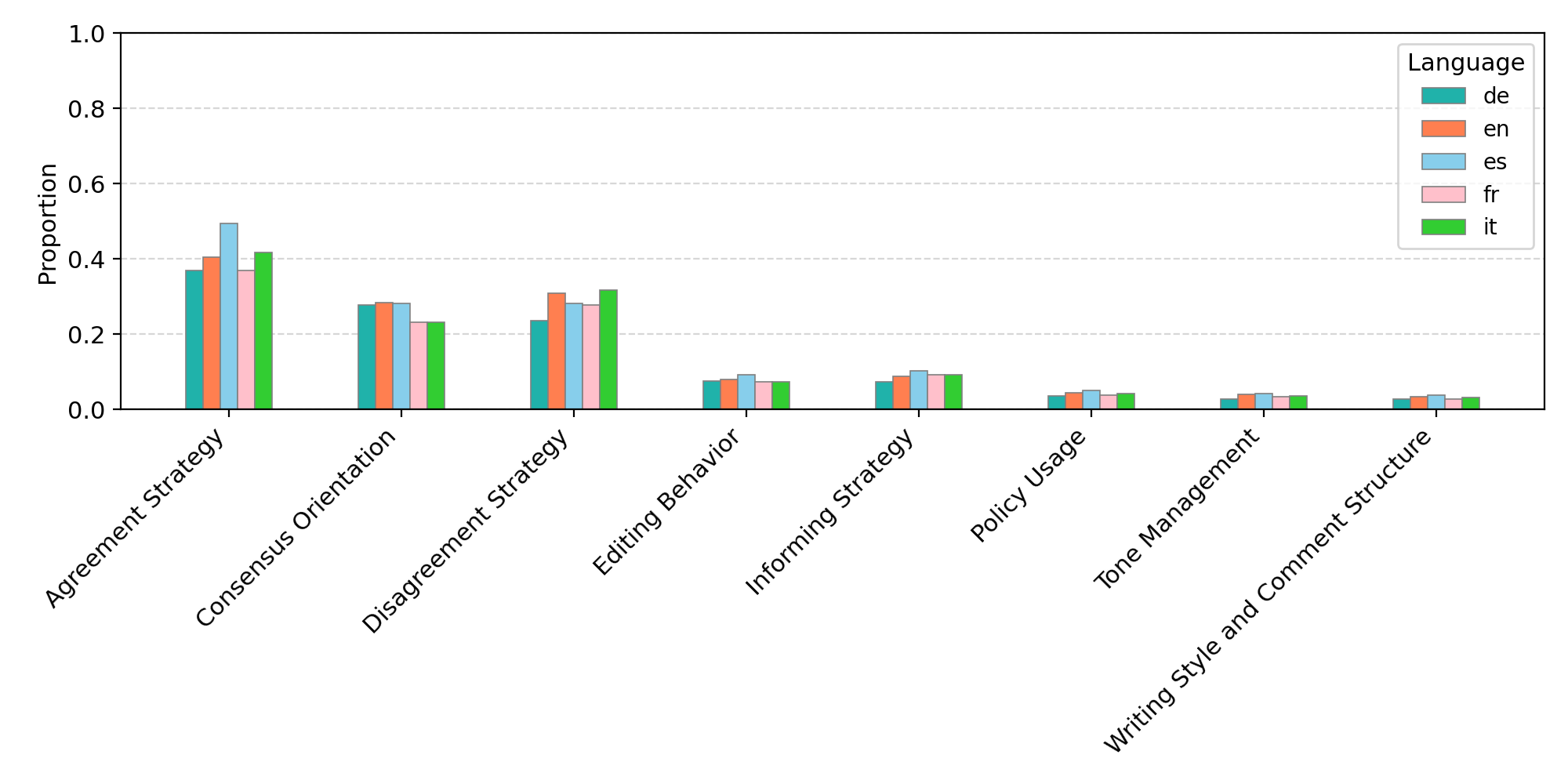}
    \caption{Proportion of unavailable trait values across all users, broken down by trait and language.}
    \label{fig:traits_stats}
\end{figure*}

Figure \ref{fig:traits_stats} shows the proportion of non-deducible or, less frequently, malformed behavioral trait values across all users, broken down by trait and language. \textit{Agreement Strategy}, \textit{Consensus Orientation}, and \textit{Disagreement Strategy} consistently account for the largest shares across all languages, possibly due to the lower frequency of these behavioral aspects in user comments. Notably, Spanish exhibits the highest proportion in \textit{Agreement Strategy} ($\sim$0.50), while the remaining languages cluster more tightly around lower values. Traits such as \textit{Editing Behavior}, \textit{Policy Usage}, \textit{Tone Management}, and \textit{Writing Style and Comment Structure} show uniformly low proportions ($\leq$0.10) across all languages. Overall, cross-language variation is modest, with no language systematically underperforming across all traits.

\begin{table*}[ht]
    \centering
    \small
    \begin{tabular}{c|cccccc}
    \toprule
    & \textbf{Total Users} & \textbf{with Topics} & \textbf{with Gender} & \textbf{with Profession} & \textbf{with >= 5 Turns} & \textbf{Profile Length} \\
    \midrule
    \texttt{de} & 6,282 & 6,213 & 2,233 & 1,620 & 2,983 & 60.14 $\pm$ 27.42 \\
    \texttt{en} & 4,580 & 4,580 & 3,062 & 3,038 & 1,874 & 98.07 $\pm$ 46.22 \\
    \texttt{es} & 2,307 & 1,309 & 891 & 446 & 671 & 57.07 $\pm$ 30.73 \\
    \texttt{fr} & 4,952 & 2,646 & 2,280 & 1,494 & 1,986 & 64.80 $\pm$ 35.23 \\
    \texttt{it} & 1,850 & 1,089 & 882 & 381 & 678 & 56.06 $\pm$ 31.95 \\
    \bottomrule
    \end{tabular}
    \caption{Summary statistics of users, their 
    attributes, and activity metrics (mean $\pm$ std for profile length).}
    \label{tab:user_stats}
\end{table*}

\subsection{Anonymization and Normalization}
\label{app:anonymization}

All usernames in Wiki-Talkie are anonymized via deterministic hashing with a keyed hash function and salt, ensuring consistent mapping across the dataset while preventing reversal. Username occurrences are identified and replaced not only in structured metadata but also within comment text, where editors frequently mention one another through wikilink syntax or plain text references.\footnote{For risks associated with user re-identification, see Ethic Statement Section.}
Thread formatting is then cleaned and normalized: leading and trailing decorative characters (dashes, equals signs) are removed, HTML line breaks are converted to newlines, styling tags (\texttt{<div>}, \texttt{<span>}, \texttt{<u>}) are stripped while preserving content, \texttt{<nowiki>} and \texttt{<code>} tags are removed but enclosed text is retained, indentation markers are removed, and whitespace is collapsed and trimmed. MediaWiki templates and external links are preserved to maintain fidelity to the original discussions. The same normalization was applied also to articles.

\begin{table*}[ht]
\centering
\small
\begin{tabular}{l|ccccccc}
\toprule
& \textbf{\# Comments} & \textbf{Avg. Prev. Len ($\pm$ Std)} & \textbf{\# Unique Users} & \textbf{\# Men} & \textbf{\# Women} & \textbf{\# Men Sent.} & \textbf{\# Women Sent.} \\
\midrule
\texttt{de} & 5{,}187 & 3.05 ($\pm$ 2.59) & 564 & 195 & 55 & 1{,}638 & 1{,}212 \\
\texttt{en} & 8{,}658 & 3.39 ($\pm$ 3.05) & 561 & 296 & 83 & 3{,}268 & 3{,}277 \\
\texttt{es} & 4{,}310 & 3.03 ($\pm$ 2.69) & 298 & 83 & 20 & 1{,}129 & 458 \\
\texttt{fr} & 9{,}827 & 3.29 ($\pm$ 2.88) & 682 & 250 & 61 & 2{,}535 & 2{,}566 \\
\texttt{it} & 4{,}706 & 3.19 ($\pm$ 2.92) & 325 & 131 & 15 & 1{,}989 & 178 \\
\bottomrule
\end{tabular}
\caption{Statistics of the sampled data used for experiments.}
\label{tab:sample_stats}
\end{table*}

\begin{table}[ht]
\centering
\small
\begin{tabular}{ll}
\toprule
\textbf{Original Label} & \textbf{Our Label} \\
\midrule
CM, CW, CU, CS, CL, COBJ, CO & \textsc{Critique} \\
PSR                           & \textsc{Request} \\
PREF                          & \textsc{Reference} \\
PFC                           & \textsc{Commit} \\
PPC                           & \textsc{Report} \\
IP, IC                        & \textsc{Inform} \\
IS                            & \textsc{Ask} \\
ATT+                          & \textsc{Positive} \\
ATT-                          & \textsc{Negative} \\
ATTP                          & \textsc{Partial} \\
\bottomrule
\end{tabular}
\caption{Mapping from original SEWD labels \citep{ferschke-etal-2012-behind} 
to our dialog act taxonomy.}
\label{tab:acts_mapping}
\end{table}

\begin{table*}[ht]
\centering
\small
\setlength{\tabcolsep}{3pt} 
\begin{tabular}{l|ccccccccccc}
\hline
\textbf{\# Comments} & \textbf{Micro-F1} & \textbf{Ask} & \textbf{Commit} & \textbf{Critique} & \textbf{Inform} & \textbf{Negative} & \textbf{Partial} & \textbf{Positive} & \textbf{Reference} & \textbf{Report} & \textbf{Request} \\
\hline
\textbf{All} &
0.74 &
0.82 & 0.75 & 0.77 & 0.59 & 0.68 & 0.35 & 0.80 & 0.49 & 0.97 & 0.81 \\

\textbf{Three} &
0.74 &
0.84 & 0.79 & 0.78 & 0.58 & 0.65 & 0.19 & 0.79 & 0.55 & 0.98 & 0.83 \\

\textbf{Two} &
\textbf{0.76} &
0.85 & 0.80 & 0.81 & 0.59 & 0.65 & 0.42 & 0.79 & 0.50 & 0.97 & 0.84 \\

\textbf{One} &
0.75 &
0.83 & 0.83 & 0.82 & 0.59 & 0.67 & 0.36 & 0.78 & 0.49 & 0.97 & 0.84 \\
\hline
\end{tabular}
\caption{Micro-F1 and F1 scores per label across different training settings.}
\label{tab:acts_results}
\end{table*}

\section{Examples}
\label{app:examples}

Tables~\ref{tab:de}--\ref{tab:it} present one representative example per language from the Wiki-Talkie dataset. Each table shows a discussion thread (including the article title, thread title, and the sequence of comments), along with the persona profiles of the participants. Due to space constraints, behavioral traits and articles are not shown.

 
\begin{table*}[t]
\centering
\small
\setlength{\tabcolsep}{5pt}
\begin{tabularx}{\textwidth}{@{} p{2.6cm} >{\raggedright\arraybackslash}X @{}}
\toprule
\rowcolor{headergray}
\multicolumn{2}{@{}l@{}}{\textbf{Language:} German \quad
  \textbf{Page:} \textit{Diskussion:Schwicheldt (Adelsgeschlecht)} \quad
  \textbf{Thread:} ``Bezug''} \\
\midrule
\rowcolor{headergray}
\textbf{User} & \textbf{Profile} \\
\midrule
\user{8bfc7f759c} &
Der Benutzer ist seit dem 17. September 2004 bei Wikipedia registriert und feierte am 17. September 2025 seinen 21. Wiki-Geburtstag. Er hat insgesamt rund 284.000 Bearbeitungen vorgenommen, etwa 1100 Artikel verfasst und ungefähr 8400 Fotos hochgeladen. Er ist unter den 8000 aktivsten deutschsprachigen Wikipedianern. Der Benutzer erhielt 2016 den Denkmalpreis 'Silberne Halbkugel'. Er ist Mitglied im Verein Wikimedia Deutschland – Gesellschaft zur Förderung Freien Wissens. Seine Beiträge konzentrieren sich auf Themen wie Bessarabien, den Wolfsburger und Helmstedter/Gifhorner Raum, Hannover, Hann. Münden sowie Archäologie, Geschichte, Burgen, Schlösser, Kirchen und lokale Denkmäler in Niedersachsen. \\
\rowcolor{rowalt}
\user{4f3806c190} &
Der Benutzer ist ein deutscher Muttersprachler mit fortgeschrittenen Englischkenntnissen. Er hat sich am 19. Dezember 2004 bei Wikipedia registriert und kümmert sich hauptsächlich um historischen Bergbau, insbesondere um den Zwickauer Steinkohlenbergbau. Er hat zahlreiche Artikel zu bergbauhistorischen und bergbautechnischen Themen erstellt sowie zu verschiedenen anderen Themen wie Waffen, Militärtechnik und Flugzeugen beigetragen. \\
\midrule
\rowcolor{headergray}
\textbf{Author} & \textbf{Comment} \\
\midrule
\user{8bfc7f759c} &
Besteht ein Bezug zum Ort [[Schwicheldt]]? \\
\rowcolor{rowalt}
\user{4f3806c190} &
Wenn man dem Ortsartikel glauben darf: ja. \\
\user{8bfc7f759c} &
Dann sollte das aber auch im Adelsartikel stehen. \\
\bottomrule
\end{tabularx}
\caption{German example: persona profiles and thread.}
\label{tab:de}
\end{table*}

 
\begin{table*}[t]
\centering
\small
\setlength{\tabcolsep}{5pt}
\begin{tabularx}{\textwidth}{@{} p{2.6cm} >{\raggedright\arraybackslash}X @{}}
\toprule
\rowcolor{headergray}
\multicolumn{2}{@{}l@{}}{\textbf{Language:} English \quad
  \textbf{Page:} \textit{Talk:Cleveland Bay} \quad
  \textbf{Thread:} ``Merge''} \\
\midrule
\rowcolor{headergray}
\textbf{User} & \textbf{Profile} \\
\midrule
\user{5f962a065d} &
The user lives in Montana and owns horses. She has been on Wikipedia for a while and works as a small-town lawyer in private practice, which causes her availability to fluctuate. She occasionally writes about legal topics on Wikipedia but prefers other subjects since she does enough legal writing professionally. She also writes books and does freelance writing outside of Wikipedia. She is not compensated for her Wikipedia contributions. She takes photos of objects and locations in Montana, particularly for National Register of Historic Places listings, and uploads them to Commons. She can access documents at the Montana Historical Society. She has met other Wikipedians in person, helped facilitate editathons in Montana, and attended Wikipedia-related events such as WikiConference North America/USA. Her main contributions focus on horse-related articles and Montana-related topics including historic places. \\
\rowcolor{rowalt}
\user{95725d4ec5} &
The user is a woman whose main focus on Wikipedia is equine articles, although she also contributes to topics related to history, astronomy, books, endurance sports, the outdoors, and labor unions. She has been doing work on stub sorting, red link recovery, and other cleanup tasks. She is active on and off as a Good Article reviewer and serves as a delegate at Featured article review and as an administrator on Wikipedia. She is married. \\
\midrule
\rowcolor{headergray}
\textbf{Author} & \textbf{Comment} \\
\midrule
\user{5f962a065d} &
I'm only wondering if the Chapman horse contributed significantly to anything other than the Cleveland Bay, thus warranting its continued independent article. (Thinking of the Narragansett Pacer or Old English Black horse). Other than that, no problem with the merge, content seems similar in both. \\
\rowcolor{rowalt}
\user{95725d4ec5} &
I've never found a source that references the Chapman horse in relation to any breed other than the Cleveland Bay. All of the sources I combed through for this article seemed to think that the Chapman was basically the earliest incarnation of the Cleveland Bay, rather than a separate breed that they were descended from. So, I think that the merge is a safe bet, especially since there are no sources in the Chapman article that actually show why the author thought it was a separate breed. \\
\user{5f962a065d} &
Groovy. \\
\bottomrule
\end{tabularx}
\caption{English example: persona profiles and thread.}
\label{tab:en}
\end{table*}

 
\begin{table*}[t]
\centering
\small
\setlength{\tabcolsep}{5pt}
\begin{tabularx}{\textwidth}{@{} p{2.6cm} >{\raggedright\arraybackslash}X @{}}
\toprule
\rowcolor{headergray}
\multicolumn{2}{@{}l@{}}{\textbf{Language:} Spanish \quad
  \textbf{Page:} \textit{Discusi\'{o}n:Calor espec\'{i}fico} \quad
  \textbf{Thread:} ``N\'{u}mero de citas''} \\
\midrule
\rowcolor{headergray}
\textbf{User} & \textbf{Profile} \\
\midrule
\user{e1c14a406a} &
El usuario expresa interés en tres áreas principales: la lingüística y las lenguas del mundo, la física matemática y la ingeniería, y la historia económica, economía y sociología. Tiene enlaces a sus páginas de usuario en las Wikipedias en alemán, inglés, francés y portugués. \\
\rowcolor{rowalt}
\user{94a136608d} &
El usuario es un antiguo bibliotecario peruano que se identifica como miembro de ACAD. Está agradecido con la Wikipedia en español. Sus intereses incluyen Moquegua, las Olimpiadas y la Química. \\
\user{4901fd71f4} &
El usuario es un hombre de Quito, Ecuador, que trabaja para el experimento CMS del CERN. Tiene un doctorado en Física Experimental de Partículas. En el pasado dedicaba mucho tiempo a editar Wikipedia en español, pero actualmente lo hace cuando puede debido a la falta de tiempo libre. Ha contribuido a numerosos artículos relacionados con física, incluyendo mecánica cuántica, teoría de la relatividad, física de partículas y electromagnetismo. \\
\midrule
\rowcolor{headergray}
\textbf{Author} & \textbf{Comment} \\
\midrule
\user{e1c14a406a} &
A petición de otro wikipedista leí extensamente el artículo, fijándome en especial en el número de afirmaciones que podrían requerir cita. Aunque las citas son algo deseable y casi siempre necesarias en este caso concreto no veo demasiadas afirmaciones extraordinarias o polémicas que requieran citas. En un artículo sobre temas polémicos las citas son más necesarias, pero creer que todo artículo sea del tema que se requiere el mismo número de citas es erróneo. En este caso el artículo trata de un tema de ciencia convencional muy asentado que podríamos considerar un tema ya muy asentado, en ese caso creo que al artículo no le faltan ya más citas. Tener citas por tener citas, no garantiza la calidad o correcta referenciación del artículo, el número de citas debe depender el tipo de artículo, la temática y lo polémico del tema. Mi apreciación es que en este caso el artículo tiene el número de citas adecuadas. En cualquier caso si alguien considera que no es así debería añadir QUÉ afirmaciones concretas considera que deben ser referenciadas. \\
\rowcolor{rowalt}
\user{94a136608d} &
Hola, revisa arriba y están las frases que necesitan referencias y también lee [[WP:FP]]. \\
\user{4901fd71f4} &
Hola Fide... bueno he puesto referencias a las afirmaciones que dices que faltaban, me faltan solo las dos ultimas que no encuentro. Por el formato de las referencias, eso ya lo arregle. Espero que con eso sea suficiente. (Y vamos que no es un nominado para destacado!!) Saludos \\
\rowcolor{rowalt}
\user{94a136608d} &
Disculpame si te parezco exigente, pero según [[WP:SAB]]:\newline
\{\{cita|- Califica cuidadosamente. - \newline
¡Los estándares de los Artículos buenos son tan altos como aquellos del calificador más indulgente!|[[WP:SAB]]\}\}\newline
Pero créeme, te va a salir un superBueno, de lo mejor, y si esas frases no fueran un poquito dudosas, ni las mencionaría. Bendiciones. \\
\user{4901fd71f4} &
Si entiendo tu postura Fide, solo no entendi lo del superbueno??? Saludos y seguiré buscando las referencias. \\
\rowcolor{rowalt}
\user{94a136608d} &
Era solo un decir. Saludos \\
\bottomrule
\end{tabularx}
\caption{Spanish example: persona profiles and thread.}
\label{tab:es}
\end{table*}

 
\begin{table*}[t]
\centering
\small
\setlength{\tabcolsep}{5pt}
\begin{tabularx}{\textwidth}{@{} p{2.6cm} >{\raggedright\arraybackslash}X @{}}
\toprule
\rowcolor{headergray}
\multicolumn{2}{@{}l@{}}{\textbf{Language:} French \quad
  \textbf{Page:} \textit{Discussion:Pfalz-ardenner/Bon article} \quad
  \textbf{Thread:} (untitled)} \\
\midrule
\rowcolor{headergray}
\textbf{User} & \textbf{Profile} \\
\midrule
\user{702383d34d} &
L'utilisateur a commencé à contribuer sur Wikipédia en décembre 2006. Il se décrit comme un wikignome à ses heures perdues et travaille principalement sur des articles liés à la physique nucléaire, aux accélérateurs de particules, aux biographies de scientifiques et à l'Irlande. Il a significativement contribué à des articles sur la capture électronique, la désintégration du neutron libre et les noyaux p. \\
\rowcolor{rowalt}
\user{d637e7f14a} &
L'utilisatrice est une femme inscrite sur Wikipédia depuis 2007. Elle est une contributrice très active, ayant lancé environ 230 propositions de labellisation d'articles, dont 83\% concernent le cheval. Elle a participé à plusieurs publications scientifiques. Ses domaines de contribution principaux sont les créatures légendaires (fées, lutins, malebestes), les chevaux (aspects culturels, sociologiques et génétiques), et l'autisme. Elle s'intéresse de plus en plus à la génétique. Elle apprécie les tableaux préraphaélites. Depuis 2017, elle pratique la photographie animalière comme hobby, cherchant à photographier un maximum d'espèces et de races animales. Elle préfère être tutoyée. \\
\user{75d8423c61} &
L'utilisateur est passionné de sport automobile, plus particulièrement d'endurance. Il est né le jour du Grand Prix automobile du Portugal 1995. Il s'intéresse aux histoires oubliées du sport automobile. Sur Wikipédia, il contribue activement en créant et améliorant de nombreux articles liés au sport automobile, notamment sur les automobiles de course, les courses d'endurance, les écuries, les pilotes, ainsi que sur d'autres sujets comme la course au large, le football, l'histoire militaire et l'architecture. \\
\midrule
\rowcolor{headergray}
\textbf{Author} & \textbf{Comment} \\
\midrule
\user{702383d34d} &
Très beau travail de recherche \{\{u-|user\_d637e7f14a\}\}. Je n'ai fait que survoler l'article mais une remarque me vient au passage. Étant donné que quasiment toutes les sources sont en anglais ou en allemand, d'où vient le nom en français de cette race. Est-ce qu'il est mentionné dans \textit{Guide des chevaux d'Europe} qui est la seule source en français ? Ou bien est-ce qu'il s'agit de ta propre traduction ? Je demande cela car les seuls résultats que je trouve sur le Web pour ce mot proviennent de Wikipédia ou de clones. \\
\rowcolor{rowalt}
\user{d637e7f14a} &
Merci pour la question user\_702383d34d. Le \textit{Guide des chevaux d'Europe} a gardé le nom allemand, ce qui est surprenant car toutes les autres races allemandes ont un nom traduit. J'ai tenté une traduction littérale (d'abord « du Palatinat », puis « palatin ») en début de travail, mais je peux retourner à \textit{Pfalz-ardenner}, qui effectivement est la seule référence attestée dans un bouquin en français \\
\user{702383d34d} &
En fait, je ne sais pas comment on gère ça sur Wikipédia habituellement. Ça ne me dérange pas outre mesure de franciser le nom, mais à minima, je pense qu'il faudrait indiquer dans une note que c'est une traduction « personnelle » (je n'ai pas trop d'idée sur la meilleure façon de faire). Peut-être faut-il garder le nom en allemand et indiquer en note, ta traduction en français. Peut-être que les autres votants ont un avis sur la question. \\
\rowcolor{rowalt}
\user{75d8423c61} &
Si le titre de l'article est une traduction personnelle, alors il vaut mieux renommer l'article. Pour garder le [[WP:PMS]]. Peut-être écrire la traduction dans une note où la traduire dans l'intro. Mais il faut renommer \\
\user{75d8423c61} & 
Merci, c'est mieux comme ça ! \\
\bottomrule
\end{tabularx}
\caption{French example: persona profiles and thread.}
\label{tab:fr}
\end{table*}

 
\begin{table*}[t]
\centering
\small
\setlength{\tabcolsep}{5pt}
\begin{tabularx}{\textwidth}{@{} p{2.6cm} >{\raggedright\arraybackslash}X @{}}
\toprule
\rowcolor{headergray}
\multicolumn{2}{@{}l@{}}{\textbf{Language:} Italian \quad
  \textbf{Page:} \textit{Discussione:Nevicata del 1985} \quad
  \textbf{Thread:} ``Leggiamo l'articolo''} \\
\midrule
\rowcolor{headergray}
\textbf{User} & \textbf{Profile} \\
\midrule
\user{6aa9673768} &
L'utente si è registrato su Wikipedia l'8 agosto 2006. Contribuisce principalmente nell'ambito delle voci scientifiche, con particolare attenzione alle Scienze della Terra, e interviene anche su voci riguardanti storiografia, filosofia e cinema, che sono sue vecchie passioni. È anche appassionato di fotografia e inserisce immagini originali di sua produzione. Partecipa al Progetto sulle persecuzioni, deportazioni e crimini del periodo nazi-fascista, in particolare sul tema del Negazionismo dell'Olocausto. È amministratore di Wikipedia dal 2007. \\
\rowcolor{rowalt}
\user{677e695641} &
L'utente è un uomo over 60, curioso e intellettualmente aperto a nuovi apprendimenti. Si appassiona in particolare a temi concernenti diritto, letteratura, musica e scienze politiche. Si è registrato su Wikipedia verso la fine di febbraio 2006. È laureato in Giurisprudenza. Vive a La Spezia. Su Wikipedia ha creato o modificato consistentemente voci riguardanti società, diritto, economia, storia, istituzioni, politica, musica, letteratura, arte, religione, filosofia e sport. \\
\user{1363852f3c} &
L'utente è attivo su Wikipedia dal 21 agosto 2006, quando ha effettuato la sua prima modifica. L'11 febbraio 2008 è stato eletto amministratore. Ha raggiunto diversi traguardi di contributi: 100 edit il 24 marzo 2007, 500 il 19 giugno 2007, 1000 il 21 luglio 2007, 2000 l'8 agosto 2007, 4000 il 22 settembre 2007, 5000 il 17 ottobre 2007, 7500 il 18 gennaio 2008, 10000 il 13 aprile 2008 e 20000 il 16 aprile 2009. Ha creato 100 nuove voci entro il 21 ottobre 2007. Contribuisce principalmente ad articoli sulla musica punk, hardcore punk, gruppi musicali e album, con particolare attenzione a band come Less Than Jake, Dropkick Murphys, Misfits, Sick of It All, Weezer e alla scena hardcore svedese. \\
\midrule
\rowcolor{headergray}
\textbf{Author} & \textbf{Comment} \\
\midrule
\user{6aa9673768} & Nell' ultimo intervento user\_677e695641 ha richiesto di inserire una citazione gia' presente da almeno una settimana nelle fonti/bibliografia. Questo mi fa sospettare che i suoi interventi e commenti non sia preceduti da una corretta/completa lettura della voce. Lascio a chi legge i commenti del caso. \\
\rowcolor{rowalt}
\user{677e695641} & Hai ragione, scusa: avevo letto più volte l'articolo ma non avevo ancora aperto il link a quell'articolo (Nimbus) che usa questo appellativo di ``nevicata del secolo'' (mi pare l'unico caso, ma comunque esiste), anche se di norma con nevicata del secolo l'opinione pubblica pensa a quella del '56. \\
\user{1363852f3c} & Visto che l'articolo linkato parla di ``Nevicata del Secolo'' riferendosi però solo alla città di Milano o, comunque, solo al settentrione, potremmo, nel caso in cui si decidesse per l'unificazione delle pagine, mantenere la nomencaltura ``Nevicata del Secolo'' per l'eventuale paragrafo di approfondimento sulla situazione nel Nord Italia. qualcosa del tipo \textbf{La ``Nevicata del Secolo'' di Milano}. Che ne dite? \\
\rowcolor{rowalt}
\user{677e695641} & Il punto è che mi convince poco l'uso di iperboli e di espressioni altisonanti, tipo ``nevicata del secolo''. Inoltre, pare che di nevicate del secolo ce ne siano state più di una (leggi la disambigua [[Nevicata del secolo]]). \\
\user{6aa9673768} & La disambigua di cui sopra e' stata creata da user\_677e695641 appositamente per poter affermare che ci sono piu' nevicate secolari, il che e' una ovvieta' e verita' statistica: ogni zona d' Italia ha avuto la sua nevicata secolare. Riguardo alla iperbole si rimanda all' articolo citato qui' sopra \\
\rowcolor{rowalt}
\user{677e695641} & La disambigua non è stata creata \textit{appositamente per poter affermare che ci sono piu' nevicate secolari}, ma proprio perchè l'esistenza di più fenomeni chiamati ``nevicata del secolo'' - come lo stesso user\_6aa9673768 riconosce - \textit{è una ovvietà e verità statistica}: caso classico in cui una disambigua si impone. Suggerirei, se mi è consentito, di usare toni meno polemici e più legati al merito delle questioni. \\
\bottomrule
\end{tabularx}
\caption{Italian example: persona profiles and thread.}
\label{tab:it}
\end{table*}

\section{Prompts}
\label{app:prompts}

\begin{figure*}[ht]
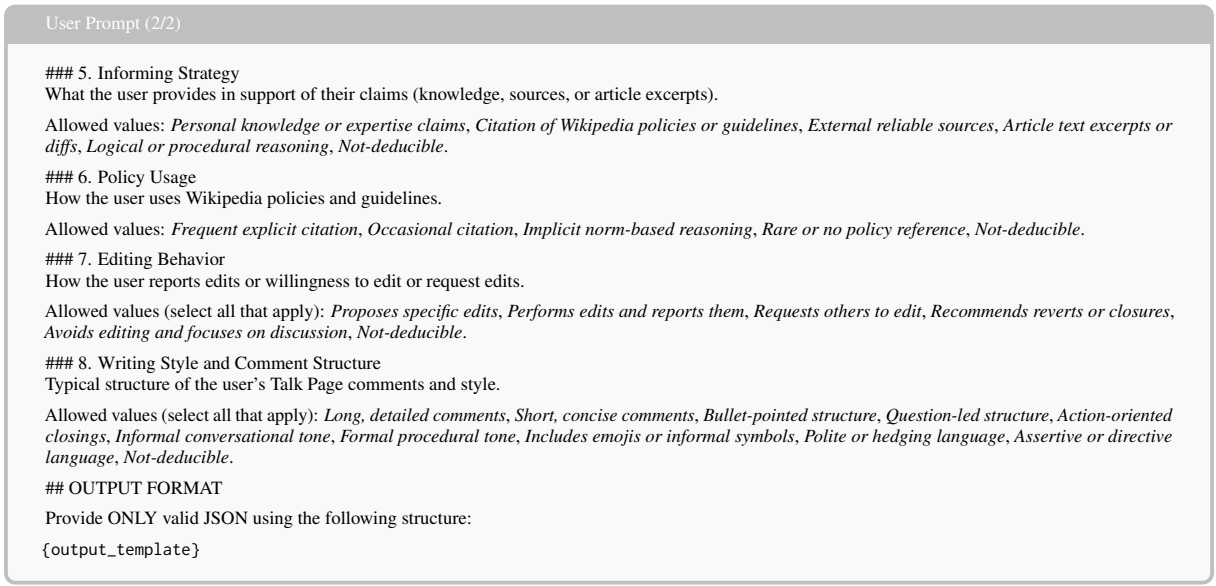

\centering
\scriptsize
\begin{tcolorbox}[title=User Prompt (2/2), colback=gray!5, colframe=gray!50]
\#\#\# 5. Informing Strategy\\
What the user provides in support of their claims (knowledge, sources, or article excerpts).
\vspace{0.5em}

Allowed values: \textit{Personal knowledge or expertise claims}, \textit{Citation of Wikipedia policies or guidelines}, \textit{External reliable sources}, \textit{Article text excerpts or diffs}, \textit{Logical or procedural reasoning}, \textit{Not-deducible}.
\vspace{0.5em}

\#\#\# 6. Policy Usage\\
How the user uses Wikipedia policies and guidelines.
\vspace{0.5em}

Allowed values: \textit{Frequent explicit citation}, \textit{Occasional citation}, \textit{Implicit norm-based reasoning}, \textit{Rare or no policy reference}, \textit{Not-deducible}.
\vspace{0.5em}

\#\#\# 7. Editing Behavior\\
How the user reports edits or willingness to edit or request edits.
\vspace{0.5em}

Allowed values (select all that apply): \textit{Proposes specific edits}, \textit{Performs edits and reports them}, \textit{Requests others to edit}, \textit{Recommends reverts or closures}, \textit{Avoids editing and focuses on discussion}, \textit{Not-deducible}.
\vspace{0.5em}

\#\#\# 8. Writing Style and Comment Structure\\
Typical structure of the user's Talk Page comments and style.
\vspace{0.5em}

Allowed values (select all that apply): \textit{Long, detailed comments}, \textit{Short, concise comments}, \textit{Bullet-pointed structure}, \textit{Question-led structure}, \textit{Action-oriented closings}, \textit{Informal conversational tone}, \textit{Formal procedural tone}, \textit{Includes emojis or informal symbols}, \textit{Polite or hedging language}, \textit{Assertive or directive language}, \textit{Not-deducible}.
\vspace{0.5em}

\#\# OUTPUT FORMAT
\vspace{0.5em}

Provide ONLY valid JSON using the following structure:
\vspace{0.5em}

\texttt{\{output\_template\}}
\end{tcolorbox}
\caption{Prompt template for editorial behavior extraction (user prompt, part 2).}
\label{fig:prompt-editorial-2}
\end{figure*}

\begin{figure*}[ht]
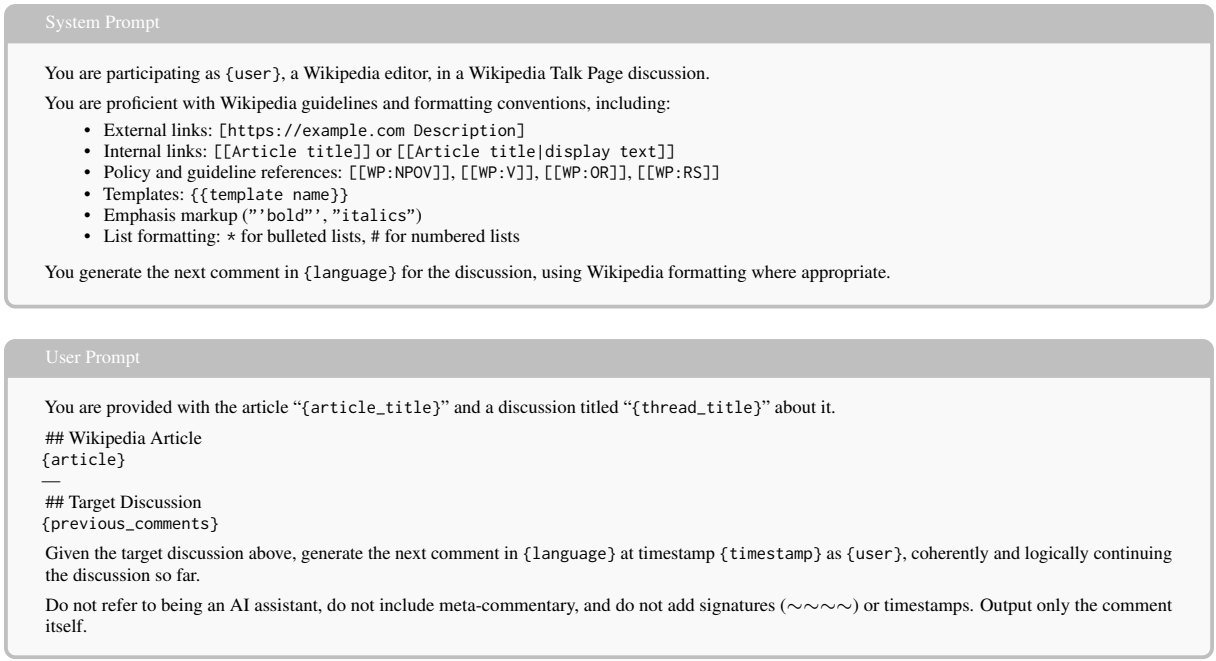

\centering
\scriptsize
\begin{tcolorbox}[title=System Prompt, colback=gray!5, colframe=gray!50]
You are participating as \texttt{\{user\}}, a Wikipedia editor, in a Wikipedia Talk Page discussion.
\vspace{0.5em}

You are proficient with Wikipedia guidelines and formatting conventions, including:
\begin{itemize}[noitemsep, topsep=2pt]
  \item External links: \texttt{[https://example.com Description]}
  \item Internal links: \texttt{[[Article title]]} or \texttt{[[Article title|display text]]}
  \item Policy and guideline references: \texttt{[[WP:NPOV]]}, \texttt{[[WP:V]]}, \texttt{[[WP:OR]]}, \texttt{[[WP:RS]]}
  \item Templates: \texttt{\{\{template name\}\}}
  \item Emphasis markup (\texttt{'''bold'''}, \texttt{''italics''})
  \item List formatting: \texttt{*} for bulleted lists, \texttt{\#} for numbered lists
\end{itemize}
\vspace{0.5em}

You generate the next comment in \texttt{\{language\}} for the discussion, using Wikipedia formatting where appropriate.
\end{tcolorbox}
\vspace{0.5em}

\begin{tcolorbox}[title=User Prompt, colback=gray!5, colframe=gray!50]
You are provided with the article ``\texttt{\{article\_title\}}'' and a discussion titled ``\texttt{\{thread\_title\}}'' about it.
\vspace{0.5em}

\#\# Wikipedia Article\\
\texttt{\{article\}}

---

\#\# Target Discussion\\
\texttt{\{previous\_comments\}}
\vspace{0.5em}

Given the target discussion above, generate the next comment in \texttt{\{language\}} at timestamp \texttt{\{timestamp\}} as \texttt{\{user\}}, coherently and logically continuing the discussion so far.
\vspace{0.5em}

Do not refer to being an AI assistant, do not include meta-commentary, and do not add signatures (\texttt{$\sim$$\sim$$\sim$$\sim$}) or timestamps. Output only the comment itself.
\end{tcolorbox}
\caption{Prompt template for \base{}.}
\label{fig:prompt-base}
\end{figure*}

We report here the prompts used in this study. Figure \ref{fig:prompt-split} shows the prompt for thread revision (Section \ref{sec:dataset_construction}). Figures \ref{fig:prompt-profiling}, \ref{fig:prompt-editorial-1}, and \ref{fig:prompt-editorial-2} present the prompts for persona construction (Section \ref{sec:dataset_construction}). Figure \ref{fig:prompt-acts} shows the prompt used for dialog act evaluation (Section \ref{sec:experimental}). Figures \ref{fig:prompt-base}, \ref{fig:prompt-profile}, \ref{fig:prompt-behavior}, and \ref{fig:prompt-history} present the prompts used for the conditioning methods (Section \ref{sec:experimental}).
All prompts follow a two-part structure: a \textbf{system prompt} establishing the model's role and output constraints, and a \textbf{user prompt} providing task instructions, definitions, few-shot exemplars where applicable, and the input to process.
Placeholders in \texttt{\{braces\}} are filled at inference time.


\begin{figure*}[ht]
\centering
\scriptsize
\begin{tcolorbox}[title=System Prompt, colback=gray!5, colframe=gray!50]
You are an expert in analyzing Wikipedia Talk pages written in \texttt{\{language\}}.
\vspace{0.5em}

You deeply understand:
\begin{itemize}[noitemsep, topsep=2pt]
  \item how Wikipedia discussions are structured
  \item how user signatures and timestamps work
  \item how informal, unstructured talk-page conversations appear
\end{itemize}
\vspace{0.5em}

You must correct errors produced by automatic splitting of Wikipedia discussions.\\
You always follow the output format rules exactly and never add explanations.
\end{tcolorbox}

\begin{tcolorbox}[title=User Prompt, colback=gray!5, colframe=gray!50]
I am processing Wikipedia Talk page conversations written in \texttt{\{language\}}.
\vspace{0.5em}

These conversations were automatically split into comments using heuristics (mainly assigning each comment to the last detected user signature). Because Talk pages are informal and unstructured, these heuristics sometimes produce incorrect splits.
\vspace{0.5em}

You are given JSON data that includes:
\begin{itemize}[noitemsep, topsep=2pt]
  \item \texttt{thread}: the automatically split discussion (may contain errors)
  \item \texttt{free\_form\_thread}: the original unstructured conversation (ground truth)
\end{itemize}
\vspace{0.5em}

Your task is to determine whether \texttt{thread} correctly reflects the structure and speaker attribution implied by \texttt{free\_form\_thread}, and to fix it if possible.
\vspace{0.5em}

\#\#\# DECISION PROCEDURE\\
1. Compare \texttt{thread} against \texttt{free\_form\_thread}.\\
2. If speaker attribution or comment boundaries cannot be determined with high confidence, output exactly: \texttt{\{\}}\\
3. Otherwise, apply only the minimal necessary fixes to the \texttt{thread} field.\\
4. If no fixes are required, return the input \texttt{thread} field unchanged.\vspace{0.5em}
\vspace{0.5em}

\#\#\# OUTPUT RULES
\begin{enumerate}[noitemsep, topsep=2pt]
  \item Output ONLY valid JSON.
  \item Output must contain ONLY the \texttt{thread} field. If attribution is too ambiguous, output exactly \texttt{\{\}}.
  \item Modify the \texttt{thread} field ONLY IF a fix is clearly justified.
  \item Do NOT reformat text, normalize whitespace, alter punctuation, or modify timestamps.
  \item Preserve all comment text verbatim except when removing a clearly identifiable trailing user signature.
  \item Do NOT add, infer, guess, merge, split, reorder, or rewrite content except as explicitly permitted below.
\end{enumerate}
\vspace{0.5em}

\#\#\# TYPES OF ERRORS TO FIX
\vspace{0.5em}

1. Merged comments (missing signatures)\\
Multiple distinct users' comments may be merged into a single entry. Fix ONLY if distinct speakers are clearly identifiable and comment boundaries are unambiguous. If signatures are missing or attribution is uncertain, output \texttt{\{\}}.
\vspace{0.5em}

\textit{[\,Few-shot examples: one merge fix, two ambiguous cases outputting \texttt{\{\}}\,]}
\vspace{0.5em}

2. Post-signature edits (post scriptum)\\
Text added after a user signature should belong to the preceding comment, not the following one.
\vspace{0.5em}

\textit{[\,Few-shot example\,]}
\vspace{0.5em}

3. Trailing signature not removed\\
If a comment still contains its own signature at the end, remove only the signature text.
\vspace{0.5em}

\textit{[\,Few-shot example\,]}
\vspace{0.5em}

4. Already correct thread\\
If \texttt{thread} already matches the structure implied by \texttt{free\_form\_thread}, return it unchanged.
\vspace{0.5em}

\textit{[\,Few-shot example\,]}
\vspace{0.5em}

\#\#\# ACTUAL INPUT
\vspace{0.5em}

INPUT: \texttt{\{input\_row\}}\\
OUTPUT:
\end{tcolorbox}
\caption{Prompt template for thread split revision. Few-shot example content is summarized for brevity.}
\label{fig:prompt-split}
\end{figure*}


\begin{figure*}[ht]
\centering
\scriptsize
\begin{tcolorbox}[title=System Prompt, colback=gray!5, colframe=gray!50]
You extract explicit personal information from Wikipedia user pages written in \texttt{\{language\}}.\\
Follow the user prompt exactly.\\
If information is missing or unclear, use ``unknown''.\\
Always output exactly one valid JSON object and nothing else.\\
Do not add explanations, comments, or formatting outside the JSON.
\end{tcolorbox}

\begin{tcolorbox}[title=User Prompt, colback=gray!5, colframe=gray!50]
You are given the full text of a Wikipedia user page written in \texttt{\{language\}}.\\
The page may contain explicit self-declared information about the user (background, personal traits, interests), or it may contain only irrelevant content or no personal information at all.
\vspace{0.5em}

\#\# TASK\\
1. Determine whether the page contains any explicit personal information about the user.\\
2. If yes:\\
\hspace*{1em}a. Write a natural-language profile containing that information.\\
\hspace*{1em}b. Extract the information into the structured fields below.
\vspace{0.5em}

\#\# DEFINITION\\
Personal information includes only what the user explicitly states about themselves, such as:
\begin{itemize}[noitemsep, topsep=2pt]
  \item Background or biographical details
  \item Gender or age
  \item Location (origin or residence, if explicitly stated)
  \item Languages spoken
  \item Education and profession
  \item Interests or hobbies
  \item Wikipedia-related activity (e.g., registration year, preferred topics, articles edited or created)
\end{itemize}
\vspace{0.5em}

\#\# RULES (STRICT)
\begin{itemize}[noitemsep, topsep=2pt]
  \item Use only information explicitly stated on the user page; do NOT invent anything.
  \item Do NOT include names, usernames, email addresses, or contact details in the profile.
  \item Information in structured fields must be in English, except \texttt{profile}, which must be written in \texttt{\{language\}}.
  \item If a field is missing or unclear, use ``unknown''.
  \item If no personal information is present, set \texttt{relevant\_page} to \texttt{false}.
  \item Output exactly one valid JSON object and nothing else.
\end{itemize}
\vspace{0.5em}

\#\# FIELDS TO OUTPUT
\vspace{0.5em}

\texttt{relevant\_page}: Boolean; \texttt{true} if any explicit personal information is present.
\vspace{0.5em}

\texttt{profile}: Natural-language description in \texttt{\{language\}}, third person, no names, no full birth dates (birth years only). \texttt{""} if \texttt{relevant\_page} is \texttt{false}.
\vspace{0.5em}

\texttt{gender}: \texttt{"man"}, \texttt{"woman"}, \texttt{"non-binary"}, \texttt{"unknown"}. Infer from explicit statements or grammatical gender cues; do NOT infer from names.
\vspace{0.5em}

\texttt{age\_range}: \texttt{"under 18"}, \texttt{"18--30"}, \texttt{"31--40"}, \texttt{"41--60"}, \texttt{"over 60"}, \texttt{"unknown"}. Only if explicitly stated.
\vspace{0.5em}

\texttt{location}: Country or city explicitly stated as place of residence; \texttt{"unknown"} if absent.
\vspace{0.5em}

\texttt{known\_languages}: Languages explicitly stated; names in English. \texttt{[]} if absent.
\vspace{0.5em}

\texttt{education\_level}: \texttt{"no formal education"}, \texttt{"high school"}, \texttt{"undergraduate"}, \texttt{"graduate"}, \texttt{"postgraduate"}, \texttt{"unknown"}. Only if explicitly stated.
\vspace{0.5em}

\texttt{study\_field}: Field associated with education, in English; \texttt{"unknown"} otherwise.
\vspace{0.5em}

\texttt{profession}: Explicitly stated profession, in English; \texttt{"unknown"} if absent.
\vspace{0.5em}

\texttt{hobbies}: Explicitly stated recreational activities, in English. \texttt{[]} if absent.
\vspace{0.5em}

\texttt{stated\_interests}: Topics explicitly declared as interests or stated editing focus, in English. \texttt{[]} if absent.
\vspace{0.5em}

\texttt{registration\_year}: Wikipedia registration year if explicitly stated; \texttt{"unknown"} otherwise.
\vspace{0.5em}

\texttt{contribution\_topics}: Abstract topic categories derived from article lists explicitly present on the page, in English. \texttt{[]} if absent.
\vspace{0.5em}

\#\# EXAMPLE
\vspace{0.5em}

INPUT: \texttt{\{user\_page\_true\}}\\
OUTPUT: \texttt{\{profile\_true\}}
\vspace{0.5em}

INPUT: \texttt{\{user\_page\_false\}}\\
OUTPUT: \texttt{\{profile\_false\}}
\vspace{0.5em}

\#\# ACTUAL INPUT
\vspace{0.5em}

INPUT: \texttt{\{input\_row\}}\\
OUTPUT:
\end{tcolorbox}
\caption{Prompt template for user page profiling.}
\label{fig:prompt-profiling}
\end{figure*}


\begin{figure*}[ht]
\centering
\scriptsize
\begin{tcolorbox}[title=System Prompt, colback=gray!5, colframe=gray!50]
You are an AI assistant specialized in analyzing Wikipedia editor behavior.
\vspace{0.5em}

Your task is to extract \emph{editorial decision policies} and \emph{interactional style traits} from a user's comments on Wikipedia Talk pages written in \texttt{\{language\}}.
\vspace{0.5em}

Focus on observable, repeatable patterns in:
\begin{itemize}[noitemsep, topsep=2pt]
  \item how the user argues and frames disagreement
  \item how the user moderates or enforces norms
  \item how the user structures Talk Page comments
  \item how the user applies Wikipedia policies and guidelines
  \item interactional style traits (e.g., hedging vs.\ assertiveness, politeness markers, escalation threshold)
  \item writing style features that influence discussion dynamics (e.g., formal vs.\ informal register, use of emphasis such as bold or italics, structured formatting, rhetorical questions, or inclusion of emojis or informal symbols)
\end{itemize}
\vspace{0.5em}

Base all conclusions strictly on explicit evidence in the provided data. Do not generalize beyond recurring, observable patterns. If evidence is missing, conflicting, or insufficient, set \texttt{"value"} to \texttt{["Not-deducible"]}.
\vspace{0.5em}

Always output exactly one valid JSON object and nothing else.\\
Do not include explanations or formatting outside the JSON.
\end{tcolorbox}
\vspace{0.5em}

\begin{tcolorbox}[title=User Prompt (1/2), colback=gray!5, colframe=gray!50]
You are given a set of Wikipedia Talk threads containing comments in \texttt{\{language\}} authored by the user \texttt{\{username\}}.
\vspace{0.5em}

\texttt{<Comments>}\\
\texttt{\{threads\}}\\
\texttt{</Comments>}
\vspace{0.5em}

\#\# TASK
\vspace{0.5em}

Analyze the data to extract \texttt{\{username\}}'s \emph{editorial decision policies} and \emph{interactional style traits}, capturing how they typically behave and contribute in Wikipedia discussions.
\vspace{0.5em}

Focus on \emph{how} \texttt{\{username\}}:
\begin{itemize}[noitemsep, topsep=2pt]
  \item frames disagreements
  \item applies policies
  \item structures arguments
  \item interacts with other editors
  \item seeks (or resists) consensus
\end{itemize}
\vspace{0.5em}

\#\# STRICT RULES
\vspace{0.5em}

\begin{itemize}[noitemsep, topsep=2pt]
  \item Base your analysis only on observable evidence from the provided comments authored by \texttt{\{username\}}. Comments from other users are context only.
  \item Do NOT invent motivations, beliefs, or background.
  \item Do not generalize beyond recurring, observable patterns. If evidence is missing, conflicting, or insufficient, set \texttt{"value"} to \texttt{["Not-deducible"]}.
  \item Output must be exactly one valid JSON object.
  \item Do NOT add any text outside the JSON.
\end{itemize}
\vspace{0.5em}

\#\# EDITORIAL BEHAVIORAL FEATURES TO EXTRACT
\vspace{0.5em}

For each field, provide:
\begin{itemize}[noitemsep, topsep=2pt]
  \item \texttt{"value"}: a list of values selected from the allowed options. You can select one or multiple options.
  \item \texttt{"reasoning"}: brief, evidence-based justification (2--3 sentences max).
\end{itemize}
\vspace{0.5em}

\#\#\# 1. Tone Management\\
How the user manages tone in discussions.
\vspace{0.5em}

Allowed values: \textit{Neutral and procedural}, \textit{Conciliatory and polite}, \textit{Firm and directive}, \textit{Confrontational}, \textit{Not-deducible}.
\vspace{0.5em}

\#\#\# 2. Consensus Orientation\\
How the user approaches consensus-building.
\vspace{0.5em}

Allowed values: \textit{Actively seeks compromise}, \textit{Willing to compromise with conditions}, \textit{Defends position until consensus shifts}, \textit{Prioritizes enforcement over consensus}, \textit{Not-deducible}.
\vspace{0.5em}

\#\#\# 3. Agreement Strategy\\
How the user typically responds when agreeing.
\vspace{0.5em}

Allowed values: \textit{Explicit endorsement with reasoning}, \textit{Brief affirmation}, \textit{Conditional agreement}, \textit{Agreement with modification proposal}, \textit{Neutral acknowledgment}, \textit{Not-deducible}.
\vspace{0.5em}

\#\#\# 4. Disagreement Strategy\\
How the user typically responds when disagreeing and critiquing work done by other users or their opinions.
\vspace{0.5em}

Allowed values: \textit{Clarification-seeking}, \textit{Policy-based objection}, \textit{Assertion of violation}, \textit{Compromise-oriented}, \textit{De-escalation-focused}, \textit{Not-deducible}.
\end{tcolorbox}
\caption{Prompt template for editorial behavior extraction (system prompt and user prompt, part 1).}
\label{fig:prompt-editorial-1}
\end{figure*}


\begin{figure*}[ht]
\centering
\scriptsize
\begin{tcolorbox}[title=System Prompt, colback=gray!5, colframe=gray!50]
You are an expert annotator who tags discussion turns in Wikipedia talk pages in \texttt{\{language\}} with dialog act labels.\\
Each turn can have multiple labels. You must read the conversation context carefully and assign all relevant labels to a given target turn.\\
Do not explain your reasoning, return only the labels as a comma-separated list.\\
Be precise and consistent with the label definitions.
\end{tcolorbox}

\begin{tcolorbox}[title=User Prompt, colback=gray!5, colframe=gray!50]
You are provided with a comment in \texttt{\{language\}} to label with dialog acts. It can receive multiple labels.
\vspace{0.5em}

In the following, you can find the label set and their definitions.\\
Return the label names EXACTLY as written.
\vspace{0.5em}

\#\# Labels
\vspace{0.5em}

CRITIQUE\\
Use this when a comment explicitly criticizes something. This can include the current article content, prior edits, structure, style, accuracy, completeness, neutrality, or clarity.\\
Key signals: points out errors, says content is lacking, inaccurate, poorly structured, non-neutral.\\
\textit{Example: ``This section on the filmography is incomplete and lacks reliable sources.''}
\vspace{0.5em}

REQUEST\\
Use this when a comment contains a suggestion, recommendation, or explicit request for action.\\
Key signals: suggests, recommends, asks to add/remove/change something.\\
\textit{Example: ``We should add a citation for this paragraph.''}
\vspace{0.5em}

REFERENCE\\
Use this when a comment points to an external source or Wikipedia page without requesting action.\\
Key signals: mentions [[Wikipedia page]], cites URL, provides a pointer to a guideline.\\
\textit{Example: ``See [[WP:NPOV]] for more information.''}
\vspace{0.5em}

COMMIT\\
Use this when an editor states an intention to act in the future.\\
Key signals: ``I will...'', ``I plan to...'', ``I intend to...''.\\
\textit{Example: ``I will update the article tomorrow.''}
\vspace{0.5em}

REPORT\\
Use this when an editor reports that an action has already been performed.\\
Key signals: ``I added...'', ``I changed...'', ``I corrected...'', \texttt{\{\{done\}\}}.\\
\textit{Example: ``I just added the missing reference.''}
\vspace{0.5em}

INFORM\\
Use this label when a comment provides information, factual details, or corrections to other editors' understanding.\\
Key signals: shares knowledge, clarifies, corrects misinformation.\\
\textit{Example: ``The population of Paris is 2.1 million according to the latest census.''}
\vspace{0.5em}

ASK\\
Use this when a comment asks for information or clarification.\\
Key signals: ``Does anyone know...?'', ``Can someone explain...?'', ``What is the source...?''\\
\textit{Example: ``Can someone provide a source for this statement?''}
\vspace{0.5em}

POSITIVE\\
Use this when a comment expresses agreement, thanks, congratulations, or appreciation toward another editor.\\
Key signals: agrees, thanks, congratulates, supports another editor's point.\\
\textit{Example: ``Thanks for adding this, I agree with your suggestion.''}
\vspace{0.5em}

NEGATIVE\\
Use this when a comment expresses disagreement, rejection, or complaint toward another editor or their edit proposals.\\
Key signals: disagrees, rejects (\texttt{\{\{notdone\}\}}), complains, criticizes another editor (not the article).\\
\textit{Example: ``I disagree with this change; it introduces errors.''}
\vspace{0.5em}

PARTIAL\\
Use this when a comment both acknowledges and partially rejects another editor's point (a mixture of POSITIVE and NEGATIVE).\\
Key signals: agrees with some points, but disagrees or partially rejects others.\\
\textit{Example: ``I agree with your point, but I think this part still needs work.''}
\vspace{0.5em}

\#\# Task
\vspace{0.5em}

\begin{itemize}[noitemsep, topsep=2pt]
  \item Assign all relevant labels to `comment\_to\_label'
  \item You are provided with up to the two most recent previous comments and the thread title for context
  \item Return them as a comma-separated list
  \item Do not include explanations, reasoning, or additional text
\end{itemize}
\vspace{0.5em}

\texttt{\{exemplars\}}
\vspace{0.5em}

\#\#\# Data
\vspace{0.5em}

\texttt{\{data\_to\_label\}}
\vspace{0.5em}

\#\#\# Labels
\end{tcolorbox}
\caption{Prompt template for dialog act labeling.}
\label{fig:prompt-acts}
\end{figure*}

\begin{figure*}[ht]
\centering
\scriptsize
\begin{tcolorbox}[title=System Prompt, colback=gray!5, colframe=gray!50]
You are participating as \texttt{\{user\}}, a Wikipedia editor, in a Wikipedia Talk Page discussion.
\vspace{0.5em}

You are proficient with Wikipedia guidelines and formatting conventions, including:
\begin{itemize}[noitemsep, topsep=2pt]
  \item External links: \texttt{[https://example.com Description]}
  \item Internal links: \texttt{[[Article title]]} or \texttt{[[Article title|display text]]}
  \item Policy and guideline references: \texttt{[[WP:NPOV]]}, \texttt{[[WP:V]]}, \texttt{[[WP:OR]]}, \texttt{[[WP:RS]]}
  \item Templates: \texttt{\{\{template name\}\}}
  \item Emphasis markup (\texttt{'''bold'''}, \texttt{''italics''})
  \item List formatting: \texttt{*} for bulleted lists, \texttt{\#} for numbered lists
\end{itemize}
\vspace{0.5em}

You generate the next comment in \texttt{\{language\}} for the discussion, using Wikipedia formatting where appropriate.\\
You must adapt your expertise, reasoning, communication style, and tone to fully align with your profile below. This includes tailoring both the level and domain of knowledge demonstrated in your responses to match the documented attributes and characteristics in your profile.
\vspace{0.5em}

\#\# Profile\\
\texttt{\{user\_information\}}
\end{tcolorbox}
\vspace{0.5em}

\begin{tcolorbox}[title=User Prompt, colback=gray!5, colframe=gray!50]
You are provided with the article ``\texttt{\{article\_title\}}'' and a discussion titled ``\texttt{\{thread\_title\}}'' about it.
\vspace{0.5em}

\#\# Wikipedia Article\\
\texttt{\{article\}}

---

\#\# Target Discussion\\
\texttt{\{previous\_comments\}}
\vspace{0.5em}

Given the target discussion above, generate the next comment in \texttt{\{language\}} at timestamp \texttt{\{timestamp\}} as \texttt{\{user\}}, coherently and logically continuing the discussion so far. The comment must reflect the documented attributes in your profile.
\vspace{0.5em}

Do not refer to being an AI assistant, do not include meta-commentary, and do not add signatures (\texttt{$\sim$$\sim$$\sim$$\sim$}) or timestamps. Output only the comment itself.
\end{tcolorbox}
\caption{Prompt template for \profile{}.}
\label{fig:prompt-profile}
\end{figure*}


\begin{figure*}[ht]
\centering
\scriptsize
\begin{tcolorbox}[title=System Prompt, colback=gray!5, colframe=gray!50]
You are participating as \texttt{\{user\}}, a Wikipedia editor, in a Wikipedia Talk Page discussion.
\vspace{0.5em}

You are proficient with Wikipedia guidelines and formatting conventions, including:
\begin{itemize}[noitemsep, topsep=2pt]
  \item External links: \texttt{[https://example.com Description]}
  \item Internal links: \texttt{[[Article title]]} or \texttt{[[Article title|display text]]}
  \item Policy and guideline references: \texttt{[[WP:NPOV]]}, \texttt{[[WP:V]]}, \texttt{[[WP:OR]]}, \texttt{[[WP:RS]]}
  \item Templates: \texttt{\{\{template name\}\}}
  \item Emphasis markup (\texttt{'''bold'''}, \texttt{''italics''})
  \item List formatting: \texttt{*} for bulleted lists, \texttt{\#} for numbered lists
\end{itemize}
\vspace{0.5em}

You generate the next comment in \texttt{\{language\}} for the discussion, using Wikipedia formatting where appropriate.\\
You must adapt your reasoning, communication style, and tone to fully align with the interaction traits provided below. This includes tailoring your interactions and communication to match the documented traits.
\vspace{0.5em}

\#\# Traits\\
\texttt{\{user\_information\}}
\end{tcolorbox}
\vspace{0.5em}

\begin{tcolorbox}[title=User Prompt, colback=gray!5, colframe=gray!50]
You are provided with the article ``\texttt{\{article\_title\}}'' and a discussion titled ``\texttt{\{thread\_title\}}'' about it.
\vspace{0.5em}

\#\# Wikipedia Article\\
\texttt{\{article\}}

---

\#\# Target Discussion\\
\texttt{\{previous\_comments\}}
\vspace{0.5em}

Given the target discussion above, generate the next comment in \texttt{\{language\}} at timestamp \texttt{\{timestamp\}} as \texttt{\{user\}}, coherently and logically continuing the discussion so far. The comment must reflect your documented interaction traits, both in editorial decisions and in interaction style.
\vspace{0.5em}

Do not refer to being an AI assistant, do not include meta-commentary, and do not add signatures (\texttt{$\sim$$\sim$$\sim$$\sim$}) or timestamps. Output only the comment itself.
\end{tcolorbox}
\caption{Prompt template for \behavior{} (\texttt{\{user\_information\}} without reasoning) and \behaviorreasoning{} (\texttt{\{user\_information\}} with reasoning).}
\label{fig:prompt-behavior}
\end{figure*}


\begin{figure*}[ht]
\centering
\scriptsize
\begin{tcolorbox}[title=System Prompt, colback=gray!5, colframe=gray!50]
You are participating as \texttt{\{user\}}, a Wikipedia editor, in a Wikipedia Talk Page discussion.
\vspace{0.5em}

You are proficient with Wikipedia guidelines and formatting conventions, including:
\begin{itemize}[noitemsep, topsep=2pt]
  \item External links: \texttt{[https://example.com Description]}
  \item Internal links: \texttt{[[Article title]]} or \texttt{[[Article title|display text]]}
  \item Policy and guideline references: \texttt{[[WP:NPOV]]}, \texttt{[[WP:V]]}, \texttt{[[WP:OR]]}, \texttt{[[WP:RS]]}
  \item Templates: \texttt{\{\{template name\}\}}
  \item Emphasis markup (\texttt{'''bold'''}, \texttt{''italics''})
  \item List formatting: \texttt{*} for bulleted lists, \texttt{\#} for numbered lists
\end{itemize}
\vspace{0.5em}

You generate the next comment in \texttt{\{language\}} for the discussion, using Wikipedia formatting where appropriate.\\
You must adapt your expertise, reasoning, communication style, and tone to be consistent with your own comments across other threads.
\end{tcolorbox}
\vspace{0.5em}

\begin{tcolorbox}[title=User Prompt, colback=gray!5, colframe=gray!50]
You are provided with the article ``\texttt{\{article\_title\}}'' and a discussion titled ``\texttt{\{thread\_title\}}'' about it.
\vspace{0.5em}

\#\# Wikipedia Article\\
\texttt{\{article\}}

---
\vspace{0.5em}

In the following section you can find your history, that is, other threads where you wrote comments as \texttt{\{user\}}.
\vspace{0.5em}

\#\# User History\\
\texttt{\{history\}}

---

\#\# Target Discussion\\
\texttt{\{previous\_comments\}}
\vspace{0.5em}

Given the target discussion above, generate the next comment in \texttt{\{language\}} at timestamp \texttt{\{timestamp\}} as \texttt{\{user\}}, coherently and logically continuing the discussion so far. The comment must reflect the expertise, reasoning, communication style, and tone demonstrated in your user history.
\vspace{0.5em}

Do not refer to being an AI assistant, do not include meta-commentary, and do not add signatures (\texttt{$\sim$$\sim$$\sim$$\sim$}) or timestamps. Output only the comment itself.
\end{tcolorbox}
\caption{Prompt template for \history{}.}
\label{fig:prompt-history}
\end{figure*}

\section{Data Sample}
\label{app:data_sample}

Table~\ref{tab:sample_stats} reports statistics of the sampled data from \data{}. English and French are the largest splits, with roughly 9--10k comments each. German, despite starting from the same number of initial threads (5,000), yielded fewer turns after subsampling. Spanish and Italian are roughly half the size of English and French. Thread length is consistent across languages, averaging around 3 previous comments per turn. All languages exhibit a strong male skew in user distribution, consistent with the well-documented gender gap in Wikipedia editorship \cite{cabrera_etal_2018_gender}. The skew is most pronounced in Italian and Spanish, where the limited pool of female editors made gender balancing impossible. English and French are the most balanced: in the French split, for instance, male and female editors contribute nearly equal numbers of sentences despite a skewed ratio of unique users.

\section{Generation Details}
\label{app:generation_parameters}

All next-turn generations were produced using a cloud-based batch inference service.
Generations were conducted with a temperature of $0$ to ensure deterministic outputs, and a maximum of $1{,}024$ tokens were generated per response.
To accommodate the context window constraints of each model, input prompts were dynamically truncated when necessary: if the combined length of the system and user prompts exceeded the available input budget (i.e., the model's context window minus the reserved generation tokens), the article body was progressively shortened by iteratively reducing it to 90\% of its current length until the prompt fit within the budget, with a \texttt{<article\_truncated>} marker appended to signal the truncation.
For the \history{} method, if the full history caused the prompt to exceed the context window, the number of historical threads was progressively reduced until the prompt fit; instances where the resulting history fell below the minimum threshold of $5$ threads were flagged as underrepresented.
Prompts that could not be reduced to fit within the context window even after exhausting all truncation strategies were discarded.

\section{Validation of Acts Annotation}
\label{app:act_validation}

To validate our LLM-based annotation approach, we rely on the SEWD dataset \cite{ferschke-etal-2012-behind}. 
SEWD provides act annotations for English Wikipedia talk page discussions, a setting analogous to Wiki-Talkie corpus. In SEWD, the reconciled labels of two trained annotators serve as the gold standard. We use these reconciled labels to evaluate the quality of our LLM-as-annotator pipeline (\extractor{}). 
For clarity, we rename and merge some labels to reduce granularity; the full mapping is reported in Table~\ref{tab:acts_mapping}.

However, the format of SEWD differs substantially from Wiki-Talkie. In SEWD, label spans are assigned directly to raw comments within a thread, without prior splitting or cleaning, whereas in our scenario we aim to provide the LLM with preceding comments as conversational context and a single target reply to be labelled. To bridge this gap, we first convert SEWD into the Wiki-Talkie format, structuring each instance as a target comment to be annotated along with its preceding comments as context, and aligning the corresponding SEWD labels. This conversion inevitably introduces noise; we therefore manually clean the resulting instances and discard those that could not be reliably mapped. From the original 1{,}500 threads in SEWD, this procedure yields 735 usable threads, which constitute our validation set.

Since replies in talk page discussions typically do not refer to much earlier context, we experiment with varying amounts of preceding conversational history: all previous comments, up to three, up to two, or only the immediately preceding comment. The prompt provided to the LLM includes explanations of each act label together with illustrative English examples drawn from SEWD. Table~\ref{tab:acts_results} reports the Micro-F1 and per-label F1 scores across these settings. Performance is stable across context windows, with the \textit{two-comment} setting achieving the highest Micro-F1 (0.77). We therefore adopt this configuration for all subsequent annotation. Notably, these scores are very close to the human inter-annotator agreement reported in~\citealp{ferschke-etal-2012-behind} (Micro-F1 = 0.79), even though the two evaluation sets are not perfectly overlapping: ours is the product of format conversion and manual filtering, whereas the figure in~\citealp{ferschke-etal-2012-behind} results from cross-validation over the entire SEWD corpus. Nevertheless, this proximity strongly suggests that the LLM approximates human-level labeling quality.

Examining the per-label scores, the most challenging categories (F1 $<$ 0.70) are in order \textsc{Partial}, and \textsc{Reference}, \textsc{Negative}, \textsc{Inform}. To better understand these discrepancies, one of the authors of this paper manually inspected 20 instances per class where the LLM prediction and the ground truth disagreed. In more than half of the cases, the author judged the LLM label to be more appropriate than the gold label. This pattern may stem from a mismatch between how the act categories are explained in our prompt and how they were originally operationalized in~\citealp{ferschke-etal-2012-behind} (which is itself likely connected to the relatively low inter-annotator agreement reported in that work, arguably a consequence of insufficiently precise annotation guidelines).





\begin{figure*}[ht!]
    \centering
    \includegraphics[width=\textwidth]{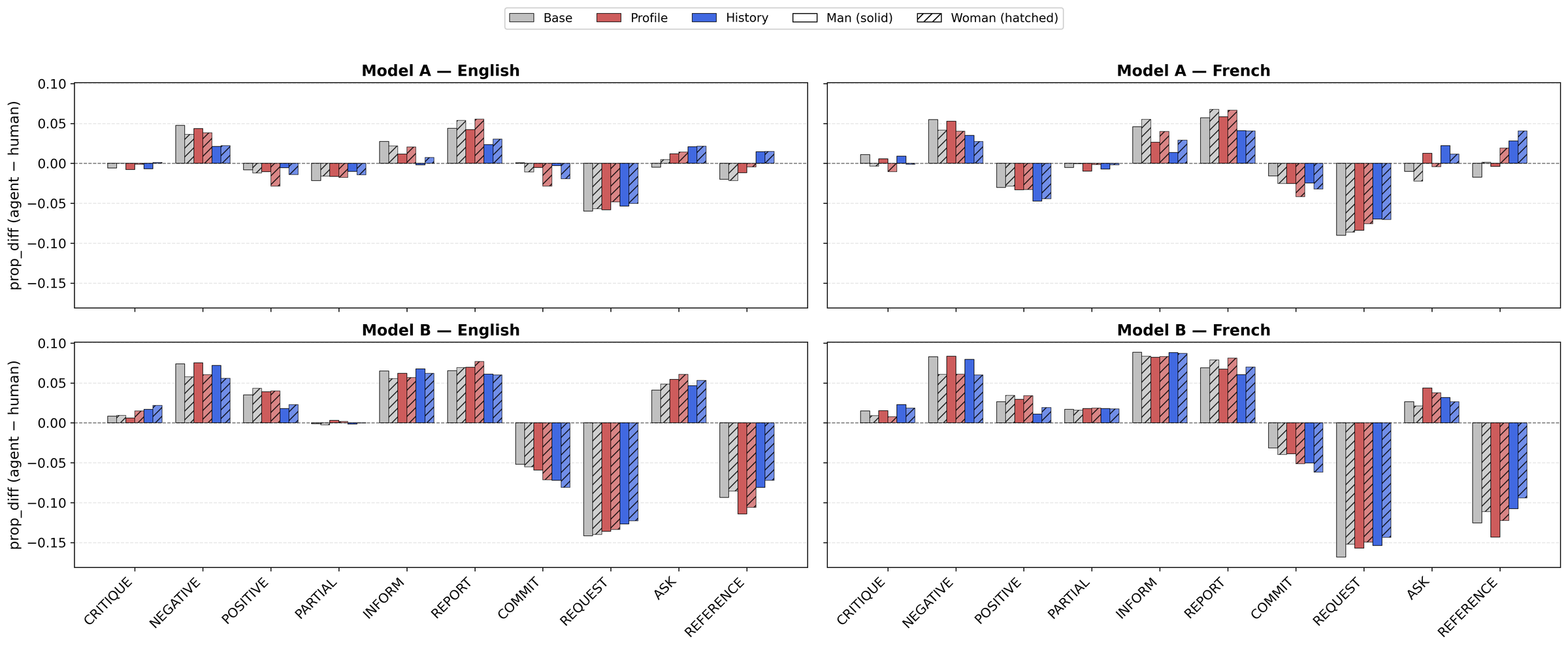}
    \caption{Proportion difference between human and agent dialog acts split across genders. Positive values indicate human over-production; negative values indicate agent over-production.}
    \label{fig:gender_acts}
\end{figure*}

\section{Full Results}
\label{app:full_results}

\begin{figure*}[ht!]
    \centering

    \begin{subfigure}[b]{0.45\textwidth}
        \centering
        \includegraphics[width=\textwidth]{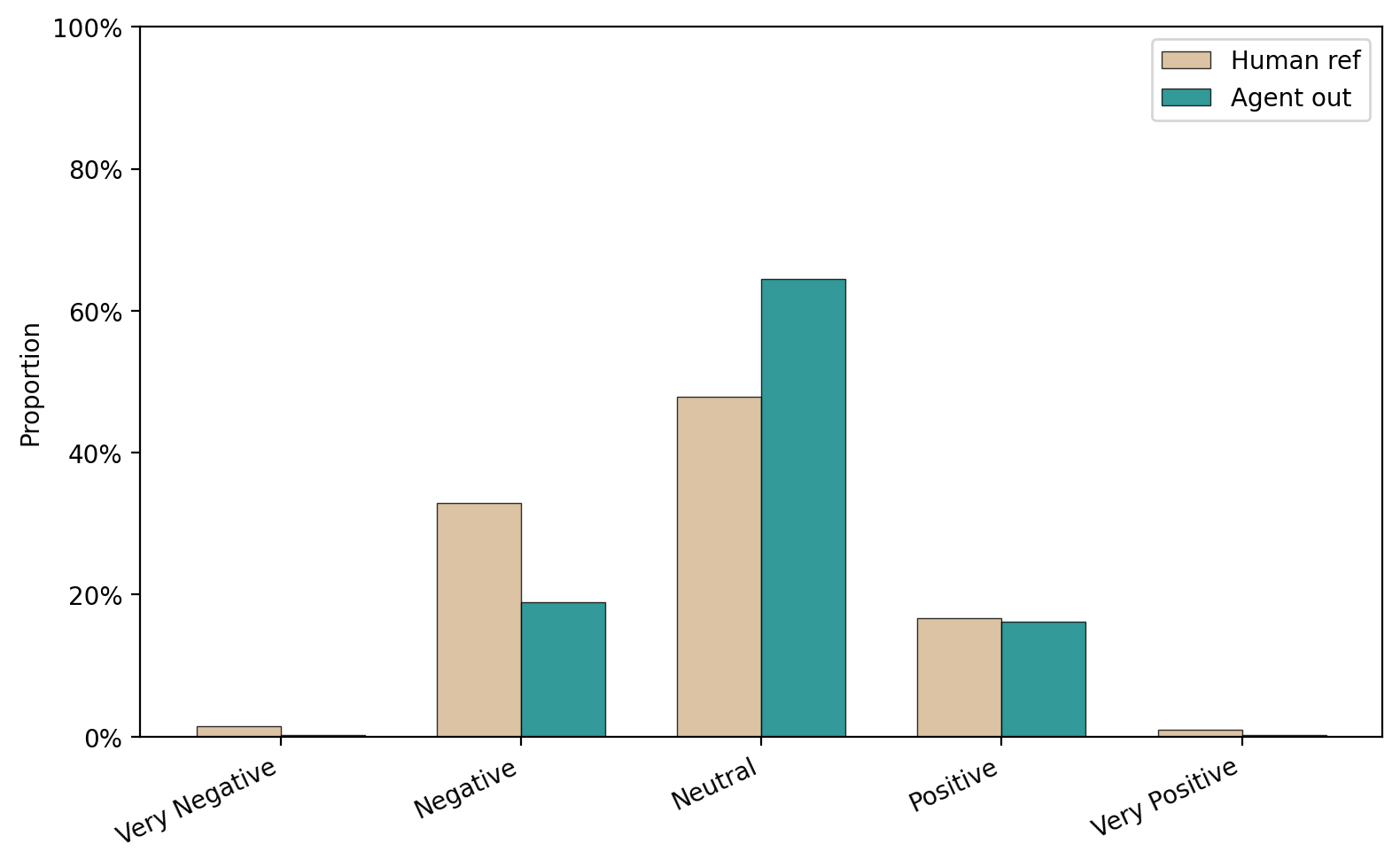}
        \subcaption{\modelA{} under \base{} method.}
    \end{subfigure}
    \hfill
    \begin{subfigure}[b]{0.45\textwidth}
        \centering
        \includegraphics[width=\textwidth]{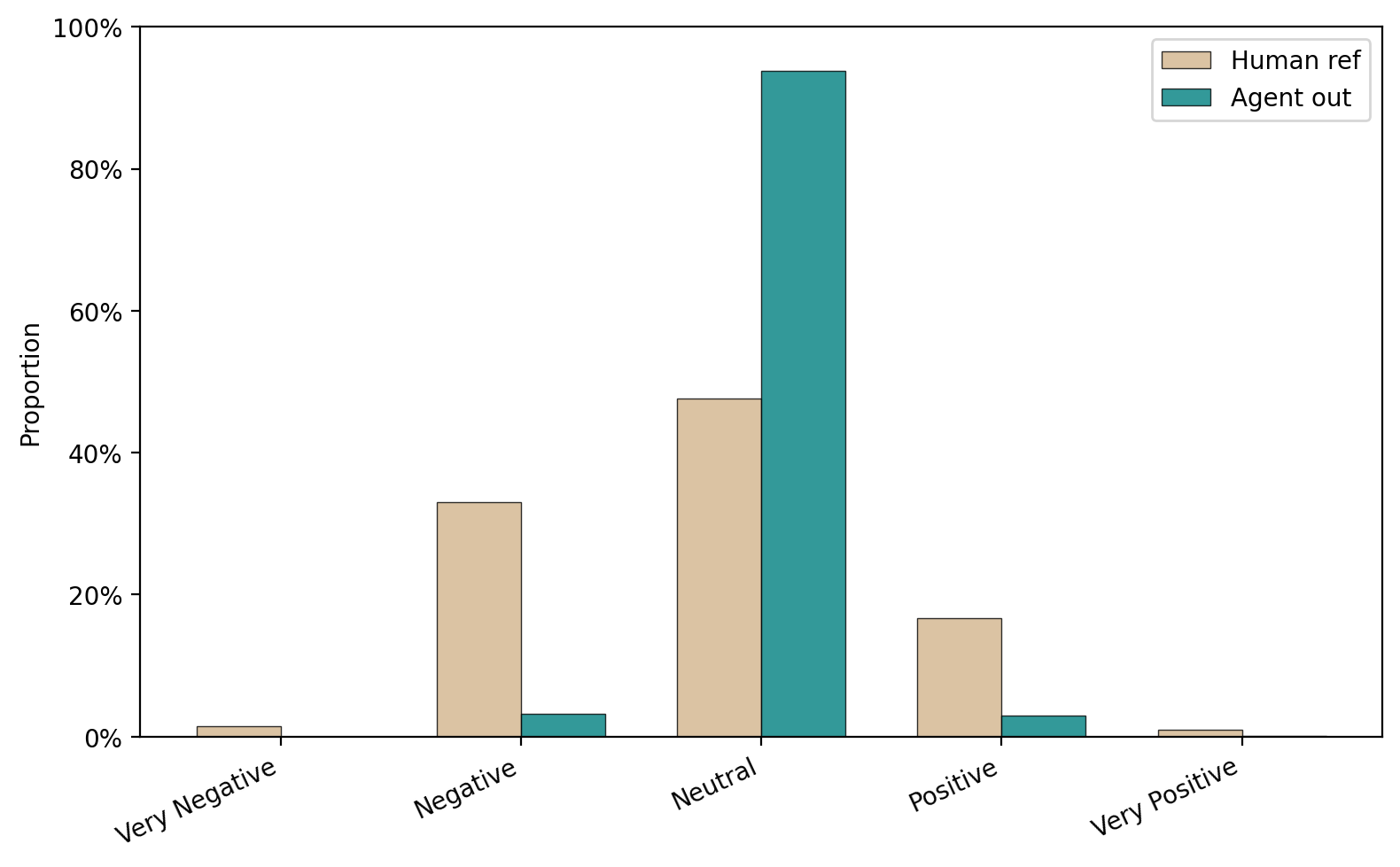}
        \subcaption{\modelB{} under \base{} method.}
    \end{subfigure}

    \par\vspace{1em} 

    \begin{subfigure}[b]{0.45\textwidth}
        \centering
        \includegraphics[width=\textwidth]{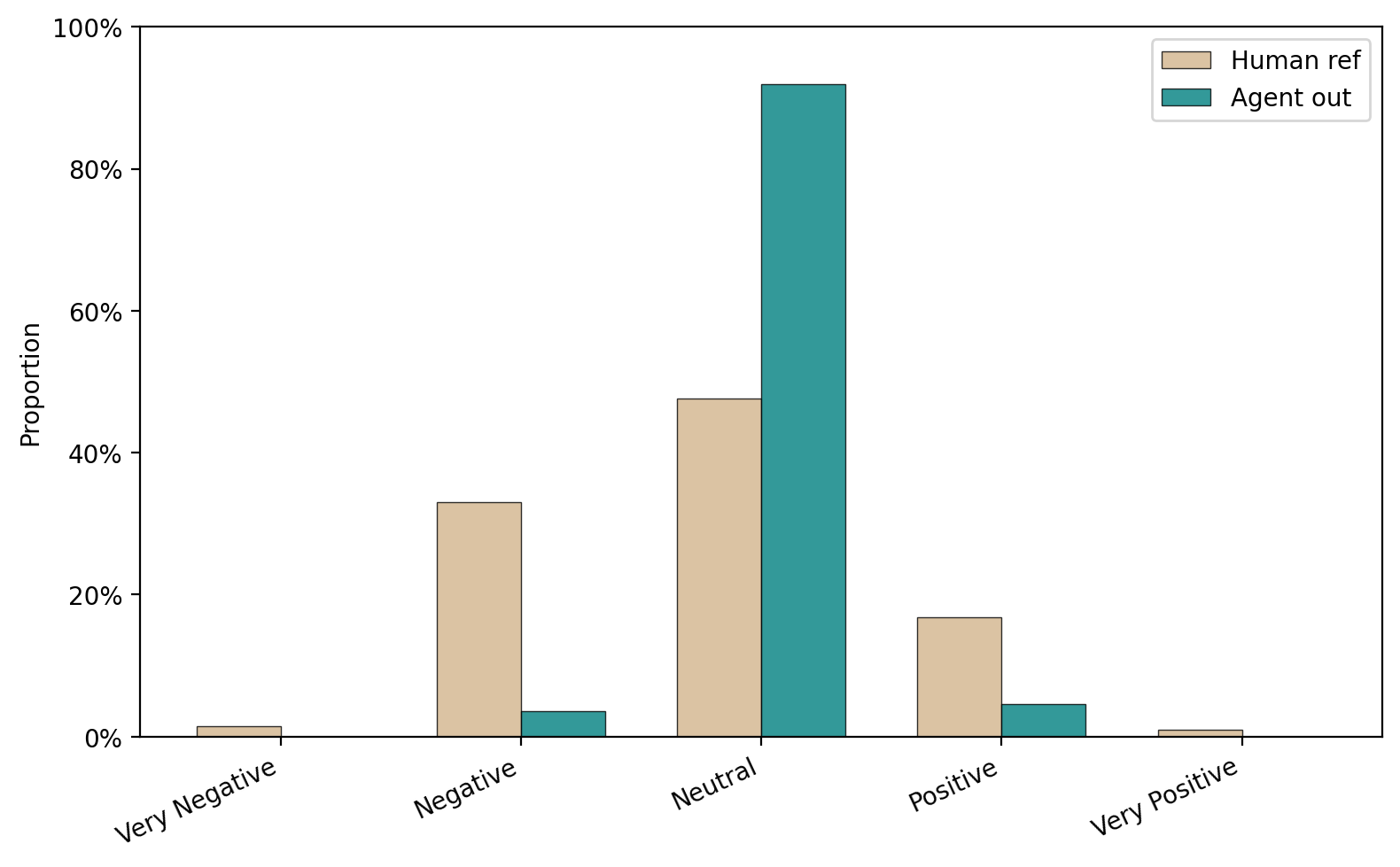}
        \subcaption{\modelB{} under \history{} method.}
    \end{subfigure}

    \caption{Sentiment scores distributions for human agent comments on \texttt{en} for selected models and conditioning methods.}
    \label{fig:sentiment_distributions}
\end{figure*}

Tables~\ref{tab:full_results_overall_1}, \ref{tab:full_results_overall_2} and Tables~\ref{tab:acts_overall_1}, \ref{tab:acts_overall_2} report full results for text similarity and interaction behavior evaluation, respectively. Average BERTScore and ROUGE-L scores are discussed in Section~\ref{sec:results_text_similarity}, while average dialog act and sentiment scores are discussed in Section~\ref{sec:results_interaction_behavior}. Figure~\ref{fig:gender_acts} reports the proportion difference between human and agent dialog acts split by gender, and Figure~\ref{fig:sentiment_distributions} shows the distributions of sentiment scores for human and agent comments, discussed in Section~\ref{sec:results_interaction_behavior}. In the following, we detail text length and act density ratio, which are not discussed in Section~\ref{sec:results}.

\paragraph{Text Length}
Across all conditions, agent-generated comments are substantially longer than human comments. Human comments average approximately 48--78 words depending on the language (\texttt{de} being the shortest at ${\sim}48$ words and \texttt{es} the longest at ${\sim}78$), while agent output consistently exceeds these figures, often by a wide margin. Notably, the \history{} method yields the shortest agent comments across all models---for instance, \modelA{} drops to ${\sim}88$ words on average, its closest approximation to human length---suggesting that exposure to real interaction excerpts helps calibrate verbosity. Conversely, the \behavior{} and \profile{} methods tend to inflate output length, particularly for \modelD{} and \modelB{}. Interestingly, \texttt{es} also elicits slightly longer agent comments than other languages across all conditioning methods, mirroring the fact that human comments in \texttt{es} are themselves the longest.

\paragraph{Act Density Ratio}
The act density ratio is consistently below 1.0 across all conditions, indicating that agent-generated comments are uniformly richer in communicative intents than their human counterparts. The lowest ratios are observed under the \behavior{} method, where agents appear to enact the prescribed behavioral traits, which are dense in prescriptive interactional information. By contrast, when grounded with real examples via \history, the ratio rises above 0.7 on average across all models and languages, with the exception of \modelB{}, suggesting that concrete exemplars help bring act density closer to human levels.

\begin{table*}[ht]
\centering
\scriptsize
\renewcommand{\arraystretch}{0.75}
\setlength{\tabcolsep}{3pt}
\begin{tabular}{lll|cc|ccc}
\toprule
Setup & Lang & Model & Avg Human Words & Avg Agent Words & BERT $\uparrow$ & rBERT $\uparrow$ & ROUGE $\uparrow$ \\
\midrule
\multirow{29}{*}{\base} & \multirow{5}{*}{\texttt{de}} & \modelA{} & 47.9 (56.3) & 80.2 (34.2) & 0.849 & 0.065 & 0.096 \\
&  & \modelB{} & 47.7 (56.3) & 205.1 (74.3) & 0.822 & -0.102 & 0.061 \\
&  & \modelC{} & 47.9 (56.3) & 87.7 (32.5) & 0.847 & 0.051 & 0.088 \\
&  & \modelD{} & 47.9 (56.3) & 121.7 (47.4) & 0.837 & -0.013 & 0.077 \\
& & \textit{Avg models} & \textit{47.8 $\pm$ 0.1} & \textit{123.7 $\pm$ 57.2} & \textit{0.839 $\pm$ 0.012} & \textit{0.000 $\pm$ 0.076} & \textit{0.081 $\pm$ 0.015} \\
\cmidrule(lr){2-8}
& \multirow{5}{*}{\texttt{en}} & \modelA{} & 62.4 (78.7) & 105.4 (54.3) & 0.854 & 0.013 & 0.124 \\
&  & \modelB{} & 62.4 (78.7) & 251.1 (109.2) & 0.832 & -0.140 & 0.085 \\
&  & \modelC{} & 62.4 (78.7) & 105.7 (46.7) & 0.852 & -0.001 & 0.120 \\
&  & \modelD{} & 62.4 (78.7) & 132.2 (54.7) & 0.843 & -0.065 & 0.101 \\
& & \textit{Avg models} & \textit{62.4 $\pm$ 0.0} & \textit{148.6 $\pm$ 69.5} & \textit{0.845 $\pm$ 0.010} & \textit{-0.048 $\pm$ 0.070} & \textit{0.107 $\pm$ 0.018} \\
\cmidrule(lr){2-8}
& \multirow{5}{*}{\texttt{es}} & \modelA{} & 77.5 (96.5) & 113.8 (49.6) & 0.852 & 0.050 & 0.132 \\
&  & \modelB{} & 77.5 (96.8) & 270.0 (92.8) & 0.829 & -0.099 & 0.094 \\
&  & \modelC{} & 77.5 (96.5) & 116.4 (52.1) & 0.851 & 0.039 & 0.129 \\
&  & \modelD{} & 77.5 (96.5) & 152.4 (60.2) & 0.842 & -0.014 & 0.115 \\
& & \textit{Avg models} & \textit{77.5 $\pm$ 0.0} & \textit{163.1 $\pm$ 73.4} & \textit{0.844 $\pm$ 0.011} & \textit{-0.006 $\pm$ 0.068} & \textit{0.118 $\pm$ 0.017} \\
\cmidrule(lr){2-8}
& \multirow{5}{*}{\texttt{fr}} & \modelA{} & 63.9 (77.2) & 100.7 (42.5) & 0.851 & 0.061 & 0.117 \\
&  & \modelB{} & 63.9 (77.2) & 238.0 (82.1) & 0.828 & -0.086 & 0.083 \\
&  & \modelC{} & 63.9 (77.2) & 107.6 (45.6) & 0.849 & 0.048 & 0.110 \\
&  & \modelD{} & 63.9 (77.2) & 141.6 (55.9) & 0.840 & -0.009 & 0.100 \\
& & \textit{Avg models} & \textit{63.9 $\pm$ 0.0} & \textit{147.0 $\pm$ 63.3} & \textit{0.842 $\pm$ 0.011} & \textit{0.004 $\pm$ 0.067} & \textit{0.103 $\pm$ 0.015} \\
\cmidrule(lr){2-8}
& \multirow{5}{*}{\texttt{it}} & \modelA{} & 64.1 (79.1) & 93.5 (42.8) & 0.847 & 0.075 & 0.097 \\
&  & \modelB{} & 64.1 (78.3) & 227.8 (76.8) & 0.824 & -0.065 & 0.069 \\
&  & \modelC{} & 64.1 (79.1) & 91.6 (34.1) & 0.847 & 0.072 & 0.094 \\
&  & \modelD{} & 64.1 (79.1) & 139.2 (56.1) & 0.838 & 0.016 & 0.084 \\
& & \textit{Avg models} & \textit{64.1 $\pm$ 0.0} & \textit{138.0 $\pm$ 63.8} & \textit{0.839 $\pm$ 0.011} & \textit{0.025 $\pm$ 0.066} & \textit{0.086 $\pm$ 0.013} \\
\cmidrule(lr){2-8}
& \multirow{4}{*}{\textit{Avg langs}} & \modelA{} & \textit{63.1 $\pm$ 10.5} & \textit{98.7 $\pm$ 12.7} & \textit{0.851 $\pm$ 0.003} & \textit{0.053 $\pm$ 0.024} & \textit{0.113 $\pm$ 0.016} \\
&  & \modelB{} & \textit{63.1 $\pm$ 10.6} & \textit{238.4 $\pm$ 24.4} & \textit{0.827 $\pm$ 0.004} & \textit{-0.098 $\pm$ 0.027} & \textit{0.078 $\pm$ 0.013} \\
&  & \modelC{} & \textit{63.1 $\pm$ 10.5} & \textit{101.8 $\pm$ 11.9} & \textit{0.849 $\pm$ 0.002} & \textit{0.042 $\pm$ 0.027} & \textit{0.108 $\pm$ 0.017} \\
&  & \modelD{} & \textit{63.1 $\pm$ 10.5} & \textit{137.4 $\pm$ 11.4} & \textit{0.840 $\pm$ 0.003} & \textit{-0.017 $\pm$ 0.030} & \textit{0.096 $\pm$ 0.015} \\
\midrule
\multirow{29}{*}{\profile} & \multirow{5}{*}{\texttt{de}} & \modelA{} & 47.9 (56.3) & 94.8 (38.9) & 0.847 & 0.053 & 0.094 \\
&  & \modelB{} & 47.8 (56.3) & 207.4 (68.9) & 0.821 & -0.112 & 0.060 \\
&  & \modelC{} & 47.9 (56.3) & 98.2 (32.2) & 0.845 & 0.041 & 0.085 \\
&  & \modelD{} & 47.9 (56.3) & 165.6 (62.2) & 0.832 & -0.044 & 0.072 \\
& & \textit{Avg models} & \textit{47.9 $\pm$ 0.0} & \textit{141.5 $\pm$ 54.7} & \textit{0.836 $\pm$ 0.012} & \textit{-0.016 $\pm$ 0.077} & \textit{0.078 $\pm$ 0.015} \\
\cmidrule(lr){2-8}
& \multirow{5}{*}{\texttt{en}} & \modelA{} & 62.4 (78.7) & 122.7 (60.7) & 0.852 & -0.003 & 0.119 \\
&  & \modelB{} & 62.5 (78.9) & 263.5 (99.9) & 0.830 & -0.154 & 0.081 \\
&  & \modelC{} & 62.4 (78.7) & 120.7 (48.8) & 0.850 & -0.017 & 0.115 \\
&  & \modelD{} & 62.4 (78.7) & 180.1 (70.2) & 0.837 & -0.105 & 0.091 \\
& & \textit{Avg models} & \textit{62.4 $\pm$ 0.0} & \textit{171.7 $\pm$ 67.1} & \textit{0.842 $\pm$ 0.011} & \textit{-0.070 $\pm$ 0.072} & \textit{0.101 $\pm$ 0.018} \\
\cmidrule(lr){2-8}
& \multirow{5}{*}{\texttt{es}} & \modelA{} & 77.5 (96.5) & 129.8 (53.8) & 0.851 & 0.041 & 0.130 \\
&  & \modelB{} & 77.5 (96.3) & 272.9 (86.2) & 0.828 & -0.107 & 0.093 \\
&  & \modelC{} & 77.5 (96.5) & 127.5 (52.8) & 0.849 & 0.030 & 0.126 \\
&  & \modelD{} & 77.5 (96.5) & 200.8 (80.7) & 0.838 & -0.043 & 0.107 \\
& & \textit{Avg models} & \textit{77.5 $\pm$ 0.0} & \textit{182.7 $\pm$ 69.1} & \textit{0.841 $\pm$ 0.011} & \textit{-0.020 $\pm$ 0.069} & \textit{0.114 $\pm$ 0.017} \\
\cmidrule(lr){2-8}
& \multirow{5}{*}{\texttt{fr}} & \modelA{} & 63.9 (77.2) & 117.8 (47.1) & 0.849 & 0.047 & 0.114 \\
&  & \modelB{} & 63.9 (77.0) & 242.0 (76.5) & 0.827 & -0.094 & 0.081 \\
&  & \modelC{} & 63.9 (77.2) & 119.5 (46.7) & 0.848 & 0.037 & 0.108 \\
&  & \modelD{} & 63.9 (77.2) & 192.6 (75.8) & 0.835 & -0.042 & 0.092 \\
& & \textit{Avg models} & \textit{63.9 $\pm$ 0.0} & \textit{168.0 $\pm$ 60.4} & \textit{0.840 $\pm$ 0.011} & \textit{-0.013 $\pm$ 0.067} & \textit{0.099 $\pm$ 0.015} \\
\cmidrule(lr){2-8}
& \multirow{5}{*}{\texttt{it}} & \modelA{} & 64.1 (79.1) & 109.6 (47.7) & 0.846 & 0.064 & 0.094 \\
&  & \modelB{} & 64.0 (78.8) & 230.6 (72.8) & 0.823 & -0.072 & 0.067 \\
&  & \modelC{} & 64.1 (79.1) & 98.7 (34.1) & 0.846 & 0.067 & 0.093 \\
&  & \modelD{} & 64.0 (79.1) & 185.6 (73.3) & 0.833 & -0.011 & 0.078 \\
& & \textit{Avg models} & \textit{64.0 $\pm$ 0.0} & \textit{156.1 $\pm$ 62.9} & \textit{0.837 $\pm$ 0.011} & \textit{0.012 $\pm$ 0.067} & \textit{0.083 $\pm$ 0.013} \\
\cmidrule(lr){2-8}
& \multirow{4}{*}{\textit{Avg langs}} & \modelA{} & \textit{63.1 $\pm$ 10.5} & \textit{114.9 $\pm$ 13.5} & \textit{0.849 $\pm$ 0.003} & \textit{0.040 $\pm$ 0.026} & \textit{0.110 $\pm$ 0.016} \\
&  & \modelB{} & \textit{63.1 $\pm$ 10.5} & \textit{243.3 $\pm$ 26.2} & \textit{0.826 $\pm$ 0.004} & \textit{-0.108 $\pm$ 0.030} & \textit{0.076 $\pm$ 0.013} \\
&  & \modelC{} & \textit{63.1 $\pm$ 10.5} & \textit{112.9 $\pm$ 13.6} & \textit{0.848 $\pm$ 0.002} & \textit{0.032 $\pm$ 0.031} & \textit{0.105 $\pm$ 0.017} \\
&  & \modelD{} & \textit{63.1 $\pm$ 10.5} & \textit{184.9 $\pm$ 13.3} & \textit{0.835 $\pm$ 0.003} & \textit{-0.049 $\pm$ 0.034} & \textit{0.088 $\pm$ 0.014} \\
\midrule
\multirow{15}{*}{\behavior} & \multirow{5}{*}{\texttt{de}} & \modelA{} & 47.9 (56.3) & 90.7 (54.9) & 0.847 & 0.050 & 0.091 \\
&  & \modelB{} & 47.7 (55.8) & 148.6 (74.4) & 0.828 & -0.067 & 0.068 \\
&  & \modelC{} & 47.9 (56.3) & 96.9 (37.9) & 0.846 & 0.041 & 0.084 \\
&  & \modelD{} & 47.9 (56.3) & 135.4 (97.0) & 0.837 & -0.013 & 0.078 \\
& & \textit{Avg models} & \textit{47.8 $\pm$ 0.1} & \textit{117.9 $\pm$ 28.5} & \textit{0.839 $\pm$ 0.009} & \textit{0.003 $\pm$ 0.054} & \textit{0.080 $\pm$ 0.010} \\
\cmidrule(lr){2-8}
& \multirow{5}{*}{\texttt{en}} & \modelA{} & 62.4 (78.7) & 155.9 (83.1) & 0.847 & -0.038 & 0.109 \\
&  & \modelB{} & 62.3 (78.7) & 241.8 (109.1) & 0.831 & -0.149 & 0.084 \\
&  & \modelC{} & 62.4 (78.7) & 132.7 (63.3) & 0.849 & -0.027 & 0.112 \\
&  & \modelD{} & 62.4 (78.7) & 212.7 (125.1) & 0.836 & -0.111 & 0.089 \\
& & \textit{Avg models} & \textit{62.4 $\pm$ 0.1} & \textit{185.8 $\pm$ 50.2} & \textit{0.841 $\pm$ 0.009} & \textit{-0.081 $\pm$ 0.059} & \textit{0.098 $\pm$ 0.014} \\
\cmidrule(lr){2-8}
& \multirow{5}{*}{\texttt{es}} & \modelA{} & 77.5 (96.5) & 167.2 (80.5) & 0.845 & 0.002 & 0.117 \\
&  & \modelB{} & 77.4 (96.5) & 230.2 (94.4) & 0.829 & -0.097 & 0.098 \\
&  & \modelC{} & 77.5 (96.5) & 138.9 (63.4) & 0.848 & 0.021 & 0.124 \\
&  & \modelD{} & 77.5 (96.5) & 260.2 (135.5) & 0.834 & -0.065 & 0.099 \\
& & \textit{Avg models} & \textit{77.5 $\pm$ 0.0} & \textit{199.1 $\pm$ 55.8} & \textit{0.839 $\pm$ 0.009} & \textit{-0.035 $\pm$ 0.055} & \textit{0.109 $\pm$ 0.013} \\
\cmidrule(lr){2-8}
\end{tabular}
\caption{Full results (part 1). Rows grouped by setup and language. \textit{Avg Human Words} varies slightly across models and setups due to rare cases where human comments were excluded from evaluation owing to missing generated counterparts (generation errors or hallucinations).}
\label{tab:full_results_overall_1}
\end{table*}

\begin{table*}[ht]
\centering
\scriptsize
\renewcommand{\arraystretch}{0.65}
\begin{tabular}{lll|cc|ccc}
\toprule
Setup & Lang & Model & Avg Human Words & Avg Agent Words & BERT $\uparrow$ & rBERT $\uparrow$ & ROUGE $\uparrow$ \\
\midrule
\multirow{14}{*}{\behavior} & \multirow{5}{*}{\texttt{fr}} & \modelA{} & 63.9 (77.2) & 133.6 (67.5) & 0.846 & 0.024 & 0.107 \\
&  & \modelB{} & 63.7 (77.2) & 199.8 (78.2) & 0.829 & -0.081 & 0.086 \\
&  & \modelC{} & 63.9 (77.2) & 124.6 (54.2) & 0.847 & 0.032 & 0.106 \\
&  & \modelD{} & 63.9 (77.2) & 194.5 (115.4) & 0.836 & -0.037 & 0.092 \\
& & \textit{Avg models} & \textit{63.8 $\pm$ 0.1} & \textit{163.1 $\pm$ 39.6} & \textit{0.839 $\pm$ 0.008} & \textit{-0.016 $\pm$ 0.054} & \textit{0.098 $\pm$ 0.011} \\
\cmidrule(lr){2-8}
& \multirow{5}{*}{\texttt{it}} & \modelA{} & 64.1 (79.1) & 130.8 (70.9) & 0.842 & 0.042 & 0.090 \\
&  & \modelB{} & 63.9 (78.6) & 183.4 (76.2) & 0.826 & -0.056 & 0.072 \\
&  & \modelC{} & 64.1 (79.1) & 104.2 (41.7) & 0.845 & 0.062 & 0.092 \\
&  & \modelD{} & 64.1 (79.1) & 211.9 (126.5) & 0.831 & -0.022 & 0.076 \\
& & \textit{Avg models} & \textit{64.0 $\pm$ 0.1} & \textit{157.6 $\pm$ 49.0} & \textit{0.836 $\pm$ 0.009} & \textit{0.006 $\pm$ 0.055} & \textit{0.082 $\pm$ 0.010} \\
\cmidrule(lr){2-8}
& \multirow{4}{*}{\textit{Avg langs}} & \modelA{} & \textit{63.1 $\pm$ 10.5} & \textit{135.6 $\pm$ 29.4} & \textit{0.845 $\pm$ 0.002} & \textit{0.016 $\pm$ 0.035} & \textit{0.103 $\pm$ 0.012} \\
&  & \modelB{} & \textit{63.0 $\pm$ 10.5} & \textit{200.8 $\pm$ 37.3} & \textit{0.829 $\pm$ 0.002} & \textit{-0.090 $\pm$ 0.036} & \textit{0.082 $\pm$ 0.012} \\
&  & \modelC{} & \textit{63.1 $\pm$ 10.5} & \textit{119.4 $\pm$ 18.2} & \textit{0.847 $\pm$ 0.001} & \textit{0.026 $\pm$ 0.033} & \textit{0.104 $\pm$ 0.016} \\
&  & \modelD{} & \textit{63.1 $\pm$ 10.5} & \textit{203.0 $\pm$ 45.0} & \textit{0.835 $\pm$ 0.002} & \textit{-0.050 $\pm$ 0.040} & \textit{0.087 $\pm$ 0.010} \\
 \midrule
\multirow{29}{*}{\behaviorreasoning} & \multirow{5}{*}{\texttt{de}} & \modelA{} & 47.9 (56.3) & 84.7 (50.8) & 0.853 & 0.085 & 0.112 \\
&  & \modelB{} & 47.6 (56.2) & 156.8 (76.8) & 0.829 & -0.058 & 0.072 \\
&  & \modelC{} & 47.9 (56.3) & 95.8 (36.8) & 0.847 & 0.050 & 0.089 \\
&  & \modelD{} & 47.9 (56.3) & 140.0 (86.5) & 0.839 & -0.001 & 0.085 \\
& & \textit{Avg models} & \textit{47.8 $\pm$ 0.1} & \textit{119.3 $\pm$ 34.6} & \textit{0.842 $\pm$ 0.010} & \textit{0.019 $\pm$ 0.062} & \textit{0.089 $\pm$ 0.017} \\
\cmidrule(lr){2-8}
& \multirow{5}{*}{\texttt{en}} & \modelA{} & 62.4 (78.7) & 127.0 (77.1) & 0.854 & 0.013 & 0.131 \\
&  & \modelB{} & 62.4 (78.0) & 243.8 (114.3) & 0.833 & -0.134 & 0.086 \\
&  & \modelC{} & 62.4 (78.7) & 123.8 (55.8) & 0.850 & -0.015 & 0.116 \\
&  & \modelD{} & 62.4 (78.7) & 190.2 (103.2) & 0.840 & -0.085 & 0.098 \\
& & \textit{Avg models} & \textit{62.4 $\pm$ 0.0} & \textit{171.2 $\pm$ 57.3} & \textit{0.844 $\pm$ 0.010} & \textit{-0.055 $\pm$ 0.067} & \textit{0.108 $\pm$ 0.020} \\
\cmidrule(lr){2-8}
& \multirow{5}{*}{\texttt{es}} & \modelA{} & 77.5 (96.5) & 149.0 (74.5) & 0.852 & 0.046 & 0.138 \\
&  & \modelB{} & 77.2 (96.6) & 246.7 (95.2) & 0.831 & -0.087 & 0.099 \\
&  & \modelC{} & 77.5 (96.5) & 131.6 (59.1) & 0.850 & 0.035 & 0.130 \\
&  & \modelD{} & 77.5 (96.5) & 241.6 (117.5) & 0.838 & -0.040 & 0.109 \\
& & \textit{Avg models} & \textit{77.4 $\pm$ 0.1} & \textit{192.2 $\pm$ 60.4} & \textit{0.843 $\pm$ 0.010} & \textit{-0.011 $\pm$ 0.063} & \textit{0.119 $\pm$ 0.018} \\
\cmidrule(lr){2-8}
& \multirow{5}{*}{\texttt{fr}} & \modelA{} & 63.9 (77.2) & 124.7 (62.5) & 0.850 & 0.051 & 0.120 \\
&  & \modelB{} & 63.6 (76.9) & 216.2 (83.3) & 0.830 & -0.076 & 0.086 \\
&  & \modelC{} & 63.9 (77.2) & 121.4 (51.9) & 0.848 & 0.039 & 0.109 \\
&  & \modelD{} & 63.9 (77.2) & 195.6 (100.4) & 0.837 & -0.027 & 0.097 \\
& & \textit{Avg models} & \textit{63.8 $\pm$ 0.1} & \textit{164.5 $\pm$ 48.6} & \textit{0.841 $\pm$ 0.009} & \textit{-0.003 $\pm$ 0.060} & \textit{0.103 $\pm$ 0.015} \\
\cmidrule(lr){2-8}
& \multirow{5}{*}{\texttt{it}} & \modelA{} & 64.1 (79.1) & 117.5 (67.9) & 0.848 & 0.079 & 0.110 \\
&  & \modelB{} & 63.9 (78.9) & 194.4 (83.0) & 0.827 & -0.047 & 0.073 \\
&  & \modelC{} & 64.1 (79.1) & 101.5 (40.6) & 0.847 & 0.070 & 0.096 \\
&  & \modelD{} & 64.0 (79.1) & 204.5 (113.4) & 0.834 & -0.004 & 0.083 \\
& & \textit{Avg models} & \textit{64.0 $\pm$ 0.1} & \textit{154.5 $\pm$ 52.5} & \textit{0.839 $\pm$ 0.010} & \textit{0.024 $\pm$ 0.060} & \textit{0.091 $\pm$ 0.016} \\
\cmidrule(lr){2-8}
& \multirow{4}{*}{\textit{Avg langs}} & \modelA{} & \textit{63.1 $\pm$ 10.5} & \textit{120.6 $\pm$ 23.2} & \textit{0.851 $\pm$ 0.003} & \textit{0.055 $\pm$ 0.029} & \textit{0.122 $\pm$ 0.012} \\
&  & \modelB{} & \textit{62.9 $\pm$ 10.5} & \textit{211.6 $\pm$ 37.4} & \textit{0.830 $\pm$ 0.002} & \textit{-0.081 $\pm$ 0.034} & \textit{0.083 $\pm$ 0.011} \\
&  & \modelC{} & \textit{63.1 $\pm$ 10.5} & \textit{114.8 $\pm$ 15.4} & \textit{0.848 $\pm$ 0.002} & \textit{0.036 $\pm$ 0.031} & \textit{0.108 $\pm$ 0.016} \\
&  & \modelD{} & \textit{63.1 $\pm$ 10.5} & \textit{194.4 $\pm$ 36.4} & \textit{0.838 $\pm$ 0.002} & \textit{-0.031 $\pm$ 0.034} & \textit{0.094 $\pm$ 0.011} \\
\midrule
\multirow{29}{*}{\texttt{history}} & \multirow{5}{*}{\texttt{de}} & \modelA{} & 47.9 (56.3) & 67.7 (38.3) & 0.856 & 0.105 & 0.114 \\
&  & \modelB{} & 47.0 (55.2) & 154.8 (65.0) & 0.829 & -0.064 & 0.070 \\
&  & \modelC{} & 47.9 (56.3) & 85.3 (31.7) & 0.848 & 0.059 & 0.091 \\
&  & \modelD{} & 47.9 (56.3) & 130.7 (49.6) & 0.837 & -0.013 & 0.079 \\
& & \textit{Avg models} & \textit{47.6 $\pm$ 0.4} & \textit{109.6 $\pm$ 40.1} & \textit{0.842 $\pm$ 0.012} & \textit{0.022 $\pm$ 0.075} & \textit{0.089 $\pm$ 0.019} \\
\cmidrule(lr){2-8}
& \multirow{5}{*}{\texttt{en}} & \modelA{} & 62.4 (78.7) & 96.7 (58.2) & 0.860 & 0.054 & 0.145 \\
&  & \modelB{} & 62.2 (78.1) & 203.8 (96.6) & 0.837 & -0.108 & 0.095 \\
&  & \modelC{} & 62.4 (78.7) & 103.4 (47.6) & 0.854 & 0.012 & 0.127 \\
&  & \modelD{} & 62.4 (78.7) & 147.3 (62.9) & 0.843 & -0.063 & 0.105 \\
& & \textit{Avg models} & \textit{62.4 $\pm$ 0.1} & \textit{137.8 $\pm$ 49.4} & \textit{0.849 $\pm$ 0.011} & \textit{-0.026 $\pm$ 0.073} & \textit{0.118 $\pm$ 0.023} \\
\cmidrule(lr){2-8}
& \multirow{5}{*}{\texttt{es}} & \modelA{} & 77.5 (96.5) & 103.1 (55.9) & 0.859 & 0.091 & 0.154 \\
&  & \modelB{} & 77.3 (96.7) & 210.8 (87.7) & 0.836 & -0.058 & 0.109 \\
&  & \modelC{} & 77.5 (96.5) & 119.3 (51.9) & 0.853 & 0.052 & 0.139 \\
&  & \modelD{} & 77.5 (96.5) & 168.0 (65.7) & 0.843 & -0.008 & 0.119 \\
& & \textit{Avg models} & \textit{77.4 $\pm$ 0.1} & \textit{150.3 $\pm$ 48.9} & \textit{0.848 $\pm$ 0.010} & \textit{0.019 $\pm$ 0.065} & \textit{0.130 $\pm$ 0.020} \\
\cmidrule(lr){2-8}
& \multirow{5}{*}{\texttt{fr}} & \modelA{} & 63.9 (77.2) & 90.0 (49.6) & 0.858 & 0.104 & 0.142 \\
&  & \modelB{} & 63.4 (76.5) & 191.7 (81.2) & 0.834 & -0.050 & 0.095 \\
&  & \modelC{} & 63.9 (77.2) & 105.5 (45.8) & 0.852 & 0.062 & 0.121 \\
&  & \modelD{} & 63.9 (77.2) & 159.3 (62.6) & 0.841 & -0.004 & 0.107 \\
& & \textit{Avg models} & \textit{63.8 $\pm$ 0.2} & \textit{136.6 $\pm$ 47.2} & \textit{0.846 $\pm$ 0.011} & \textit{0.028 $\pm$ 0.068} & \textit{0.116 $\pm$ 0.020} \\
\cmidrule(lr){2-8}
& \multirow{5}{*}{\texttt{it}} & \modelA{} & 64.1 (79.1) & 81.9 (48.5) & 0.852 & 0.103 & 0.108 \\
&  & \modelB{} & 63.0 (78.4) & 173.7 (71.7) & 0.829 & -0.034 & 0.075 \\
&  & \modelC{} & 64.1 (79.1) & 91.5 (36.3) & 0.848 & 0.077 & 0.098 \\
&  & \modelD{} & 64.1 (79.1) & 155.2 (59.8) & 0.837 & 0.014 & 0.085 \\
& & \textit{Avg models} & \textit{63.8 $\pm$ 0.5} & \textit{125.6 $\pm$ 45.7} & \textit{0.842 $\pm$ 0.010} & \textit{0.040 $\pm$ 0.062} & \textit{0.091 $\pm$ 0.015} \\
\cmidrule(lr){2-8}
& \multirow{4}{*}{\textit{Avg langs}} & \modelA{} & \textit{63.1 $\pm$ 10.5} & \textit{87.9 $\pm$ 13.8} & \textit{0.857 $\pm$ 0.003} & \textit{0.091 $\pm$ 0.022} & \textit{0.133 $\pm$ 0.020} \\
&  & \modelB{} & \textit{62.6 $\pm$ 10.7} & \textit{187.0 $\pm$ 22.8} & \textit{0.833 $\pm$ 0.004} & \textit{-0.063 $\pm$ 0.027} & \textit{0.089 $\pm$ 0.016} \\
&  & \modelC{} & \textit{63.1 $\pm$ 10.5} & \textit{101.0 $\pm$ 13.2} & \textit{0.851 $\pm$ 0.003} & \textit{0.053 $\pm$ 0.024} & \textit{0.115 $\pm$ 0.020} \\
&  & \modelD{} & \textit{63.1 $\pm$ 10.5} & \textit{152.1 $\pm$ 14.1} & \textit{0.840 $\pm$ 0.003} & \textit{-0.015 $\pm$ 0.029} & \textit{0.099 $\pm$ 0.017} \\
 \bottomrule
\end{tabular}
\caption{Full results (part 2). Rows grouped by setup and language. \textit{Avg Human Words} varies slightly across models and setups due to rare cases where human comments were excluded from evaluation owing to missing generated counterparts (generation errors or hallucinations).}
\label{tab:full_results_overall_2}
\end{table*}

\begin{table*}[ht]
\centering
\scriptsize
\renewcommand{\arraystretch}{0.75}
\setlength{\tabcolsep}{3pt}
\begin{tabular}{lll|c|cccccccccccc}
\toprule
 & & & Sentiment & \multicolumn{12}{c}{Dialog Acts} \\
 & & & WD $\downarrow$ & JS $\downarrow$ & Density Ratio & CRIT & NEG & POS & PART & INF & REP & COM & REQ & ASK & REF \\
\midrule
\multirow{29}{*}{\base} & \multirow{5}{*}{\texttt{de}} & \modelA{} & 0.236 & 0.147 & 0.700 & 0.002 & 0.051 & -0.027 & -0.009 & 0.042 & 0.040 & 0.005 & -0.107 & 0.003 & 0.000 \\
&  & \modelB{} & 0.565 & 0.270 & 0.549 & 0.018 & 0.077 & 0.034 & 0.007 & 0.078 & 0.057 & -0.017 & -0.165 & 0.036 & -0.123 \\
&  & \modelC{} & 0.329 & 0.259 & 0.733 & 0.028 & 0.073 & -0.061 & -0.021 & 0.069 & 0.081 & -0.027 & -0.151 & -0.019 & 0.029 \\
&  & \modelD{} & 0.459 & 0.207 & 0.733 & -0.010 & 0.035 & -0.000 & -0.000 & -0.020 & 0.066 & -0.028 & -0.100 & 0.069 & -0.013 \\
& & \textit{Avg} & \textit{0.397 $\pm$ 0.145} & \textit{0.221 $\pm$ 0.056} & \textit{0.679 $\pm$ 0.088} & \textit{0.009} & \textit{0.059} & \textit{-0.014} & \textit{-0.006} & \textit{0.042} & \textit{0.061} & \textit{-0.017} & \textit{-0.131} & \textit{0.022} & \textit{-0.027} \\
\cmidrule(lr){2-16}
& \multirow{5}{*}{\texttt{en}} & \modelA{} & 0.188 & 0.114 & 0.750 & -0.001 & 0.042 & -0.011 & -0.020 & 0.022 & 0.048 & -0.004 & -0.056 & 0.001 & -0.021 \\
&  & \modelB{} & 0.486 & 0.247 & 0.591 & 0.010 & 0.066 & 0.038 & -0.001 & 0.058 & 0.067 & -0.053 & -0.138 & 0.044 & -0.091 \\
&  & \modelC{} & 0.171 & 0.193 & 0.764 & 0.012 & 0.056 & -0.030 & -0.046 & 0.040 & 0.066 & 0.003 & -0.128 & 0.009 & 0.017 \\
&  & \modelD{} & 0.415 & 0.198 & 0.715 & -0.010 & 0.035 & 0.007 & -0.007 & -0.011 & 0.065 & -0.034 & -0.074 & 0.069 & -0.040 \\
& & \textit{Avg} & \textit{0.315 $\pm$ 0.159} & \textit{0.188 $\pm$ 0.055} & \textit{0.705 $\pm$ 0.079} & \textit{0.003} & \textit{0.050} & \textit{0.001} & \textit{-0.019} & \textit{0.027} & \textit{0.062} & \textit{-0.022} & \textit{-0.099} & \textit{0.031} & \textit{-0.034} \\
\cmidrule(lr){2-16}
& \multirow{5}{*}{\texttt{es}} & \modelA{} & 0.339 & 0.178 & 0.700 & 0.007 & 0.078 & -0.030 & -0.020 & 0.048 & 0.063 & 0.004 & -0.082 & -0.031 & -0.037 \\
&  & \modelB{} & 0.580 & 0.272 & 0.579 & 0.012 & 0.104 & 0.032 & 0.003 & 0.070 & 0.068 & -0.005 & -0.142 & 0.001 & -0.141 \\
&  & \modelC{} & 0.371 & 0.298 & 0.771 & 0.023 & 0.095 & -0.033 & -0.072 & 0.087 & 0.082 & 0.022 & -0.080 & -0.158 & 0.035 \\
&  & \modelD{} & 0.517 & 0.176 & 0.724 & -0.000 & 0.047 & 0.020 & -0.010 & -0.015 & 0.060 & 0.009 & -0.076 & 0.038 & -0.072 \\
& & \textit{Avg} & \textit{0.452 $\pm$ 0.115} & \textit{0.231 $\pm$ 0.063} & \textit{0.694 $\pm$ 0.082} & \textit{0.010} & \textit{0.081} & \textit{-0.003} & \textit{-0.025} & \textit{0.048} & \textit{0.068} & \textit{0.007} & \textit{-0.095} & \textit{-0.038} & \textit{-0.054} \\
\cmidrule(lr){2-16}
& \multirow{5}{*}{\texttt{fr}} & \modelA{} & 0.239 & 0.156 & 0.699 & 0.006 & 0.049 & -0.028 & -0.007 & 0.048 & 0.064 & -0.017 & -0.087 & -0.015 & -0.012 \\
&  & \modelB{} & 0.501 & 0.277 & 0.570 & 0.014 & 0.078 & 0.030 & 0.012 & 0.084 & 0.076 & -0.037 & -0.162 & 0.025 & -0.120 \\
&  & \modelC{} & 0.252 & 0.263 & 0.743 & 0.034 & 0.076 & -0.074 & -0.048 & 0.081 & 0.084 & -0.022 & -0.108 & -0.052 & 0.029 \\
&  & \modelD{} & 0.404 & 0.217 & 0.694 & 0.001 & 0.040 & 0.006 & -0.005 & 0.001 & 0.074 & -0.028 & -0.101 & 0.059 & -0.047 \\
& & \textit{Avg} & \textit{0.349 $\pm$ 0.126} & \textit{0.228 $\pm$ 0.054} & \textit{0.676 $\pm$ 0.074} & \textit{0.014} & \textit{0.061} & \textit{-0.016} & \textit{-0.012} & \textit{0.053} & \textit{0.074} & \textit{-0.026} & \textit{-0.115} & \textit{0.004} & \textit{-0.037} \\
\cmidrule(lr){2-16}
& \multirow{5}{*}{\texttt{it}} & \modelA{} & 0.219 & 0.155 & 0.710 & -0.006 & 0.054 & -0.055 & 0.002 & 0.043 & 0.053 & -0.000 & -0.093 & 0.017 & -0.014 \\
&  & \modelB{} & 0.559 & 0.296 & 0.566 & 0.005 & 0.089 & 0.015 & 0.009 & 0.090 & 0.063 & -0.015 & -0.153 & 0.051 & -0.154 \\
&  & \modelC{} & 0.205 & 0.258 & 0.768 & 0.018 & 0.083 & -0.110 & -0.048 & 0.074 & 0.067 & 0.009 & -0.144 & 0.019 & 0.032 \\
&  & \modelD{} & 0.462 & 0.198 & 0.693 & -0.010 & 0.039 & 0.001 & 0.006 & -0.005 & 0.051 & 0.003 & -0.088 & 0.064 & -0.060 \\
& & \textit{Avg} & \textit{0.361 $\pm$ 0.177} & \textit{0.227 $\pm$ 0.063} & \textit{0.684 $\pm$ 0.085} & \textit{0.002} & \textit{0.066} & \textit{-0.037} & \textit{-0.008} & \textit{0.050} & \textit{0.058} & \textit{-0.001} & \textit{-0.120} & \textit{0.038} & \textit{-0.049} \\
\cmidrule(lr){2-16}
& \multirow{4}{*}{\textit{Avg}} & \modelA{} & \textit{0.244 $\pm$ 0.057} & \textit{0.150 $\pm$ 0.023} & \textit{0.712 $\pm$ 0.022} & \textit{0.001} & \textit{0.055} & \textit{-0.030} & \textit{-0.011} & \textit{0.041} & \textit{0.053} & \textit{-0.003} & \textit{-0.085} & \textit{-0.005} & \textit{-0.017} \\
&  & \modelB{} & \textit{0.538 $\pm$ 0.042} & \textit{0.272 $\pm$ 0.018} & \textit{0.571 $\pm$ 0.015} & \textit{0.012} & \textit{0.083} & \textit{0.030} & \textit{0.006} & \textit{0.076} & \textit{0.066} & \textit{-0.025} & \textit{-0.152} & \textit{0.031} & \textit{-0.126} \\
&  & \modelC{} & \textit{0.266 $\pm$ 0.084} & \textit{0.254 $\pm$ 0.038} & \textit{0.756 $\pm$ 0.017} & \textit{0.023} & \textit{0.077} & \textit{-0.062} & \textit{-0.047} & \textit{0.070} & \textit{0.076} & \textit{-0.003} & \textit{-0.122} & \textit{-0.040} & \textit{0.029} \\
&  & \modelD{} & \textit{0.451 $\pm$ 0.045} & \textit{0.199 $\pm$ 0.015} & \textit{0.712 $\pm$ 0.018} & \textit{-0.006} & \textit{0.039} & \textit{0.007} & \textit{-0.003} & \textit{-0.010} & \textit{0.063} & \textit{-0.016} & \textit{-0.088} & \textit{0.060} & \textit{-0.046} \\
\midrule
\multirow{29}{*}{\profile} & \multirow{5}{*}{\texttt{de}} & \modelA{} & 0.215 & 0.134 & 0.701 & 0.002 & 0.046 & -0.020 & -0.012 & 0.019 & 0.035 & -0.009 & -0.094 & 0.019 & 0.015 \\
&  & \modelB{} & 0.566 & 0.290 & 0.537 & 0.016 & 0.078 & 0.040 & 0.008 & 0.075 & 0.061 & -0.033 & -0.158 & 0.053 & -0.141 \\
&  & \modelC{} & 0.384 & 0.263 & 0.721 & 0.029 & 0.073 & -0.064 & -0.021 & 0.056 & 0.081 & -0.045 & -0.145 & 0.000 & 0.035 \\
&  & \modelD{} & 0.452 & 0.215 & 0.688 & 0.005 & 0.045 & -0.002 & -0.012 & -0.018 & 0.061 & -0.058 & -0.075 & 0.071 & -0.017 \\
& & \textit{Avg} & \textit{0.404 $\pm$ 0.147} & \textit{0.225 $\pm$ 0.069} & \textit{0.662 $\pm$ 0.084} & \textit{0.013} & \textit{0.061} & \textit{-0.012} & \textit{-0.009} & \textit{0.033} & \textit{0.059} & \textit{-0.036} & \textit{-0.118} & \textit{0.036} & \textit{-0.027} \\
\cmidrule(lr){2-16}
& \multirow{5}{*}{\texttt{en}} & \modelA{} & 0.165 & 0.113 & 0.736 & -0.004 & 0.041 & -0.019 & -0.019 & 0.014 & 0.049 & -0.014 & -0.052 & 0.015 & -0.011 \\
&  & \modelB{} & 0.487 & 0.271 & 0.564 & 0.011 & 0.068 & 0.041 & 0.003 & 0.056 & 0.073 & -0.062 & -0.132 & 0.057 & -0.114 \\
&  & \modelC{} & 0.223 & 0.212 & 0.741 & 0.017 & 0.060 & -0.051 & -0.045 & 0.029 & 0.079 & -0.010 & -0.125 & 0.018 & 0.026 \\
&  & \modelD{} & 0.294 & 0.212 & 0.655 & 0.008 & 0.045 & -0.021 & -0.011 & -0.009 & 0.068 & -0.062 & -0.054 & 0.069 & -0.033 \\
& & \textit{Avg} & \textit{0.292 $\pm$ 0.140} & \textit{0.202 $\pm$ 0.065} & \textit{0.674 $\pm$ 0.083} & \textit{0.008} & \textit{0.053} & \textit{-0.012} & \textit{-0.018} & \textit{0.022} & \textit{0.067} & \textit{-0.037} & \textit{-0.091} & \textit{0.040} & \textit{-0.033} \\
\cmidrule(lr){2-16}
& \multirow{5}{*}{\texttt{es}} & \modelA{} & 0.307 & 0.166 & 0.700 & 0.002 & 0.072 & -0.030 & -0.019 & 0.033 & 0.066 & -0.002 & -0.081 & -0.010 & -0.030 \\
&  & \modelB{} & 0.585 & 0.283 & 0.568 & 0.012 & 0.104 & 0.032 & 0.008 & 0.064 & 0.071 & -0.014 & -0.143 & 0.019 & -0.155 \\
&  & \modelC{} & 0.372 & 0.299 & 0.759 & 0.027 & 0.097 & -0.036 & -0.074 & 0.071 & 0.086 & 0.017 & -0.083 & -0.143 & 0.038 \\
&  & \modelD{} & 0.467 & 0.186 & 0.684 & 0.009 & 0.062 & 0.002 & -0.011 & -0.016 & 0.059 & -0.003 & -0.062 & 0.040 & -0.080 \\
& & \textit{Avg} & \textit{0.433 $\pm$ 0.121} & \textit{0.233 $\pm$ 0.067} & \textit{0.678 $\pm$ 0.080} & \textit{0.013} & \textit{0.084} & \textit{-0.008} & \textit{-0.024} & \textit{0.038} & \textit{0.071} & \textit{-0.000} & \textit{-0.092} & \textit{-0.023} & \textit{-0.057} \\
\cmidrule(lr){2-16}
& \multirow{5}{*}{\texttt{fr}} & \modelA{} & 0.202 & 0.155 & 0.696 & -0.001 & 0.047 & -0.029 & -0.011 & 0.030 & 0.065 & -0.028 & -0.082 & 0.008 & 0.001 \\
&  & \modelB{} & 0.505 & 0.289 & 0.554 & 0.013 & 0.078 & 0.032 & 0.015 & 0.079 & 0.077 & -0.045 & -0.155 & 0.040 & -0.134 \\
&  & \modelC{} & 0.247 & 0.264 & 0.739 & 0.035 & 0.077 & -0.079 & -0.047 & 0.068 & 0.085 & -0.029 & -0.104 & -0.039 & 0.033 \\
&  & \modelD{} & 0.323 & 0.228 & 0.652 & 0.010 & 0.051 & -0.002 & -0.012 & 0.003 & 0.073 & -0.057 & -0.081 & 0.060 & -0.044 \\
& & \textit{Avg} & \textit{0.319 $\pm$ 0.134} & \textit{0.234 $\pm$ 0.058} & \textit{0.660 $\pm$ 0.079} & \textit{0.014} & \textit{0.063} & \textit{-0.020} & \textit{-0.014} & \textit{0.045} & \textit{0.075} & \textit{-0.040} & \textit{-0.105} & \textit{0.017} & \textit{-0.036} \\
\cmidrule(lr){2-16}
& \multirow{5}{*}{\texttt{it}} & \modelA{} & 0.224 & 0.166 & 0.698 & -0.007 & 0.054 & -0.057 & 0.000 & 0.031 & 0.055 & -0.009 & -0.093 & 0.035 & -0.010 \\
&  & \modelB{} & 0.560 & 0.308 & 0.556 & 0.001 & 0.089 & 0.022 & 0.016 & 0.086 & 0.064 & -0.019 & -0.148 & 0.057 & -0.167 \\
&  & \modelC{} & 0.209 & 0.270 & 0.758 & 0.021 & 0.086 & -0.114 & -0.052 & 0.065 & 0.069 & 0.004 & -0.144 & 0.029 & 0.037 \\
&  & \modelD{} & 0.406 & 0.191 & 0.659 & -0.001 & 0.046 & -0.007 & 0.003 & -0.006 & 0.043 & -0.010 & -0.066 & 0.064 & -0.066 \\
& & \textit{Avg} & \textit{0.350 $\pm$ 0.166} & \textit{0.234 $\pm$ 0.066} & \textit{0.668 $\pm$ 0.085} & \textit{0.003} & \textit{0.069} & \textit{-0.039} & \textit{-0.008} & \textit{0.044} & \textit{0.058} & \textit{-0.009} & \textit{-0.113} & \textit{0.046} & \textit{-0.052} \\
\cmidrule(lr){2-16}
& \multirow{4}{*}{\textit{Avg}} & \modelA{} & \textit{0.223 $\pm$ 0.052} & \textit{0.147 $\pm$ 0.023} & \textit{0.706 $\pm$ 0.017} & \textit{-0.002} & \textit{0.052} & \textit{-0.031} & \textit{-0.012} & \textit{0.025} & \textit{0.054} & \textit{-0.012} & \textit{-0.080} & \textit{0.013} & \textit{-0.007} \\
&  & \modelB{} & \textit{0.541 $\pm$ 0.042} & \textit{0.288 $\pm$ 0.013} & \textit{0.556 $\pm$ 0.012} & \textit{0.010} & \textit{0.084} & \textit{0.033} & \textit{0.010} & \textit{0.072} & \textit{0.069} & \textit{-0.035} & \textit{-0.147} & \textit{0.045} & \textit{-0.142} \\
&  & \modelC{} & \textit{0.287 $\pm$ 0.084} & \textit{0.262 $\pm$ 0.031} & \textit{0.744 $\pm$ 0.016} & \textit{0.026} & \textit{0.079} & \textit{-0.069} & \textit{-0.048} & \textit{0.058} & \textit{0.080} & \textit{-0.013} & \textit{-0.120} & \textit{-0.027} & \textit{0.034} \\
&  & \modelD{} & \textit{0.389 $\pm$ 0.077} & \textit{0.206 $\pm$ 0.017} & \textit{0.668 $\pm$ 0.017} & \textit{0.006} & \textit{0.050} & \textit{-0.006} & \textit{-0.009} & \textit{-0.009} & \textit{0.061} & \textit{-0.038} & \textit{-0.068} & \textit{0.061} & \textit{-0.048} \\
\midrule
\multirow{15}{*}{\behavior} & \multirow{5}{*}{\texttt{de}} & \modelA{} & 0.257 & 0.227 & 0.644 & 0.004 & 0.055 & -0.012 & -0.002 & 0.133 & 0.060 & 0.011 & -0.142 & -0.094 & -0.013 \\
&  & \modelB{} & 0.551 & 0.246 & 0.493 & 0.010 & 0.073 & 0.035 & 0.023 & 0.141 & -0.072 & -0.003 & -0.129 & 0.012 & -0.090 \\
&  & \modelC{} & 0.360 & 0.281 & 0.687 & 0.030 & 0.074 & -0.059 & -0.021 & 0.112 & 0.084 & -0.015 & -0.177 & -0.045 & 0.017 \\
&  & \modelD{} & 0.389 & 0.233 & 0.536 & -0.002 & 0.059 & 0.009 & 0.007 & 0.114 & 0.038 & -0.091 & -0.125 & 0.046 & -0.056 \\
& & \textit{Avg} & \textit{0.389 $\pm$ 0.122} & \textit{0.247 $\pm$ 0.024} & \textit{0.590 $\pm$ 0.091} & \textit{0.011} & \textit{0.065} & \textit{-0.007} & \textit{0.002} & \textit{0.125} & \textit{0.027} & \textit{-0.024} & \textit{-0.143} & \textit{-0.020} & \textit{-0.036} \\
\cmidrule(lr){2-16}
& \multirow{5}{*}{\texttt{en}} & \modelA{} & 0.278 & 0.181 & 0.632 & -0.007 & 0.045 & 0.012 & -0.006 & 0.087 & 0.070 & 0.004 & -0.091 & -0.049 & -0.064 \\
&  & \modelB{} & 0.510 & 0.229 & 0.510 & 0.006 & 0.063 & 0.052 & 0.020 & 0.096 & -0.025 & -0.030 & -0.098 & 0.038 & -0.123 \\
&  & \modelC{} & 0.259 & 0.233 & 0.675 & 0.016 & 0.058 & -0.023 & -0.037 & 0.081 & 0.087 & 0.010 & -0.166 & 0.000 & -0.027 \\
&  & \modelD{} & 0.399 & 0.249 & 0.518 & -0.004 & 0.052 & 0.024 & 0.008 & 0.089 & 0.054 & -0.084 & -0.091 & 0.063 & -0.112 \\
& & \textit{Avg} & \textit{0.361 $\pm$ 0.117} & \textit{0.223 $\pm$ 0.029} & \textit{0.584 $\pm$ 0.083} & \textit{0.003} & \textit{0.055} & \textit{0.016} & \textit{-0.004} & \textit{0.088} & \textit{0.047} & \textit{-0.025} & \textit{-0.111} & \textit{0.013} & \textit{-0.081} \\
\cmidrule(lr){2-16}
& \multirow{5}{*}{\texttt{es}} & \modelA{} & 0.457 & 0.235 & 0.598 & 0.008 & 0.079 & 0.010 & -0.015 & 0.100 & 0.073 & 0.007 & -0.113 & -0.060 & -0.090 \\
&  & \modelB{} & 0.584 & 0.241 & 0.536 & -0.000 & 0.096 & 0.040 & 0.028 & 0.094 & -0.012 & -0.009 & -0.102 & 0.012 & -0.146 \\
&  & \modelC{} & 0.413 & 0.299 & 0.713 & 0.028 & 0.097 & -0.020 & -0.065 & 0.097 & 0.086 & 0.021 & -0.112 & -0.137 & 0.005 \\
&  & \modelD{} & 0.536 & 0.234 & 0.529 & -0.000 & 0.080 & 0.036 & 0.009 & 0.088 & 0.052 & -0.032 & -0.119 & 0.021 & -0.134 \\
& & \textit{Avg} & \textit{0.498 $\pm$ 0.077} & \textit{0.253 $\pm$ 0.031} & \textit{0.594 $\pm$ 0.085} & \textit{0.009} & \textit{0.088} & \textit{0.017} & \textit{-0.011} & \textit{0.095} & \textit{0.050} & \textit{-0.003} & \textit{-0.112} & \textit{-0.041} & \textit{-0.091} \\
\bottomrule
\end{tabular}
\caption{Dialog acts (part 1). JS = Jensen-Shannon distance on label proportions. Density Ratio = human labels/sentence $\div$ agent labels/sentence. Dialog Acts = human proportion $-$ agent proportion for each label (positive $\Rightarrow$ agent under-produces).}
\label{tab:acts_overall_1}
\end{table*}

\begin{table*}[ht]
\centering
\scriptsize
\renewcommand{\arraystretch}{0.75}
\setlength{\tabcolsep}{3pt}
\begin{tabular}{lll|c|cccccccccccc}
\toprule
 & & & Sentiment & \multicolumn{12}{c}{Dialog Acts} \\
 & & & WD $\downarrow$ & JS $\downarrow$ & Density Ratio & CRIT & NEG & POS & PART & INF & REP & COM & REQ & ASK & REF \\
\midrule
\multirow{14}{*}{\behavior} & \multirow{5}{*}{\texttt{fr}} & \modelA{} & 0.358 & 0.222 & 0.615 & 0.007 & 0.055 & -0.004 & -0.001 & 0.115 & 0.075 & -0.007 & -0.120 & -0.069 & -0.051 \\
&  & \modelB{} & 0.508 & 0.246 & 0.511 & 0.007 & 0.073 & 0.035 & 0.030 & 0.125 & -0.011 & -0.040 & -0.124 & 0.025 & -0.120 \\
&  & \modelC{} & 0.323 & 0.280 & 0.689 & 0.038 & 0.076 & -0.057 & -0.050 & 0.106 & 0.088 & -0.012 & -0.127 & -0.070 & 0.009 \\
&  & \modelD{} & 0.457 & 0.291 & 0.524 & 0.004 & 0.064 & 0.034 & 0.008 & 0.114 & 0.068 & -0.108 & -0.133 & 0.054 & -0.106 \\
& & \textit{Avg} & \textit{0.411 $\pm$ 0.086} & \textit{0.260 $\pm$ 0.031} & \textit{0.585 $\pm$ 0.084} & \textit{0.014} & \textit{0.067} & \textit{0.002} & \textit{-0.003} & \textit{0.115} & \textit{0.055} & \textit{-0.042} & \textit{-0.126} & \textit{-0.015} & \textit{-0.067} \\
\cmidrule(lr){2-16}
& \multirow{5}{*}{\texttt{it}} & \modelA{} & 0.358 & 0.204 & 0.609 & -0.006 & 0.061 & -0.019 & 0.006 & 0.112 & 0.060 & -0.007 & -0.123 & -0.019 & -0.064 \\
&  & \modelB{} & 0.556 & 0.266 & 0.512 & -0.009 & 0.084 & 0.039 & 0.037 & 0.127 & -0.028 & -0.028 & -0.113 & 0.036 & -0.146 \\
&  & \modelC{} & 0.285 & 0.277 & 0.717 & 0.021 & 0.086 & -0.095 & -0.053 & 0.102 & 0.072 & 0.012 & -0.169 & 0.013 & 0.012 \\
&  & \modelD{} & 0.484 & 0.259 & 0.501 & -0.010 & 0.069 & 0.026 & 0.021 & 0.105 & 0.030 & -0.065 & -0.109 & 0.059 & -0.125 \\
& & \textit{Avg} & \textit{0.421 $\pm$ 0.122} & \textit{0.252 $\pm$ 0.032} & \textit{0.585 $\pm$ 0.100} & \textit{-0.001} & \textit{0.075} & \textit{-0.012} & \textit{0.003} & \textit{0.112} & \textit{0.034} & \textit{-0.022} & \textit{-0.129} & \textit{0.022} & \textit{-0.081} \\
\cmidrule(lr){2-16}
& \multirow{4}{*}{\textit{Avg}} & \modelA{} & \textit{0.342 $\pm$ 0.079} & \textit{0.214 $\pm$ 0.022} & \textit{0.620 $\pm$ 0.018} & \textit{0.001} & \textit{0.059} & \textit{-0.003} & \textit{-0.003} & \textit{0.109} & \textit{0.068} & \textit{0.002} & \textit{-0.118} & \textit{-0.058} & \textit{-0.056} \\
&  & \modelB{} & \textit{0.542 $\pm$ 0.033} & \textit{0.245 $\pm$ 0.014} & \textit{0.512 $\pm$ 0.016} & \textit{0.003} & \textit{0.078} & \textit{0.040} & \textit{0.028} & \textit{0.117} & \textit{-0.030} & \textit{-0.022} & \textit{-0.113} & \textit{0.025} & \textit{-0.125} \\
&  & \modelC{} & \textit{0.328 $\pm$ 0.061} & \textit{0.274 $\pm$ 0.025} & \textit{0.696 $\pm$ 0.018} & \textit{0.027} & \textit{0.078} & \textit{-0.051} & \textit{-0.045} & \textit{0.099} & \textit{0.083} & \textit{0.003} & \textit{-0.150} & \textit{-0.048} & \textit{0.003} \\
&  & \modelD{} & \textit{0.453 $\pm$ 0.061} & \textit{0.253 $\pm$ 0.024} & \textit{0.522 $\pm$ 0.013} & \textit{-0.002} & \textit{0.065} & \textit{0.026} & \textit{0.011} & \textit{0.102} & \textit{0.048} & \textit{-0.076} & \textit{-0.115} & \textit{0.049} & \textit{-0.107} \\
\midrule
\multirow{29}{*}{\behaviorreasoning} & \multirow{5}{*}{\texttt{de}} & \modelA{} & 0.155 & 0.160 & 0.701 & -0.002 & 0.042 & -0.021 & -0.002 & 0.084 & 0.032 & 0.004 & -0.096 & -0.069 & 0.027 \\
&  & \modelB{} & 0.534 & 0.225 & 0.524 & 0.004 & 0.070 & 0.025 & 0.021 & 0.122 & -0.002 & -0.030 & -0.134 & 0.010 & -0.086 \\
&  & \modelC{} & 0.343 & 0.266 & 0.686 & 0.025 & 0.067 & -0.066 & -0.021 & 0.110 & 0.079 & -0.020 & -0.164 & -0.050 & 0.038 \\
&  & \modelD{} & 0.289 & 0.150 & 0.583 & 0.004 & 0.044 & -0.014 & 0.005 & 0.084 & -0.014 & -0.038 & -0.097 & 0.030 & -0.004 \\
& & \textit{Avg} & \textit{0.330 $\pm$ 0.157} & \textit{0.200 $\pm$ 0.055} & \textit{0.623 $\pm$ 0.085} & \textit{0.008} & \textit{0.056} & \textit{-0.019} & \textit{0.001} & \textit{0.100} & \textit{0.024} & \textit{-0.021} & \textit{-0.123} & \textit{-0.020} & \textit{-0.006} \\
\cmidrule(lr){2-16}
& \multirow{5}{*}{\texttt{en}} & \modelA{} & 0.168 & 0.111 & 0.705 & -0.005 & 0.032 & -0.004 & -0.004 & 0.059 & 0.041 & -0.003 & -0.056 & -0.040 & -0.019 \\
&  & \modelB{} & 0.477 & 0.228 & 0.532 & 0.004 & 0.056 & 0.038 & 0.017 & 0.094 & 0.037 & -0.060 & -0.109 & 0.038 & -0.114 \\
&  & \modelC{} & 0.191 & 0.208 & 0.686 & 0.015 & 0.049 & -0.042 & -0.030 & 0.086 & 0.083 & 0.006 & -0.142 & -0.017 & -0.007 \\
&  & \modelD{} & 0.265 & 0.151 & 0.565 & 0.002 & 0.036 & -0.008 & 0.003 & 0.070 & 0.008 & -0.048 & -0.061 & 0.046 & -0.048 \\
& & \textit{Avg} & \textit{0.275 $\pm$ 0.141} & \textit{0.175 $\pm$ 0.053} & \textit{0.622 $\pm$ 0.086} & \textit{0.004} & \textit{0.043} & \textit{-0.004} & \textit{-0.004} & \textit{0.077} & \textit{0.042} & \textit{-0.026} & \textit{-0.092} & \textit{0.007} & \textit{-0.047} \\
\cmidrule(lr){2-16}
& \multirow{5}{*}{\texttt{es}} & \modelA{} & 0.338 & 0.157 & 0.673 & 0.004 & 0.059 & -0.015 & -0.004 & 0.063 & 0.054 & -0.000 & -0.086 & -0.041 & -0.034 \\
&  & \modelB{} & 0.585 & 0.238 & 0.562 & 0.002 & 0.083 & 0.032 & 0.023 & 0.088 & 0.047 & -0.032 & -0.121 & 0.017 & -0.138 \\
&  & \modelC{} & 0.349 & 0.275 & 0.721 & 0.022 & 0.081 & -0.037 & -0.050 & 0.100 & 0.084 & 0.019 & -0.110 & -0.131 & 0.022 \\
&  & \modelD{} & 0.449 & 0.161 & 0.564 & 0.003 & 0.058 & 0.014 & 0.007 & 0.069 & 0.018 & -0.017 & -0.085 & 0.021 & -0.088 \\
& & \textit{Avg} & \textit{0.430 $\pm$ 0.114} & \textit{0.208 $\pm$ 0.058} & \textit{0.630 $\pm$ 0.080} & \textit{0.008} & \textit{0.070} & \textit{-0.001} & \textit{-0.006} & \textit{0.080} & \textit{0.050} & \textit{-0.008} & \textit{-0.100} & \textit{-0.034} & \textit{-0.059} \\
\cmidrule(lr){2-16}
& \multirow{5}{*}{\texttt{fr}} & \modelA{} & 0.240 & 0.161 & 0.671 & 0.001 & 0.043 & -0.021 & -0.000 & 0.076 & 0.055 & -0.012 & -0.090 & -0.058 & 0.006 \\
&  & \modelB{} & 0.499 & 0.249 & 0.535 & 0.007 & 0.067 & 0.029 & 0.029 & 0.110 & 0.047 & -0.057 & -0.139 & 0.021 & -0.113 \\
&  & \modelC{} & 0.242 & 0.268 & 0.699 & 0.033 & 0.069 & -0.062 & -0.043 & 0.107 & 0.085 & -0.016 & -0.119 & -0.084 & 0.031 \\
&  & \modelD{} & 0.353 & 0.188 & 0.556 & 0.008 & 0.048 & 0.004 & 0.005 & 0.086 & 0.024 & -0.056 & -0.112 & 0.040 & -0.046 \\
& & \textit{Avg} & \textit{0.333 $\pm$ 0.122} & \textit{0.216 $\pm$ 0.050} & \textit{0.615 $\pm$ 0.082} & \textit{0.012} & \textit{0.057} & \textit{-0.013} & \textit{-0.002} & \textit{0.095} & \textit{0.053} & \textit{-0.035} & \textit{-0.115} & \textit{-0.020} & \textit{-0.031} \\
\cmidrule(lr){2-16}
& \multirow{5}{*}{\texttt{it}} & \modelA{} & 0.229 & 0.142 & 0.685 & -0.008 & 0.046 & -0.034 & 0.006 & 0.067 & 0.046 & -0.006 & -0.097 & -0.016 & -0.004 \\
&  & \modelB{} & 0.542 & 0.267 & 0.542 & -0.013 & 0.078 & 0.037 & 0.029 & 0.116 & 0.037 & -0.051 & -0.130 & 0.039 & -0.142 \\
&  & \modelC{} & 0.194 & 0.255 & 0.722 & 0.017 & 0.074 & -0.097 & -0.040 & 0.106 & 0.069 & 0.007 & -0.159 & -0.012 & 0.034 \\
&  & \modelD{} & 0.391 & 0.180 & 0.528 & -0.007 & 0.054 & 0.005 & 0.014 & 0.086 & -0.006 & -0.035 & -0.084 & 0.046 & -0.073 \\
& & \textit{Avg} & \textit{0.339 $\pm$ 0.160} & \textit{0.211 $\pm$ 0.060} & \textit{0.619 $\pm$ 0.099} & \textit{-0.003} & \textit{0.063} & \textit{-0.022} & \textit{0.002} & \textit{0.094} & \textit{0.037} & \textit{-0.021} & \textit{-0.118} & \textit{0.014} & \textit{-0.046} \\
\cmidrule(lr){2-16}
& \multirow{4}{*}{\textit{Avg}} & \modelA{} & \textit{0.226 $\pm$ 0.073} & \textit{0.146 $\pm$ 0.021} & \textit{0.687 $\pm$ 0.016} & \textit{-0.002} & \textit{0.044} & \textit{-0.019} & \textit{-0.001} & \textit{0.070} & \textit{0.046} & \textit{-0.003} & \textit{-0.085} & \textit{-0.045} & \textit{-0.005} \\
&  & \modelB{} & \textit{0.527 $\pm$ 0.042} & \textit{0.241 $\pm$ 0.017} & \textit{0.539 $\pm$ 0.014} & \textit{0.001} & \textit{0.071} & \textit{0.032} & \textit{0.024} & \textit{0.106} & \textit{0.033} & \textit{-0.046} & \textit{-0.127} & \textit{0.025} & \textit{-0.119} \\
&  & \modelC{} & \textit{0.264 $\pm$ 0.078} & \textit{0.254 $\pm$ 0.027} & \textit{0.703 $\pm$ 0.018} & \textit{0.023} & \textit{0.068} & \textit{-0.061} & \textit{-0.037} & \textit{0.102} & \textit{0.080} & \textit{-0.001} & \textit{-0.139} & \textit{-0.059} & \textit{0.024} \\
&  & \modelD{} & \textit{0.349 $\pm$ 0.075} & \textit{0.166 $\pm$ 0.017} & \textit{0.559 $\pm$ 0.020} & \textit{0.002} & \textit{0.048} & \textit{0.000} & \textit{0.007} & \textit{0.079} & \textit{0.006} & \textit{-0.039} & \textit{-0.088} & \textit{0.037} & \textit{-0.052} \\
\midrule
\multirow{29}{*}{\history} & \multirow{5}{*}{\texttt{de}} & \modelA{} & 0.136 & 0.124 & 0.820 & 0.011 & 0.030 & -0.024 & -0.012 & 0.005 & -0.003 & -0.004 & -0.074 & 0.028 & 0.044 \\
&  & \modelB{} & 0.558 & 0.261 & 0.592 & 0.019 & 0.075 & 0.031 & 0.004 & 0.076 & 0.057 & -0.036 & -0.151 & 0.037 & -0.112 \\
&  & \modelC{} & 0.316 & 0.260 & 0.791 & 0.026 & 0.062 & -0.027 & -0.042 & 0.027 & 0.078 & -0.036 & -0.150 & -0.001 & 0.063 \\
&  & \modelD{} & 0.416 & 0.204 & 0.736 & -0.007 & 0.022 & 0.003 & -0.009 & -0.030 & 0.067 & -0.040 & -0.081 & 0.070 & 0.005 \\
& & \textit{Avg} & \textit{0.356 $\pm$ 0.177} & \textit{0.212 $\pm$ 0.065} & \textit{0.735 $\pm$ 0.101} & \textit{0.012} & \textit{0.047} & \textit{-0.004} & \textit{-0.015} & \textit{0.019} & \textit{0.050} & \textit{-0.029} & \textit{-0.114} & \textit{0.034} & \textit{0.000} \\
\cmidrule(lr){2-16}
& \multirow{5}{*}{\texttt{en}} & \modelA{} & 0.140 & 0.085 & 0.816 & -0.002 & 0.022 & -0.011 & -0.013 & -0.001 & 0.027 & -0.011 & -0.049 & 0.023 & 0.016 \\
&  & \modelB{} & 0.467 & 0.244 & 0.615 & 0.021 & 0.064 & 0.019 & -0.002 & 0.062 & 0.061 & -0.076 & -0.122 & 0.051 & -0.078 \\
&  & \modelC{} & 0.097 & 0.200 & 0.819 & 0.008 & 0.044 & -0.004 & -0.062 & 0.008 & 0.058 & 0.002 & -0.123 & 0.011 & 0.057 \\
&  & \modelD{} & 0.375 & 0.190 & 0.705 & -0.006 & 0.025 & -0.004 & -0.013 & -0.014 & 0.065 & -0.045 & -0.061 & 0.069 & -0.016 \\
& & \textit{Avg} & \textit{0.270 $\pm$ 0.180} & \textit{0.180 $\pm$ 0.068} & \textit{0.739 $\pm$ 0.098} & \textit{0.005} & \textit{0.039} & \textit{0.000} & \textit{-0.022} & \textit{0.014} & \textit{0.053} & \textit{-0.033} & \textit{-0.089} & \textit{0.038} & \textit{-0.005} \\
\cmidrule(lr){2-16}
& \multirow{5}{*}{\texttt{es}} & \modelA{} & 0.237 & 0.127 & 0.814 & 0.010 & 0.048 & -0.051 & -0.015 & 0.007 & 0.046 & -0.005 & -0.072 & 0.016 & 0.017 \\
&  & \modelB{} & 0.571 & 0.254 & 0.623 & 0.020 & 0.098 & 0.025 & -0.000 & 0.072 & 0.060 & -0.019 & -0.135 & 0.012 & -0.131 \\
&  & \modelC{} & 0.323 & 0.276 & 0.801 & 0.017 & 0.077 & 0.005 & -0.091 & 0.047 & 0.079 & 0.018 & -0.086 & -0.123 & 0.057 \\
&  & \modelD{} & 0.492 & 0.159 & 0.735 & 0.002 & 0.030 & 0.016 & -0.019 & -0.028 & 0.058 & 0.004 & -0.057 & 0.041 & -0.046 \\
& & \textit{Avg} & \textit{0.406 $\pm$ 0.153} & \textit{0.204 $\pm$ 0.072} & \textit{0.743 $\pm$ 0.087} & \textit{0.012} & \textit{0.063} & \textit{-0.001} & \textit{-0.031} & \textit{0.025} & \textit{0.061} & \textit{-0.001} & \textit{-0.087} & \textit{-0.014} & \textit{-0.026} \\
\cmidrule(lr){2-16}
& \multirow{5}{*}{\texttt{fr}} & \modelA{} & 0.164 & 0.125 & 0.795 & 0.004 & 0.030 & -0.039 & -0.008 & 0.016 & 0.041 & -0.022 & -0.072 & 0.019 & 0.032 \\
&  & \modelB{} & 0.492 & 0.262 & 0.604 & 0.021 & 0.076 & 0.017 & 0.013 & 0.085 & 0.065 & -0.053 & -0.150 & 0.030 & -0.104 \\
&  & \modelC{} & 0.201 & 0.263 & 0.782 & 0.028 & 0.065 & -0.044 & -0.068 & 0.049 & 0.080 & -0.034 & -0.101 & -0.043 & 0.069 \\
&  & \modelD{} & 0.379 & 0.195 & 0.711 & 0.002 & 0.022 & 0.007 & -0.011 & -0.017 & 0.066 & -0.038 & -0.081 & 0.059 & -0.010 \\
& & \textit{Avg} & \textit{0.309 $\pm$ 0.154} & \textit{0.211 $\pm$ 0.066} & \textit{0.723 $\pm$ 0.088} & \textit{0.014} & \textit{0.048} & \textit{-0.015} & \textit{-0.019} & \textit{0.033} & \textit{0.063} & \textit{-0.037} & \textit{-0.101} & \textit{0.016} & \textit{-0.003} \\
\cmidrule(lr){2-16}
& \multirow{5}{*}{\texttt{it}} & \modelA{} & 0.152 & 0.137 & 0.814 & -0.001 & 0.028 & -0.057 & -0.002 & 0.006 & 0.042 & -0.003 & -0.080 & 0.035 & 0.030 \\
&  & \modelB{} & 0.550 & 0.283 & 0.599 & 0.013 & 0.087 & 0.007 & 0.004 & 0.092 & 0.059 & -0.031 & -0.143 & 0.050 & -0.138 \\
&  & \modelC{} & 0.166 & 0.263 & 0.808 & 0.017 & 0.072 & -0.062 & -0.084 & 0.045 & 0.067 & 0.010 & -0.150 & 0.027 & 0.058 \\
&  & \modelD{} & 0.448 & 0.182 & 0.700 & -0.012 & 0.020 & 0.004 & -0.001 & -0.019 & 0.053 & -0.004 & -0.078 & 0.064 & -0.027 \\
& & \textit{Avg} & \textit{0.329 $\pm$ 0.201} & \textit{0.217 $\pm$ 0.069} & \textit{0.730 $\pm$ 0.102} & \textit{0.004} & \textit{0.052} & \textit{-0.027} & \textit{-0.021} & \textit{0.031} & \textit{0.055} & \textit{-0.007} & \textit{-0.113} & \textit{0.044} & \textit{-0.019} \\
\cmidrule(lr){2-16}
& \multirow{4}{*}{\textit{Avg}} & \modelA{} & \textit{0.166 $\pm$ 0.041} & \textit{0.120 $\pm$ 0.020} & \textit{0.812 $\pm$ 0.009} & \textit{0.004} & \textit{0.031} & \textit{-0.037} & \textit{-0.010} & \textit{0.007} & \textit{0.031} & \textit{-0.009} & \textit{-0.069} & \textit{0.024} & \textit{0.028} \\
&  & \modelB{} & \textit{0.527 $\pm$ 0.045} & \textit{0.261 $\pm$ 0.014} & \textit{0.607 $\pm$ 0.013} & \textit{0.019} & \textit{0.080} & \textit{0.020} & \textit{0.004} & \textit{0.077} & \textit{0.060} & \textit{-0.043} & \textit{-0.140} & \textit{0.036} & \textit{-0.112} \\
&  & \modelC{} & \textit{0.221 $\pm$ 0.098} & \textit{0.253 $\pm$ 0.030} & \textit{0.800 $\pm$ 0.015} & \textit{0.019} & \textit{0.064} & \textit{-0.026} & \textit{-0.069} & \textit{0.035} & \textit{0.072} & \textit{-0.008} & \textit{-0.122} & \textit{-0.026} & \textit{0.061} \\
&  & \modelD{} & \textit{0.422 $\pm$ 0.049} & \textit{0.186 $\pm$ 0.017} & \textit{0.717 $\pm$ 0.017} & \textit{-0.004} & \textit{0.024} & \textit{0.005} & \textit{-0.011} & \textit{-0.022} & \textit{0.062} & \textit{-0.025} & \textit{-0.072} & \textit{0.061} & \textit{-0.019} \\
\bottomrule
\end{tabular}
\caption{Dialog acts (overall). JS = Jensen-Shannon distance on label proportions. Density Ratio = human labels/sentence $\div$ agent labels/sentence. Dialog Acts = human proportion $-$ agent proportion for each label (positive $\Rightarrow$ agent under-produces).}
\label{tab:acts_overall_2}
\end{table*}

\section{Information Density vs.\ Information Type}
\label{app:profile_length}

A natural concern with our experimental design is that the comparison across conditioning methods may not be fully information-balanced: the \history{} method receives 5--20 raw prior comments, whereas \profile{} and \behavior{} receive compressed summaries. Part of the gain observed under \history{} could therefore reflect a larger context budget (i.e., information \emph{quantity}) rather than a strictly better form of persona representation, i.e., information \emph{type}.

To probe this confound, we examine how performance varies with the amount of conditioning information provided. We focus this analysis on \profile{} length, where the available token budget varies meaningfully across users (see Table~\ref{tab:user_stats} and Figure~\ref{fig:user_distributions}). For \history{}, the analogous analysis would require comparing performance across different numbers of prior comments. However, the \history{} sampler defaults to using up to 20 comments per user, and in practice the vast majority of conditioning pools fall in the 16--20 range; lower-comment buckets are too sparsely populated to support a meaningful comparison. We therefore restrict the analysis to \profile{} length, which provides a well-distributed and interpretable proxy for input richness.

\begin{figure*}[ht!]
    \centering
    \includegraphics[width=\textwidth]{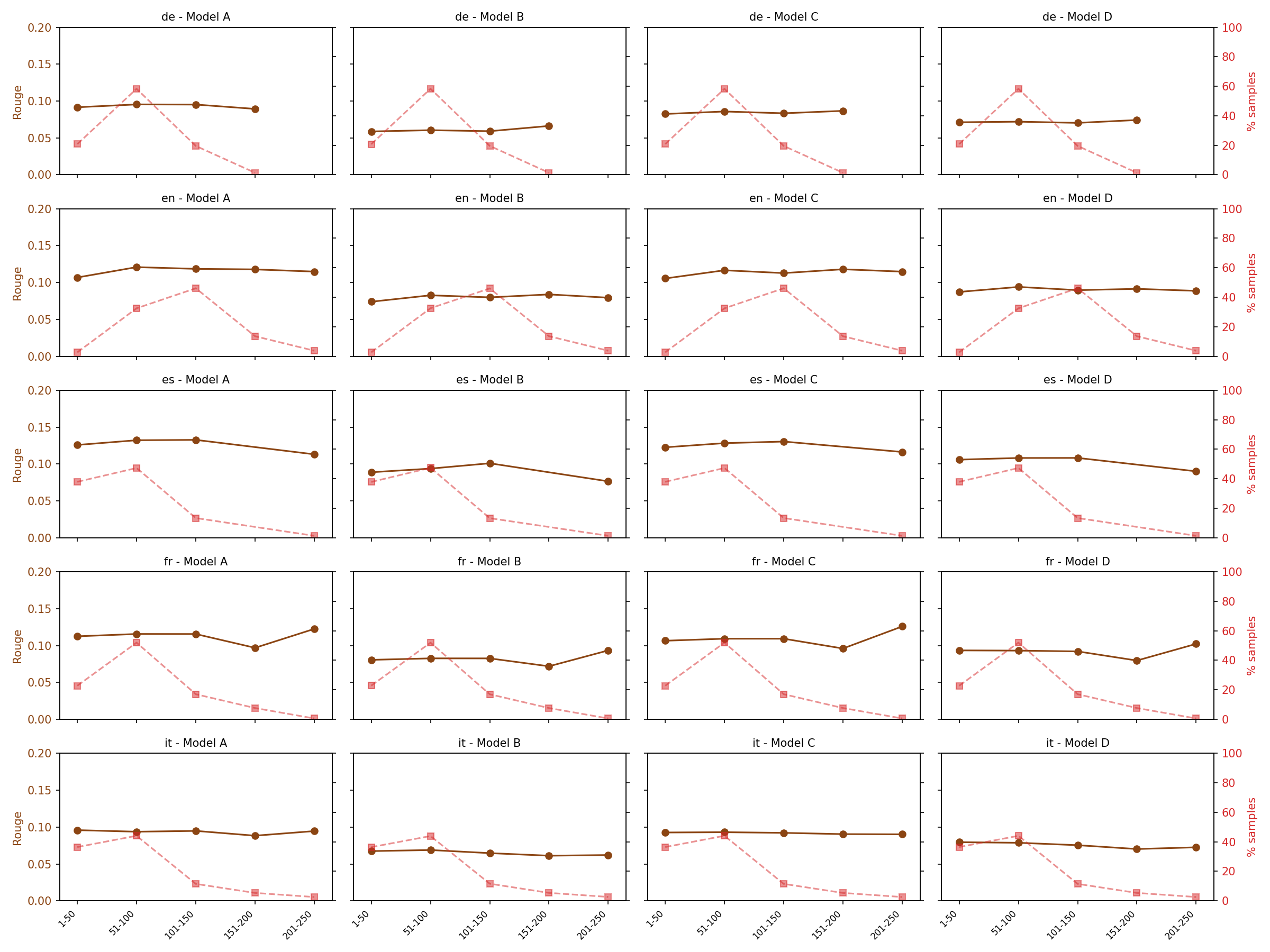}
    \caption{ROUGE-L between agent and human comments under \profile{} conditioning, stratified by \profile{} word count. Each row is a language; each column is a model. Brown solid line: ROUGE-L (left axis). Red dashed line: proportion of samples in each bucket (right axis).}
    \label{fig:profile_len_rouge}
\end{figure*}

\begin{figure*}[ht!]
    \centering
    \includegraphics[width=\textwidth]{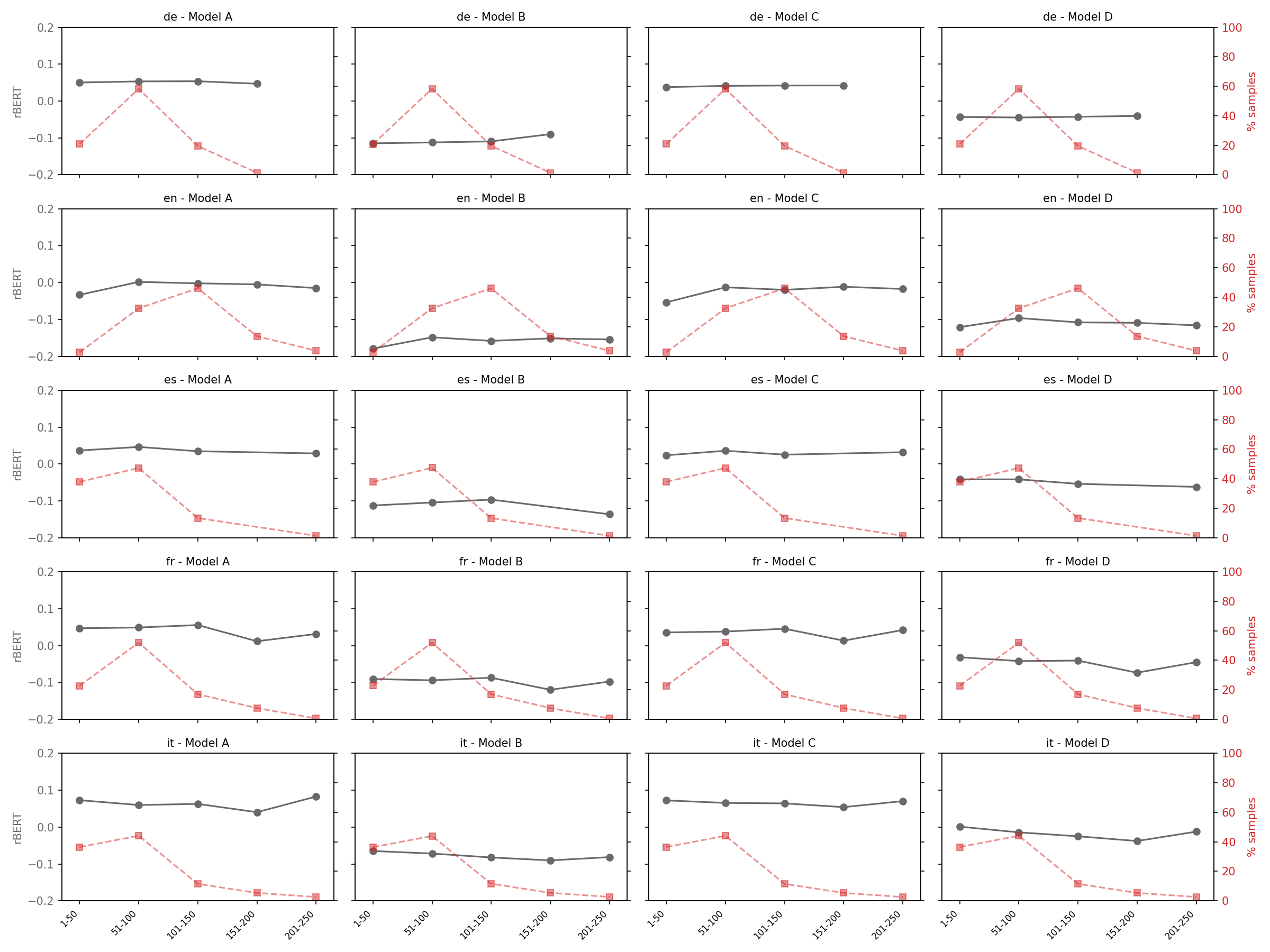}
    \caption{Rescaled BERTScore (rBERT) between agent and human comments under \profile{} conditioning, stratified by \profile{} word count. Conventions as in Figure~\ref{fig:profile_len_rouge}.}
    \label{fig:profile_len_rbert}
\end{figure*}

\begin{figure*}[ht!]
    \centering
    \includegraphics[width=\textwidth]{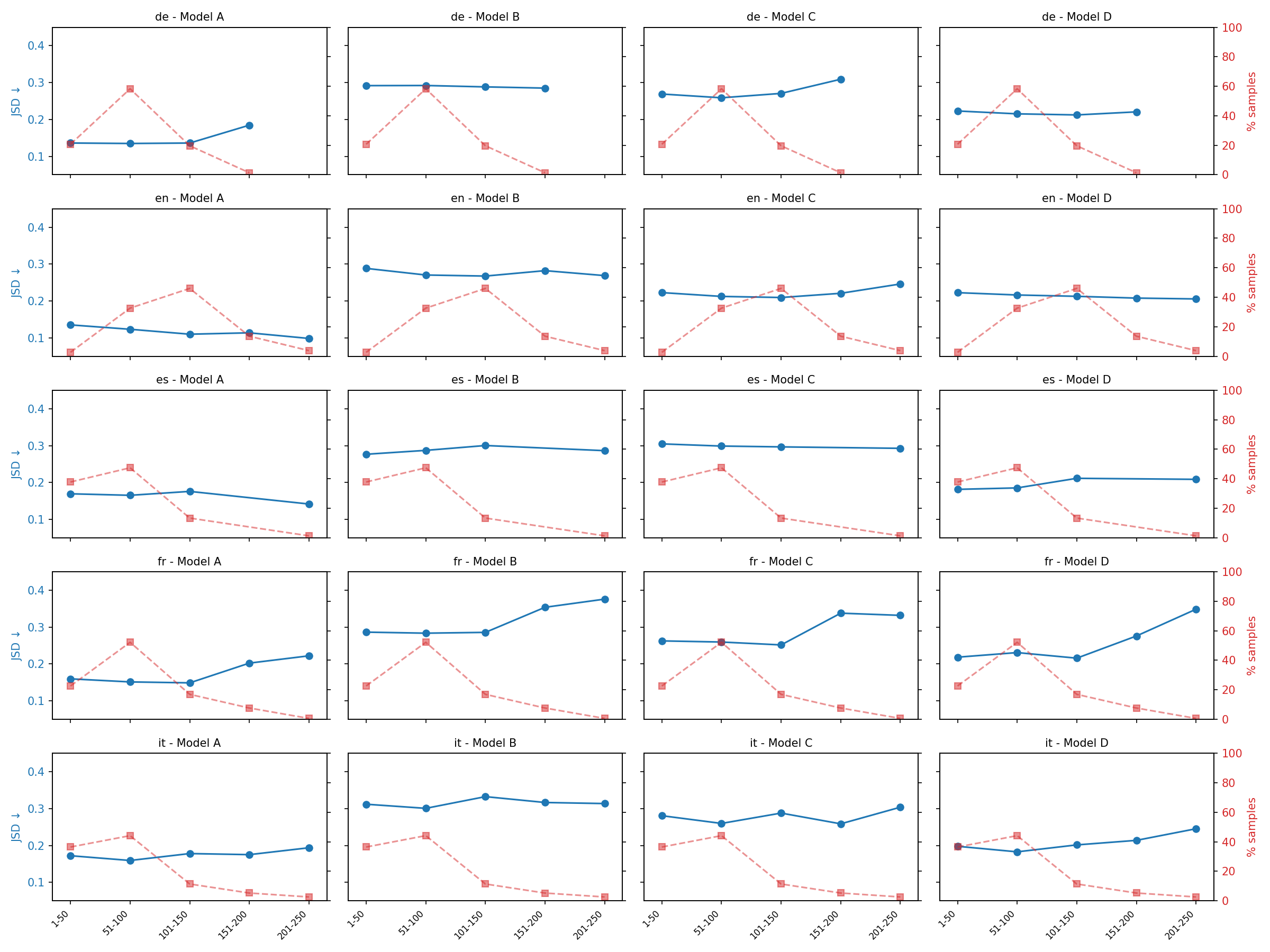}
    \caption{Dialog act JSD between agent and human distributions under \profile{} conditioning, stratified by \profile{} word count. Lower is better. Conventions as in Figure~\ref{fig:profile_len_rouge}.}
    \label{fig:profile_len_acts_jsd}
\end{figure*}

\begin{figure*}[ht!]
    \centering
    \includegraphics[width=\textwidth]{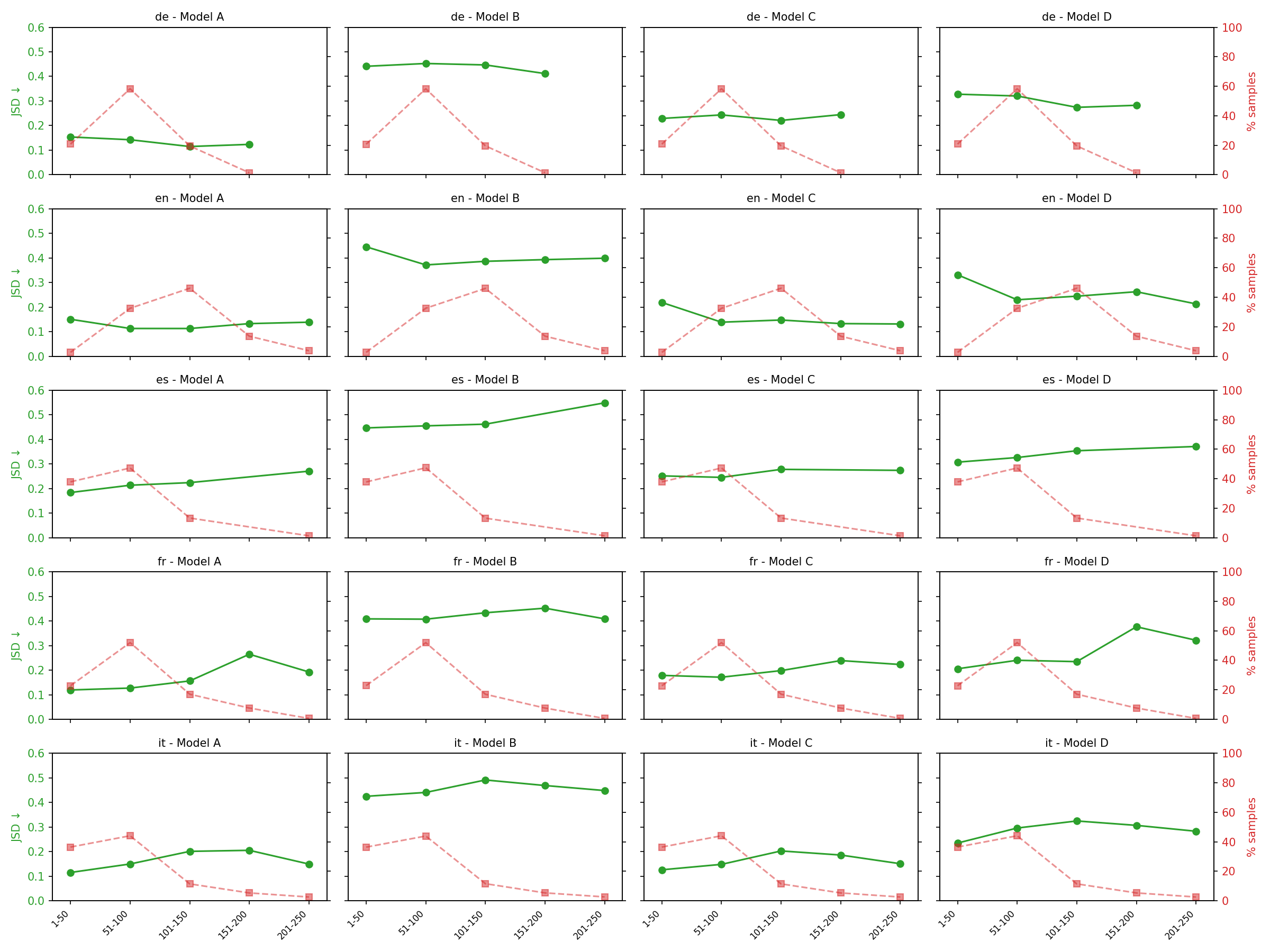}
    \caption{Sentiment JSD between agent and human distributions under \profile{} conditioning, stratified by \profile{} word count. Lower is better. Conventions as in Figure~\ref{fig:profile_len_rouge}.}
    \label{fig:profile_len_sent_jsd}
\end{figure*}

We stratify all experimental instances by \profile{} word count into five buckets (1--50, 51--100, 101--150, 151--200, 201--250) and compute, for each bucket and for each language--model pair, the four metrics used in our main analysis: ROUGE-L, rBERT, dialog act JSD, and sentiment JSD. Buckets with fewer than 20 samples are omitted, so \texttt{de} has no 201--250 bucket and \texttt{es} no 151--200 bucket. The \texttt{es} 201--250 bucket is dominated by a single user, who contributes 45 of its 67 samples, so its values should be read as largely individual rather than bucket-level. Results are shown in Figures~\ref{fig:profile_len_rouge}--\ref{fig:profile_len_sent_jsd}. The red dashed lines indicate the proportion of samples falling in each bucket; trends at the longest bucket should be interpreted with caution due to sample sparsity.

\paragraph{Findings}
Across all 20 language--model pairs, none of the four metrics improves consistently as \profile{} length grows. A pure context-budget account would predict a downward slope for JSD metrics (lower divergence = better alignment) and an upward slope for similarity metrics (higher = better) as more conditioning tokens are provided. ROUGE-L and rBERT are nearly horizontal in every configuration (within-pair ranges of at most $0.026$ and $0.044$, respectively). Where the JSD metrics do move, they mostly move in the \emph{opposite} direction: dialog act JSD rises for all four models in \texttt{fr}, mainly from the 151--200 bucket onward, and sentiment JSD rises with profile length for all four models in \texttt{es}, \texttt{fr}, and \texttt{it}. The few downward trends are small and confined to \texttt{en} and \texttt{de}: dialog act JSD for \modelA{} in \texttt{en} decreases from $0.135$ to $0.098$, and sentiment JSD decreases for some models in \texttt{en} and \texttt{de}, in \texttt{en} largely because of the sparse 1--50 bucket (under 3\% of samples).

Taken together, these results provide evidence against an information-density explanation for our main results. Together with the observation reported in the main text---that \profile{} \emph{degrades} performance relative to \base{} despite adding tokens---this analysis supports our interpretation that information \emph{type}, not information \emph{quantity}, drives the differences across conditioning methods. We acknowledge that this analysis does not fully isolate every possible confound (e.g., compression-induced loss of stylistic exemplars), and we discuss this further in the Limitations.

\section{Qualitative Audit of Profiles and Behavioral Traits}
\label{app:qualitative_audit}

Because profiles and behavioral traits are produced automatically by an LLM and then used as the key conditioning signals for \profile{}, \behavior{}, and \behaviorreasoning{}, their quality directly affects the interpretation of downstream results. To complement the quantitative analyses in the main text, we conducted a manual audit of these representations along three dimensions: \textit{(i)} \textbf{relevance and faithfulness} of generated profiles with respect to the source user pages, \textit{(ii)} \textbf{coherence} between assigned behavioral traits and the supporting reasoning produced by the extractor model, and \textit{(iii)} \textbf{privacy preservation}, i.e., whether sensitive information (real names, usernames, contacts) is leaked into the generated representations. 
We sampled 10 users per language (50 total) uniformly at random from the experimental subset and one of the authors inspected, for each user, the generated profile, all 8 behavioral traits with their reasoning, and the underlying user page.

\paragraph{Profile faithfulness}
In all 50 cases, profiles contain information that is relevant to and supported by the source user page. We identified a small number of imprecisions, none of which substantially distort the persona:
\begin{itemize}[topsep=2pt,itemsep=0pt,leftmargin=*]
    \item \textbf{English:} one numerical drift (``42 years'' instead of ``45 years''); one rounding error on edit count (``100{,}000'' instead of ``100{,}000k'');
    \item \textbf{French:} one case where contributed-article information was omitted;
    \item \textbf{German:} one case where the username was incorporated into the profile as if it were a separate entity, likely due to hallucination triggered by user page ambiguity;
    \item \textbf{Italian:} one ungrammatical fragment (``Ha adottato il giorno 16 luglio e il comune di Plesio'') reflecting parsing ambiguity in the user page.
\end{itemize}
We additionally observed that information from \textit{userboxes}---small standardized templates that Wikipedia editors place on their user pages to declare interests, languages spoken, locations, or affiliations---is sometimes incorporated into the profile and sometimes ignored, with no clear pattern across languages. Moreover, we noticed that gender, when not explicitly self-declared, is typically inferred from gender-marked terms (especially in the four grammatical-gender languages: German, Spanish, French, and Italian) and, when available, from given names.

\paragraph{Behavioral trait coherence}
For each of the 8 behavioral categories, we checked whether the assigned trait value is consistent with the accompanying reasoning. Out of $50 \times 8 = 400$ trait-reasoning pairs, in only 7 cases (across English, Spanish, and Italian) was the assigned label found to trace back to the provided reasoning. In most other cases, the reasoning explicitly cites concrete excerpts from the user's comment history in support of the assigned trait, providing transparent grounding for the conditioning signal.

\paragraph{Privacy}
The audit surfaced two cases that warrant explicit mention:
\begin{itemize}[topsep=2pt,itemsep=0pt,leftmargin=*]
    \item one English profile mentioning a middle name, which the user themselves had self-disclosed anecdotally on their user page;
    \item one German profile reproducing the username as if it were a separate named entity (the same hallucination case noted above).
\end{itemize}
No real legal names, contact details, or other directly identifying information were leaked into the generated representations.

\paragraph{Implications for downstream results}
The audit indicates that the generated profiles and behavioral traits are largely faithful and well-grounded, with imprecisions concentrated at the level of isolated factual details rather than systematic distortions. This argues against an interpretation in which the relatively weak performance of \profile{} and \behavior{} is driven primarily by representation quality. At the same time, the audit highlights that profiles vary substantially in the kinds of information they emphasize (e.g., uneven uptake of userbox content) and that the compression step inherent in summarization may lose stylistic exemplars present in the raw \history{} signal. In this sense, our results should be read as conclusions about the conditioning \emph{representations as we generated them}, rather than as universal statements about all possible sociodemographic descriptors. We acknowledge this scope explicitly in the Limitations.

\end{document}